%% file: main.tex
\documentclass[]{arxiv}
\input{prelude.tex}

\begin{document}

\ABSTRACT{We consider what we call the offline-to-online learning setting, focusing on stochastic finite-armed bandit problems.
In offline-to-online learning, a learner starts with offline data collected from interactions with an unknown environment in a way that is not under the learner’s control. Given this data, the learner begins interacting with the environment, gradually improving its initial strategy as it collects more data to maximize its total reward.
The learner in this setting faces a fundamental dilemma: if the policy is deployed for only a short period, a suitable strategy (in a number of senses) is the Lower Confidence Bound (LCB) algorithm, which is based on pessimism. LCB can effectively compete with any policy that is sufficiently "covered" by the offline data. However, for longer time horizons, a preferred strategy is the Upper Confidence Bound (UCB) algorithm, which is based on optimism. Over time, UCB converges to the performance of the optimal policy at a rate that is nearly the best possible among all online algorithms.
In offline-to-online learning, however, UCB initially explores excessively, leading to worse short-term performance compared to LCB. This suggests that a learner not in control of how long its policy will be in use should start with LCB for short horizons and gradually transition to a UCB-like strategy as more rounds are played. This article explores how and why this transition should occur.
Our main result shows that our new algorithm performs nearly as well as the better of LCB and UCB at any point in time. The core idea behind our algorithm is broadly applicable, and we anticipate that our results will extend beyond the multi-armed bandit setting.
}


\maketitle

\input{intro}

\input{litreview}
\input{setting}
\input{algorithm}
\input{comparing}

\input{experiments}

\input{proofs}

\input{conc}
\input{acknowledgements}

\bibliography{sample}
\newpage
\input{appendix}

\end{document}

%% file: prelude.tex
\usepackage[english]{babel}
\usepackage{amsmath,amssymb,amsfonts}

\usepackage{xspace}
\usepackage{natbib}
\TheoremsNumberedThrough
\usepackage[letterpaper,top=2cm,bottom=2cm,left=3cm,right=3cm,marginparwidth=1.75cm]{geometry}
\usepackage[shortlabels]{enumitem}
\usepackage{dsfont}
\usepackage{mathtools}

\usepackage{caption}
\usepackage{subcaption}
\usepackage{graphicx}
\usepackage[colorlinks=true, allcolors=blue]{hyperref}
\usepackage[size=tiny]{todonotes}

\usepackage{titlesec}
\titleformat{\paragraph}
{\normalfont\normalsize\bfseries}{\theparagraph}{1em}{}
\titlespacing*{\paragraph}
{0pt}{3.25ex plus 1ex minus .2ex}{1.5ex plus .2ex}

\usepackage[capitalize]{cleveref}
\usepackage[ruled,linesnumbered]{algorithm2e}
\newcommand{\algoname}{\textsc{OtO}\xspace}

\crefname{algocf}{Algorithm}{algs.}
\Crefname{algocf}{Algorithm}{Algorithms}

\newcommand{\algucb}{\textsc{UCB}\xspace}
\newcommand{\alglcb}{\textsc{LCB}\xspace}

\usepackage{tikz}
\usepackage{standalone}
\ARTICLEAUTHORS{%
\AUTHOR{Flore Sentenac}
\AFF{HEC Paris,
\EMAIL{sentenac@hec.fr}}
\AUTHOR{Ilbin Lee}
\AFF{University of Alberta,
\EMAIL{}}
\AUTHOR{Csaba Szepesvari}
\AFF{University of Alberta,
Deepmind London}
}

\usepackage{soul}
\usepackage{booktabs}

\usepackage{multirow}

\TITLE{Balancing optimism and pessimism in offline-to-online learning}

\newcommand{\cP}{\mathcal{P}}
\newcommand{\R}{\mathbb{R}}

%% file: intro.tex
\section{Introduction}

Sequential learning methods have been developed for a wide range of applications, from clinical trials \citep{thompsonclinical, zhou2024sequential} to digital advertising \citep{advertising, bastani2020online}. Typically, the focus has been to find a good balance between \textit{exploration}, which is gathering new information, and \textit{exploitation}, which is leveraging the already available information to maximize collected reward. 
Much of the online learning literature assumes that no information outside of the one collected by the learner is available. However, in many applications to which sequential learning methods have been applied, past interaction data is available upfront before any interaction happens. 

This is where offline-to-online learning is relevant. In offline-to-online learning, the agent has access to a historical dataset resulting from past interactions with the learning environment and then has the opportunity to interact and collect rewards in the environment for a number of rounds. The number of rounds may be fixed and known to the learner or unknown. This setting is a mix between the fully online and fully offline ones and thus presents its own set of challenges. Interest in this problem has grown in recent years, as good offline-to-online methods hold the promise of reducing the cost of online exploration by fully leveraging available offline data \citep{zheng2023adaptivepolicylearningofflinetoonline,lee2021offlinetoonlinereinforcementlearningbalanced,Song2022-vg,Li2023-lv,bu2021onlinepricingofflinedata,MABwithHistory,chen2022data}. 

Offline-to-online learning covers the continuum from the fully offline setting to the fully online one as the number of rounds of interaction increases. However, we believe that the current literature lacks a theoretical treatment encompassing the whole spectrum of offline-to-online learning, as we elaborate in the next section.   
Intuitively, if the number of online rounds is ``small’’ (can be as small as one), there is limited opportunity for exploration, and one should consider exploiting early while relying heavily on the offline data. Conversely, a large number of rounds gives ample room for exploration, and promoting exploration is a reasonable approach in this case. This suggests that different approaches are required at different points along the offline-to-online spectrum, a notion supported by past work, where studies at the two extremes adopt distinct and sometimes opposing strategies.

As mentioned before, in the fully online setting, where historical data are not available, the main challenge lies in finding a good balance between exploration and exploitation. Various strategies have been developed to address this trade-off. Although there are other approaches, such as the $\varepsilon$-greedy strategy and Thompson Sampling \citep{agrawal2012analysis}, a much-celebrated principle to balance the trade-off is optimism in the face of uncertainty.  For bandit problems, a standard way to implement the principle is the Upper Confidence Bound (UCB) algorithm, using statistical confidence intervals for the value of different arms and selecting an arm whose confidence interval has the highest upper end \citep{auer2010ucb}. This method has been shown to enjoy asymptotic and minimax optimality \citep{lattimore2020bandit}.

In offline learning, a setting also called \textit{batched policy optimization}, the learner must infer a good policy based on a given dataset, with no possibility to further adjust the policy found based on interactions with the environment. The literature on this setting is not as vast as that of online learning, but it has also gained traction over the past decades. 
The difficulty of offline learning is inherently related to the quality of offline dataset: insufficient coverage of the environment may cause erroneous overestimation of values, which can lead to sub-optimal policies 
(e.g., \citealt{fujimoto2019offpolicydeepreinforcementlearning}). Past studies have proposed using the pessimism principle to mitigate this problem. 
The idea of pessimism has been studied broadly in various literature, often under different names, including robust optimization \citep{ben2002robust, bertsimas2011theory}. For example, the principle has been used extensively for inventory and pricing problems (e.g., \citealt{bertsimas2006robust, bu2023offline, perakis2008regret, xu2022robust}). The pessimism idea has also been studied recently in the reinforcement learning literature \citep{swaminathan2015batch, wu2019behavior,kidambi2020morel, li2022pessimismofflinelinearcontextual, jin2022policy, rashidinejad2023bridgingofflinereinforcementlearning, yin2021towards} and theoretical guarantees have been established. In bandit problems, this principle is implemented as the Lower Confidence Bound (LCB) algorithm, which, just like UCB, also builds a statistical confidence interval for each arm but then picks the arm with the highest \emph{lower} bound. It has been shown that, while both UCB and LCB are minimax optimal for offline learning in multi-armed bandits (MABs), LCB has a better guarantee than UCB for a weighted minimax criterion that is more refined to reflect the inherent difficulty in learning an optimal policy \citep{xiao2021optimalitybatchpolicyoptimization}.
Another advantage of pessimism shown in the literature is the competitiveness against the policy that generated offline data, called the logging policy. Because the logging policy is often the current policy that has been used in practice, it is critical to ensure that the performance of a new policy is not degraded compared to it.  Several approaches based on the pessimism principle have been introduced with theoretical guarantees for the regret against the logging policy \citep{xie2022armormodelbasedframeworkimproving, cheng2022adversarially}. 

From this overview, we could extract the following principle: optimism is a natural approach for online learning, while pessimism is preferred in offline learning. Then, what about offline-to-online learning? Intuitively, one might expect pessimism to perform better for small horizons, with optimism taking over for larger horizons. How significant is this effect? Moreover, one expects that the point at which optimism surpasses pessimism is problem-dependent. Can we identify that ``inflection point'', or better yet,  can we develop an algorithm that automatically finds it? Note that those questions arise regardless of whether the horizon is known or unknown to the learner.

In this paper, we focus on answering these questions for the MAB setting. Although MAB is the simplest reinforcement learning problem, it has been applied to various applications, including assortment selection and online advertising \citep{caro2007dynamic,pandey2007bandits, gur2020adaptive}, and insights gained in the bandit setting often generalize beyond them. However, the critical questions have not been studied in the literature, as is explained and discussed in more detail in the literature review.

As outlined, finding the right balance between pessimism and optimism, represented by the classic LCB and UCB algorithms, is the central challenge in the offline-to-online setting. Our first contribution is a new algorithm that automatically finds this balance, adjusting between pessimistic and optimistic approaches as needed. 

We support the design of our algorithm via a thorough study of the performance of UCB and LCB, and our algorithm. We study across the whole offline-to-online spectrum, by which we mean for varying horizon $T$.  While the minimax criterion is commonly used to evaluate algorithms, it fails to distinguish between LCB and UCB in the offline setting, as both are minimax optimal \citep{xiao2021optimalitybatchpolicyoptimization}. However, a comparison against the logging policy highlights an additional advantage of pessimism in offline learning. Motivated by this, we evaluate performance using both regret against the logging policy and regret against optimality. 
As expected, we find that neither  UCB nor LCB outperforms the other across both metrics along the whole spectrum of offline-to-online learning and under different offline data compositions. In contrast, our algorithm remains competitive with both UCB and LCB under both regret measures, achieving a principled balance between optimism and pessimism across the offline-to-online learning continuum.

%% file: litreview.tex
\section{Related Works}

\textbf{Offline Learning}: We begin by providing an overview of the research conducted in offline (or batch) learning, with a particular focus on works addressing the multi-armed bandit (MAB) setting in detail. A central challenge in offline learning is data coverage: does the offline dataset contain sufficient information about an optimal or near-optimal policy? There are at least two extreme offline data types \citep{rashidinejad2023bridgingofflinereinforcementlearning}: expert data sets, produced by a near-optimal policy, and uniform coverage data sets, which provide equal representation of all actions. In expert data sets, imitation learning algorithms are shown to have a small sub-optimality gap against the logging policy \citep{Imitationlearningross,rajaraman2020fundamentallimitsimitationlearning}. Meanwhile, theoretical guarantees for many offline RL algorithms depend on the coverage of offline data—often quantified through a concentrability coefficient that measures how well the dataset covers the policies being evaluated \citep{rashidinejad2023bridgingofflinereinforcementlearning}; \cite{rashidinejad2023bridgingofflinereinforcementlearning}) or the set of policies with which the algorithms compete is limited to those covered in offline data \citep{cheng2022adversarially,  yin2020nearoptimalprovableuniformconvergence}. In this paper, we also consider different offline data compositions (uniform vs. skewed), but the main focus is the spectrum from offline to online learning.

A widely accepted intuition in offline learning is that a good policy should avoid under-explored regions of the environment. This motivates the use of the pessimism principle, which manifests in various forms: learning a pessimistic value function \citep{swaminathan2015counterfactualriskminimizationlearning,wu2019behaviorregularizedofflinereinforcement,li2022pessimismofflinelinearcontextual}, pessimistic surrogate \citep{buckman2020importancepessimismfixeddatasetpolicy}, or planning with a pessimistic model \citep{MorelPessimistisOfflineLearning,yu2020mopomodelbasedofflinepolicy}. 
Despite the surge of work in that direction, a theoretical base for the principle has only recently been developed. 
\cite{xiao2021optimalitybatchpolicyoptimization} analyzed pessimism in MABs, proving that both UCB and \alglcb are minimax optimal (up to logarithmic factors) but that \alglcb outperforms UCB under a weighted minimax criterion reflecting the difficulty of learning the optimal policy.  Notably, this result implies that \alglcb outperforms \algucb in cases where the offline data is generated by a near-optimal policy. \cite{rashidinejad2023bridgingofflinereinforcementlearning} further studied \alglcb, introducing a concentrability coefficient to measure how close the offline dataset is to being an expert dataset. They showed that \alglcb is adaptively optimal in the entire concentrability coefficient range for contextual bandits and MDP. They also showed that for MAB, \alglcb is also adaptively optimal for a wide set of concentrability coefficients, excluding only datasets where the optimal arm is drawn with probability larger than $1/2$ and where "play the most played arm" achieves an exponential convergence rate to the optimal policy.
It is important to note that these findings do not contradict the results of \cite{xiao2021optimalitybatchpolicyoptimization}.  \cite{rashidinejad2023bridgingofflinereinforcementlearning} fixed the concentrability coefficient in the minimax analysis, whereas \cite{xiao2021optimalitybatchpolicyoptimization} did not. Finally, \cite{rashidinejad2023bridgingofflinereinforcementlearning} also showed that the \alglcb algorithm can compete with any target policy that is covered in the offline data, regardless of the performance of that policy.

Two key insights emerge from these theoretical studies:
minimax regret does not fully capture the performance gap between UCB and \alglcb in offline learning, and \alglcb outperforms UCB when the logging policy is close to optimal. 
Because it is crucial to find a policy that is at least comparable to a previously deployed policy, there has been a series of works in offline RL literature that studied the regret against the logging policy or any baseline policy. Some of those methods are based on the pessimism principle. \cite{xie2022armormodelbasedframeworkimproving} proposed optimizing the worst-case regret against a baseline policy, which they called relative pessimism. They showed that the method finds a policy that performs as well as the baseline policy and learns the best policy among those supported in offline data when the baseline policy is also supported in the data. \cite{cheng2022adversarially} proposed a different optimization formulation also based on relative pessimism and showed that the algorithm improves over the logging policy. They also showed that the method can compete with policies covered in offline data. Other methods to improve over the logging policy that are not (directly) based on pessimism have been proposed as well, for example, those based on the idea that the learned policy should not be too far from the logging policy (e.g., \citealt{fujimoto2021minimalist}). However, these methods focus exclusively on the offline setting and thus only explore one extreme of the offline-to-online spectrum. Our work shares their objective of competing against the logging policy but extends beyond purely offline learning.  

\textbf{Offline-to-Online} 
While the literature on offline-to-online learning is still new, it is rapidly growing. A first line of work can be considered answering the following question: how can we adapt an online learning algorithm to achieve good performance when the offline data set is rich enough to learn a near-optimal policy? In some cases, classical online learning methods that incorporate offline data naturally achieve strong performance. For MABs, \cite{MABwithHistory} showed that a logarithmic amount of offline data can reduce the regret from logarithmic to constant, and their algorithm implements optimism but incorporates offline data. Results of \cite{gur2020adaptive} imply a lower bound for regret in offline-to-online MABs (see Appendix G of \citealt{bu2021onlinepricingofflinedata} for a summary), offering more insights into how additional data influences achievable regret. Their setting is broader than ours, allowing for new information to arrive sequentially, with offline data as a special case. They, too, employed classical online learning methods, including Thompson sampling and UCB-like algorithms augmented with offline data.
This question was also studied in dynamic pricing by \cite{bu2021onlinepricingofflinedata},  a class of contextual bandits. The authors of this last paper proposed an optimism-driven algorithm and showed that the achievable regret as a function of $T$ decreases notably as the size of the offline data set increases, which the authors coined as a ``phase transition''. It is also notable that for MABs, the optimal regret phase transition is achieved only when offline data is balanced over the arms \citep{bu2021onlinepricingofflinedata}. \cite{chen2022data} also investigated offline-to-online learning, focusing on healthcare applications where offline and online data distributions may differ. Their approach, like the others, is based on optimism or Thompson sampling.

Our work builds on this literature in a fundamental way. The above papers enhance classical online learning methods, such as UCB-like or Thompson-sampling, by incorporating additional offline data and studying how this affects regret against optimality. Here, we bring offline learning methods and metrics into the offline-to-online learning setting and thus put both offline and online learning into perspective. 

Another stream of offline-to-online works focuses on minimizing computational and sample costs rather than minimizing regret over a horizon. This setting with that objective is more commonly called hybrid reinforcement learning. Their goal is to find a good policy using both offline and online data while minimizing the number of samples or computational resources required \citep{Song2022-vg,xie2022policyfinetuningbridgingsampleefficient,ball2023efficient,
wagenmaker2023leveraging,Li2023-lv, li2024reward, zhou2023offlinedataenhancedonpolicy}. These works differ from ours in both goal and methodology: they do not address the central question of how best to navigate the transition from offline to online learning as TT varies. Furthermore, our focus on MABs sidesteps computational efficiency concerns.

Despite these prior contributions, the theoretical understanding of offline-to-online learning remains incomplete, particularly regarding how to balance conservatism and exploration as learning shifts from purely offline to purely online. Our paper directly addresses this gap, with a specific focus on MABs.

%% file: setting.tex
\section{Setting and Notation}

We consider a multi-armed bandit problem with $K$ arms. At the beginning the learner is given access to $m_i$ rewards sampled from $\mathcal{P}_i$, a distribution associated with arm $1\leq i \leq K$, which is initially unknown to the learner. We let $\mu_i \in [0,1]$ denote the mean of $\mathcal{P}_i$, which is assumed to be a 1-subgaussian. That is, for any $\lambda\in \R$ real number, for any $1\leq i \leq K$,
\begin{align*}
\int \exp(\lambda (x-\mu_i)) \cP_i(dx) \le \exp( \lambda^2/2) \,.
\end{align*}
We let $\Theta= \mathcal{G}_1^K \times \mathbb{N}^K$ and for $\textbf{m}\in \mathbb{N}^K$, we let $\Theta_{\mathbf{m}}= \mathcal{G}_1^K \times \textbf{m}$ where $\mathcal{G}_1$ is the set of 1-subgaussian distributions over the reals.
 We denote the maximum value of the means by $\mu^*:= \max_i \mu_i$, and denote the suboptimality gap of an arm $i$ by $\Delta_i=\mu^*-\mu_i$.  We denote $\hat{\mu}_i^0$ the empirical mean of the $m_i$ a priori observed rewards from arm $i$ observations, and let $m=\sum_{i \in [K]}m_i$. The implicit assumption here is that the means and sample sizes contain all the information the learner will need to know about the data. We also use the shorthands:
\[
\hat{\mu}_0:= \frac{\sum_{i \in [K]}m_i\hat{\mu}_i^0}{m},\quad  \mu_0:= \frac{\sum_{i \in [K]}m_i\mu_i}{m} \text{ and }\Delta_0= \mu_*-\mu_0.
\]
Here $\mu_0$ is the mean reward of the policy that chooses arm $i$ with probability $\pi_i= \frac{m_i}{m}$. While we keep $\mathbf{m}= \left(m_i\right)_{i \in [K]}$ non-random, we will think of the data as being collected from following the policy $\pi= \left(\pi_i\right)_{i \in [K]}$, which we call the logging policy. We expect our results to extend with little change to the case when $\mathbf{m}= \left(m_i\right)_{i \in [K]}$ is the random number of times arm $i$ is chosen when $\pi$ is followed for $m$ steps.
At every round  $t=1,\ldots,T$, the learner pulls arm $I(t)$ and receives reward $x^t_{I(t)}$, where $x_i^t\sim \mathcal{P}_i$, $i\in [K]$, denotes the reward of arm $i$ at time $t$.  Over horizon $T$, we define the pseudo-regret as

\[
R(T)= 
 T\mu^* - \sum_{t=1}^T\mu_{I(t)}.
\]

While we define the regret with respect to the logging policy as

\[
R^{\text{log}}(T)=
T\mu_0 - \sum_{t=1}^T\mu_{I(t)}.
\]

Each combination of an instance $\theta\in \Theta$ and an algorithm $\mathcal{A}$ followed by a learner induces a probability distribution over $([K]\times \R)^\mathbb{N}$, 
the sequences of arm pulls and rewards of arbitrary length.
We denote by the subscripts $\theta$ and $\mathcal{A}$ the expectations and probabilities induced by the pair $\theta$ and $\mathcal{A}$. The worst-case expected regret of an algorithm $\mathcal{A}$ over $\Theta_{\mathbf{m}}$ and horizon $T$ is  defined as
\[
\mathcal{R}_\mathcal{A}(T)= \sup_{\theta \in \Theta_{\mathbf{m}}}\mathbb{E}_{\theta, \mathcal{A}}[R(T)].
\]
Similarly, we let
\[
\mathcal{R}^{\text{log}}_\mathcal{A}(T)= \sup_{\theta \in \Theta_{\mathbf{m}}}\mathbb{E}_{\theta, \mathcal{A}}[R^{\text{log}}(T)],
\]
with the understanding that $\mathbf{m}$ will be clear from the context when these are used.
Note that the expectations are taken with respect to the randomness of both the offline data and the rewards incurred during the interaction, in addition to any randomness coming from the algorithm. Also, in the above definitions, the number of observations per arm is fixed.

At round $t$, we will let $T_i(t)$ denote the number of times arm $i$ has been pulled so far and $\hat{\mu}_i^t$ the empirical mean for arm $i$ combining offline and online samples. We also let

\[
\overline{\mu}_i(t)=\hat{\mu}_i^t+\sqrt{\frac{\log(K/\delta)}{2(m_i+T_i(t))}},
\]
and 
\[
\underline{\mu}_i(t)=\max_{t'\leq t} \left\{\hat{\mu}_i^{t'}-\sqrt{\frac{\log(K/\delta)}{2(m_i+T_i(t'))}}\right\}.
\]
These values are constructed so that $\underline{\mu}_i(t)$ is increasing with $t$ and the following lemma holds:
\begin{lemma}\label{lem:hoeff}
    For any instance $\theta$, any algorithm $\mathcal{A}$, and any $T\geq0$ it holds that
    \[
    \mathbb{P}_{\theta,\mathcal{A}}\left(\exists i \in [K] \text{ and } \exists 0\leq t\leq T \text{ s.t. } \overline{\mu}_i(t)<\mu_i\right)\leq T\delta,
    \]
    and
     \[
    \mathbb{P}_{\theta,\mathcal{A}}\left(\exists i \in [K]\text{ and } \exists 0\leq t\leq T \text{ s.t. } \underline{\mu_i}(t)>\mu_i\right)\leq T\delta.
    \]
\end{lemma}

\textit{Proof}: Both inequalities hold by Hoeffding's inequality \citep{Vershynin_2018} and a union bound over  $K$ arms and $T$ rounds. \hfill \(\Box\)

In the rest of the paper, we omit the subscripts $\theta$ and $\mathcal{A}$ when it is clear from the context.

%% file: algorithm.tex

\section{Offline to Online (\algoname) Algorithm}\label{sec:algorithm}

As it was already mentioned in the introduction and will be demonstrated in \cref{sec:intersetanalysis}, neither \alglcb nor \algucb is the best algorithm for the whole range of horizon length in the offline-to-online setting. Before introducing our algorithm, it will be instrumental in taking a peek at what goes wrong with these methods.

For \algucb, consider a case where all arms but one are observed many times in the offline data. If the horizon is short (can be as short as $1$), then \algucb will pull only the unexplored arm, even if there are good options among the sampled arms. Thus, for a short horizon, it will over-explore. In particular, \algucb will suffer large regret against the logging policy. We will see in \cref{sec:intersetanalysis} that this phenomenon can hinder the performance of \algucb even in cases less extreme than this. Now, as to the failure of \alglcb, its lack of exploration poses a significant issue in cases when good arms are under-sampled in the offline data. As such, it will suffer large regret relative to the optimal arm, especially for large horizons.

As opposed to these, our new algorithm, \algoname, will be shown to achieve both a low regret with respect to the optimal arm and a low regret with respect to the logging policy, no matter what the horizon length is, and thus, automatically finds a balance between optimism and pessimism. Notably, it retains the advantages of both \alglcb and \algucb while avoiding their weaknesses.

At every round, our algorithm computes an exploration budget, detailed in the following paragraph. If the budget is high enough to explore safely, then \algucb, i.e., $U(t):= \argmax_{i \in [K]} \overline{\mu}_i(t)$  is played. Otherwise, the algorithm defaults to the safe option, which is \alglcb, i.e., $L(t):=\argmax_{i \in [K]} \underline{\mu}_i(t)$. At the high level, our algorithm design was inspired by conservative bandits algorithms \citep{wu2016conservative}. 

Now, let us detail the computation of the exploration budget. We will start with the description in the case where the horizon $T$ is known. We explain later how to deal with the unknown horizon case.  Define
\[
\beta\coloneq\frac{\sum_i \sqrt{m_i}}{m}\sqrt{2\log(K/\delta)},
\]
and for $\alpha>0$, a tuning parameter whose role will be explained later, let
\[
\gamma\coloneq \underline{\mu}_{L(0)}(0)-\alpha\beta,
\]
a ``safe'' lower bound on reward collected by the logging policy. In fact, $\gamma$ is also a safe lower bound on the reward that will be collected by the LCB action at every time step. This is because,
by construction, at every iteration $t$, $\underline{\mu}_{L(t)}(t)\geq \underline{\mu}_{L(0)}(0)$, hence $\underline{\mu}_{L(t)}(t)-\alpha\beta \geq \gamma$. As will be seen later, the budget will use $\gamma$ as a benchmark, and this last inequality will imply that at every play of \alglcb, the reward collected exceeds our chosen benchmark by at least $\alpha \beta$. 
The reason we use a lower bound on the mean reward for LCB action at time step $0$ as the benchmark, as opposed to using, say, the actual reward (or an upper bound), is because meeting this benchmark will be easier, which makes the algorithm explore ``earlier'', which is expected to improve performance, while the algorithm is still able to compete with the logging policy.

For each arm $i \in [K]$, we denote $T_i^U(t)$ the number of times arm $i$ has been explored as the \algucb arm, i.e., in iterations where the budget was high enough. We also denote  $T^L_i(t)$ the number of times where arm $i$ was played as the \alglcb arm. Note that $T_i(t)=T^L_i(t)+T_i^U(t)$. We also introduce the notation $T^L(t)\coloneq\sum_{i \in [K]}T_i^L(t)$ for the number of times \alglcb has been played up to iteration $t$. The budget at iteration $t$ is defined as
\begin{equation}\label{eq:budget}
B_{T}(t)\coloneq \sum_{i=1}^{K}T_i^U(t-1)(\underline{\mu}_i(t)-\gamma)+\underline{\mu}_{U(t)}(t)-\gamma+(T^L(t-1)+T-t)\alpha \beta.   
\end{equation}

This budget is the sum of three parts. As will be shown in the analysis, the first one, $\sum_{i=1}^{K}T_i^U(t-1)(\underline{\mu}_i(t)-\gamma)$, is a lower bound of the difference between the total cumulated reward of \algucb iterations played until iteration $t-1$, and a benchmark, $T^U_i(t-1)\gamma$. The second part, $\underline{\mu}_{U(t)}(t)-\gamma$ is a high probability lower bound of the reward that would be obtained by playing \algucb minus $\gamma$ in the current iteration. The rest of \eqref{eq:budget} reflect that, by design, each play of \alglcb increases the budget by $\alpha\beta$. The term $T^{L}(t-1) \alpha \beta$ is the budget accumulated by playing \alglcb up to iteration $t$. The term $(T-t)\alpha \beta$ is the total budget that would be accumulated if the algorithm plays \alglcb from iteration $t+1$ until the end. The underlying function of the budget is to ensure that the reward cumulated by the algorithm does not fall too far below the reward that \alglcb would cumulate.
Lastly, the parameter $\alpha$ determines how stringent the budget constraint is. As will be discussed after the statement of the main theorem, different values for $\alpha$ may be preferred depending on the favored objective.

In the case where the horizon $T$ is unknown, we can use a horizon-doubling technique to set the budget.  Let $\Tilde{T}$ denote a proxy horizon. Initially, we let $\Tilde{T}=2$. The budget is defined as before, but with $\Tilde{T}-t$ instead of $T-t$ before the last closing parenthesis. The algorithm proceeds as before, except that if $t$, the current iteration number, exceeds $\Tilde{T}$, we double the value of $\Tilde{T}$. 

Note that this modification alone is insufficient to handle the unknown horizon. In practice, the parameter $\delta$ in $\underline{\mu}_i(t)$ and $\overline{\mu}_i(t)$ is typically chosen as a function of the horizon, such as $\delta = 1/T^2$. To address this, one can replace $\delta$ with a time-step-dependent parameter, $\delta_t = \delta_0 / t^2$. Using this adjustment, the bounds derived for a fixed $\delta$ extend to the varying $\delta_t$, as $\delta_t \geq \delta_0 / T^2$ and $\sum_{t=1}^T \delta_t \leq \frac{\pi^2}{6} \delta_0$.
While effective, this approach could be further refined. Prior work (e.g., \citet{lattimore2016regretanalysisanytimeoptimally}) has explored more sophisticated methods for addressing unknown horizons in confidence-bound-based algorithms. Adapting such methods to our setting presents an interesting avenue for future research. However, we chose not to pursue this extension here to maintain focus on the core contributions of our work while avoiding additional technical complexity.

The full pseudo-code of the algorithm is given in  \cref{alg:algo1}. 

\begin{remark} 
The specific budget formula may seem a bit arbitrary to the reader. There can be multiple ways to define a budget to achieve the same purpose and ultimately obtain the same theoretical guarantee. We show in the additional experiments section of the appendix, \cref{app:extrasim} a few options and discuss in more detail the rationale behind the chosen formula.
\end{remark}

\begin{algorithm}[ht]
\DontPrintSemicolon
\SetKwInOut{Input}{input}
\Input{ $m_i, \hat{\mu}_i^0$ for $i \in [K]$, parameters $\alpha\geq 0,\delta$, horizon $T$ if known }
Let  $\beta:=\frac{\sum_i \sqrt{m_i}}{m}\sqrt{2\log(\frac{K}{\delta})}$\;
	Let $L(0):=\argmax_{i \in [K]}\underline{\mu}_{i}(0)$ and $\gamma= \underline{\mu}_{L(0)}(0)-\alpha\beta$\; 
 If horizon $T$ unknown, let $\Tilde{T}:=2$, if known let $\Tilde{T}:=T$\;
 \For{$t=1, \ldots,T$}{
 
 \If{$t>\Tilde{T}$}{
 Let $\Tilde{T}:=2\cdot \Tilde{T}$ \tcp*[f]{Update Horizon}
 }
 Compute $\underline{\mu}_i(t)$ and $\overline{\mu}_i(t)$ for each arm $i \in [K]$\;
 Let $U(t):=\argmax_{i \in [K]} \overline{\mu}_i(t)$\;
 Let $B_{\tilde{T}}(t):=\sum_{i=1}^{K}T_i^U(t-1)(\underline{\mu}_i(t)-\gamma)+\underline{\mu}_{U(t)}(t)-\gamma+(T^L(t-1)+\tilde{T}-t)\alpha \beta$\;
 \uIf(\tcp*[f]{Check Budget}){$B_{\tilde{T}}(t)>0$}{Pull $U(t)$ \tcp*[f]{Play \algucb}}
	\uElse{Pull $L(t)$ \tcp*[f]{Play \alglcb}}
 }
	\caption{\algoname} \label{alg:algo1}
\end{algorithm}

\begin{theorem}\label{thm:boundofftoon}
    For any $\theta\in \Theta$, any $1\leq t\leq T$, with probability at least $1-2T\delta$, the following hold:
    \begin{align}\label{eq:regretlogOtO}
            R^{\text{log}}_{\algoname}(t)\leq t\left(1+(1+\mathds{1}_{\text{T unknown}})\alpha\right)\beta\,,
    \end{align}
\begin{align}\label{eq:regretOtO}
       R_{\algoname}(t)\leq \sum_{i=1}^K\Delta_i\left(\frac{4\log(K/\delta)}{\Delta_i^2}-m_i\right)_++\frac{12 K\log(K/\delta)}{\alpha \beta}+K
\end{align} 
and
\begin{align}\label{eq:minimaxregretOtO}
    \mathcal{R}_\algoname(t)\leq
   \max_{J\subseteq [K]}2t\sqrt{\frac{2|J|\log(K/\delta)}{t+\sum_{j\in J}m_j}}+|J|+\frac{12 K\log(K/\delta)}{\alpha \beta}+2t^2\delta.
\end{align}   
\end{theorem}
\begin{remark}
    \cref{thm:boundofftoon} gives bounds for a fixed \(\delta\) only. We now provide some details on how these bounds would generalize to a time-varying \(\delta_t = \delta_0 / t^2\). First, bounds \cref{eq:regretlogOtO,eq:regretOtO} would hold with probability at least \(1 - \frac{\pi^2}{3}\delta_0\). Note that, conceptually, the value of the parameter \(\beta\) should not be updated at every iteration. Instead, it should retain the value it has on the first iteration, as it reflects the uncertainty about the value of the logging policy based on offline data, which does not need to be re-evaluated at each iteration. Specifically, \(\beta = \frac{\sum_i \sqrt{m_i}}{m} \sqrt{2\log(K/\delta_0)}\). In all of \cref{eq:regretlogOtO,eq:regretOtO,eq:minimaxregretOtO}, the \(\delta\) in the logarithmic factors should be replaced with \(\frac{\delta_0}{T^2}\), and \(t^2\delta\) in \cref{eq:minimaxregretOtO} should be replaced with \(\frac{\pi^2}{3}T\delta_0\).
\end{remark}

\textit{Sketch of Proof}: We bound separately the number of pulls of each sub-optimal arm at those iterations where the budget is insufficient, i.e., the arm is chosen by \alglcb, and at those iterations where there is enough budget, i.e., the \algucb arm is chosen. We bound the number of times a suboptimal arm is chosen by \algucb through the usual argument that the upper bound of a sub-optimal arm's value can only surpass $\mu_*$ a limited number of times when the confidence intervals hold. To bound the number of times a sub-optimal arm is chosen by \alglcb, we bound the total number of iterations where the budget is insufficient. A full proof is available in \cref{sec:boundregretalgo}. \hfill \( \Box\)

A few comments on the regret bounds of the algorithm. First, the upper bound on the regret against the logging policy, \cref{eq:regretlogOtO}, can be related to the bound on \alglcb's minimax regret against the logging policy (which will be obtained in \cref{prop:regretlogginglcb}): the leading terms in the two regret bounds are almost the same, $T\beta$ in the upper bound for \alglcb, and $T\left(1+(1+\mathds{1}_{\text{T unknown}})\alpha\right)\beta $ in the upper bound on \algoname. We can also relate \cref{eq:regretOtO} and \cref{eq:minimaxregretOtO} to the regret of \algucb (\cref{prop:regretminimaxucb}), where here the upper bound on the regret of \algoname\ (against optimality) is the sum of the upper bound on the regret of \algucb and a term that scales with $\frac{1}{\alpha}$. The additional term is the cost of the budget constraint. As will be seen in the next section, the bounds in \cref{prop:regretlogginglcb,prop:regretminimaxucb} are tight (up to poly-logarithmic factors) in a range of parameters $\mathbf{m}$. Within this range, it follows then that \algoname\ behaves marginally no worse than the best of the two algorithms along both of the objectives, the regret against an optimal arm and the regret against the logging policy.

\begin{remark}
High values of $\alpha$ loosen the budget constraint. As a consequence, the algorithm behaves closer to \algucb when $\alpha$ is large. The additional cost in \cref{eq:minimaxregretOtO} due to the budget constraint decreases with $\alpha$ unsurprisingly. On the other hand, low values of $\alpha$ strengthen the budget constraint. In consequence, the algorithm behaves closer to \alglcb, and the guarantee against the logging policy is improved. In fact, for $\alpha=0$, the budget is never strictly positive, and the algorithm reduces to \alglcb. In our experiments, setting $\alpha\in [0.1,1]$ gave the best results.

\end{remark}

%% file: comparing.tex
\section{Regret Analysis of UCB and LCB}\label{sec:intersetanalysis}

In this section, we systematically investigate the performance of UCB and LCB as promised. The goal here is to paint a picture as comprehensive as possible over the whole spectrum of offline-to-online learning and different compositions of offline data, which may be of interest on its own. The first step will be to focus on the minimax regret against an optimal arm. Then, we will study the two approaches'  regret against the logging policy.

Before presenting the precise statements of the results, we provide a summary with tables and figures for illustration. The first line of results presented concerns the pseudo regret against optimality and is summarized in \cref{fig:pseudoregret}. 
The first and second rows of the table correspond to UCB and LCB, respectively. The columns are labeled with a specific horizon. Here, the choices $T=1$ and $T\gg m$ correspond to short and long horizons, respectively. That horizon $T=m$ corresponds to the case when the amount of online collected data matches the size of the offline data; this is the time when we expect the online data to make some difference for the first time.

First, it is shown that \algucb\ achieves minimax optimality (up to poly-logarithmic factors) for all horizons $T\geq 0$, with a more refined dependence on the composition of the offline data compared to prior results in the literature. As for \alglcb, it is also minimax optimal for $T=1$, but this is no longer the case as $T$ increases. It would be tempting to stop here and simply accept that \algucb, being minimax optimal over the whole range of $T$, is a good algorithm for offline-to-online setting. However, the second line of results paints a different picture.

\begin{table}[bth] 
\begin{center}
\renewcommand{\arraystretch}{1.4}
\begin{tabular}{>{\centering\arraybackslash}m{3cm} | |>{\centering\arraybackslash}m{3cm} |>{\centering\arraybackslash}m{3cm} |>{\centering\arraybackslash}m{3cm} }
\toprule
$T$ & $T = 1$ & $T = m$ & $T \gg m$ \\
 \hline\hline
 \addlinespace

$\mathcal{R}_{\text{UCB}}(T)$ & $\sqrt{\frac{1}{\min_i m_i}}$ & $m\sqrt{\frac{K}{\sum_i m_i}}$ & $\sqrt{KT}$ \\
\addlinespace
\hline
\addlinespace

$\mathcal{R}_{\text{LCB}}(T)$ & $\sqrt{\frac{1}{\min_i m_i}}$ & $m\sqrt{\frac{1}{\min_i m_i}}$ & $T\sqrt{\frac{1}{\min_i m_i}}$ \\
\bottomrule
\end{tabular}
\vspace{0.2cm}
\caption{Evolution of the pseudo regret of \alglcb\ and \algucb\ as $T$ grows (matching upper and lower bounds-ignoring poly log terms-are displayed, exact expressions in the lemmas).}
\label{fig:pseudoregret}
\end{center}
\end{table}

\cite{Xiao2021OnTO} showed that the minimax criterion was not enough to distinguish \algucb{} and \alglcb, and suggested an additional criterion. In offline learning, despite the similar minimax guarantees, \alglcb{} is often favored over \algucb. There is an intuitive explanation for this: even when there is only a single iteration, \algucb\ will explore if a single arm does not have any sample, and the gathered knowledge has no chance of being exploited. \alglcb{} does the opposite in offline learning and sticks to what the offline data indicates as a ``good enough'' arm. Intuition suggests that if offline data are focused on good arms, then exploration may affect the performance negatively, whereas \alglcb{} exploits the information immediately. Also, in the literature, pessimistic algorithms have been shown to be competitive against the logging policy in offline learning. These suggest that competing against the data generating policy, i.e., the logging policy, can be another reasonable criterion for offline-to-online learning. This is what we set out to do in this second line of results, summarized in \cref{fig:regretlogging}.
 First, we observe that at $T=1$, which corresponds to offline learning, LCB has a lower regret whenever offline data is not perfectly balanced. The gap is greater when offline data are completely concentrated on one arm, i.e., $m_i=m$ for some $i \in [K]$. As $T$ increases, the gap between the two guarantees for the two algorithms decreases slowly. The point at which it closes depends on the repartition of offline data.

\begin{table}[htb]
\begin{center}
\renewcommand{\arraystretch}{1.4}
\begin{tabular}{>{\centering\arraybackslash}m{1.5cm} | |>{\centering\arraybackslash}m{1.5cm}|>{\centering\arraybackslash}m{1.5cm} |>{\centering\arraybackslash}m{3.cm}|>{\centering\arraybackslash}m{2cm}  |>{\centering\arraybackslash}m{1.5cm}|>{\centering\arraybackslash}m{1.5cm} }
 \toprule
 $T$ & \multicolumn{2}{c|}{$1$}&\multicolumn{2}{c|}{$T=m$} &\multicolumn{2}{c}{$T\gg m$ } \\ 
  \cline{2-7}
    & LB & UB  &LB & UB&LB & UB\\
     \hline
     \hline
\addlinespace
 $\mathcal{R}^{\text{log}}_\algucb(T)$ & \multicolumn{2}{c|}{$\sqrt{\frac{1}{\min_i m_i}}$} & {\small $\sum_{i=1}^K\left(\frac{m}{K}-m_i\right)\rho_i$}&$\sqrt{KT}$& $0$&$\sqrt{KT}$ \\ 
 \addlinespace
  \hline
\addlinespace
 $\mathcal{R}^{\text{log}}_\alglcb(T)$ & $\frac{\sqrt{m_2}}{m}$&$\frac{\sum_{i}\sqrt{m_i}}{m} $ &  $\sqrt{m_2}$ & $\sum_{i=1}^m\sqrt{m_i}$&$T\frac{\sqrt{m_2}}{m}$&$T\frac{\sum_{i}\sqrt{m_i}}{m} $\\ 
 \addlinespace
 \hline
\end{tabular}
\end{center}
\vspace{0.2cm}
\caption{Evolution of the regrets against the logging policy as $T$ grows (ignoring poly log terms, exact expressions in the Lemmas), assuming wlog $m_1\geq m_2\geq \ldots\geq m_K$, and with $\rho_i=\left[\sqrt{\frac{1}{m_i+\frac{m}{K}}}-\sqrt{\frac{1}{m_1+\frac{m}{K}}}\right]$.}
\label{fig:regretlogging}
\end{table}

One lesson from these two lines of results is that if $T\gg m$, then \algucb\ dominates \alglcb\ in both objectives. It is also the case when offline data are uniformly spread over the arms, i.e., $m_i=\frac{m}{K}$ for all $i \in [K]$. However, when the horizon is short ($m\gg T$) and offline data do not cover the arms uniformly, then \alglcb\ dominates \algucb in both objectives. Thus, neither algorithm dominates the other for all horizons and all possible data coverage. The algorithm we introduced, \algoname, finds a trade-off between the two strategies.

\begin{figure}[htb]
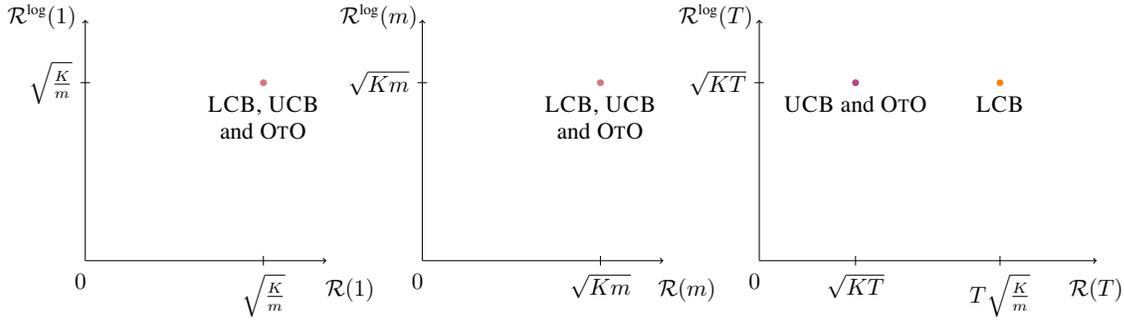

\begin{center}
  \includestandalone[scale=0.8]{plots/pareto_same_sample}
  \caption{Evolution of the two regret measures when offline samples are uniformly spread between the arms, i.e., $m_i=\frac{m}{K}$ for all $ I \in [K]$. The best algorithms are the ones closest to $(0,0)$. From left to right, the horizon is $T=1$, $T=m$ and $T\gg m. $ In the last plot, we assume $T\gg m$. The \algoname\ in the plot uses $\alpha=1$.}
\label{fig:tradeoffucblcbbanlanced}
\end{center}
\end{figure}

\begin{figure}[htb]
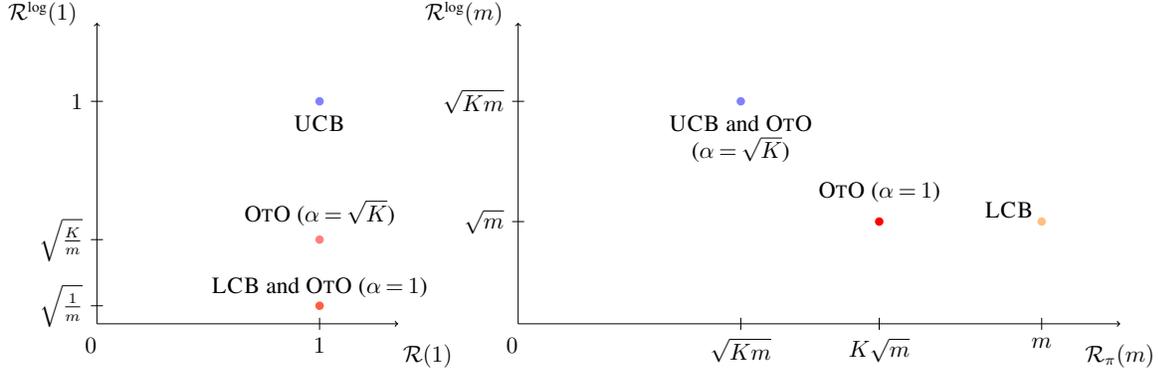

\begin{center}
  \includestandalone[scale=0.8]{plots/pareto_one_sample}
  \caption{Evolution of the two regrets when all offline samples are concentrated on two arms, i.e. $m_1=m_2=\frac{m}{2K}$. From left to right, the horizon is $T=1$ and $T=m$. For readability purposes, we do not plot $T\gg m$, as the relative behavior of the algorithms for that horizon would be quite similar to the one at $T=m$, but with an even larger ratio between the horizontal and vertical axis. }
\label{fig:tradeoffucblcbconcentrated}
\end{center}
\end{figure}

\cref{fig:tradeoffucblcbbanlanced,fig:tradeoffucblcbconcentrated} illustrate these points visually. In the figures, we plot the upper bound on the regret (ignoring poly-log terms) of all three algorithms for each specific composition of the offline arm counts and varying values of $T$. For the chosen offline arm counts, the upper bounds on \algucb\ and \alglcb\ are tight up to poly-log terms. 
In \cref{fig:tradeoffucblcbbanlanced}, we see the evolution of the regret for perfectly balanced datasets. Here, \algucb\ dominates \alglcb\ for all horizons, and \algoname\ behaves like \algucb.
In \cref{fig:tradeoffucblcbconcentrated}, we see the evolution of the regret for highly skewed offline datasets: all the samples are concentrated on the first two arms. 
Here, the picture is somewhat more complicated: \algucb\ dominates in the regret against optimality, whereas \alglcb\ dominates in the regret with respect to the logging policy. We see that \algoname finds a trade-off between the two, and the trade-off point depends on the value of the parameter $\alpha$ chosen. With $\alpha=\sqrt{K}$, for example, \algoname\ improves upon the regret against the logging policy of \algucb without downgrading (up to a multiplicative constant that does not appear on the plot) the regret against optimality. With $\alpha=1$, \algoname\ improves on the regret against optimality of \alglcb, without downgrading (again up to a multiplicative constant) the regret against the logging policy.

The rest of the section is dedicated to stating the results summarized in \cref{fig:pseudoregret,fig:regretlogging} precisely.

\subsection{Minimax regret}
We will start by proving a lower bound on the minimax regret of any algorithm for offline-to-online learning. We also derive a matching upper bound given by \algucb. 

\begin{theorem}[Lower bound on the minimax regret of any algorithm for offline-to-online learning]\label{prop:lowerboundminimax}
For any $T\geq 1$ and for any algorithm $\mathcal{A}$, we have
    \begin{equation*}
 \mathcal{R}_{\mathcal{A}}(T)\geq \frac{1}{31} T\sqrt{\max_{J\subseteq [K]}\frac{|J|}{T-1+\sum_{j\in J} m_j}}.
\end{equation*}
\end{theorem}
\textit{Sketch of Proof}: The proof is a refined application of classical information-theoretic lower bounds. The main technical challenge lies in defining an appropriate threshold, $\Delta$, which determines when arms cannot be reliably identified as suboptimal. This threshold is given by:  
\[
\Delta = \sqrt{\max_{J \subseteq [K]} \frac{|J|}{2(T - 1 + \sum_{j \in J} m_j)}}.
\]  Details can be found in \cref{sec:omittedproofs}.

The above bound may be hard to interpret. To get a sense of the behavior of the lower bound, we look at two special cases. Assume that $m_1\geq \ldots \geq m_K$. Then letting $J=\left\{m_2,\ldots, m_K\right\}$, we get
\begin{equation*}
 \mathcal{R}_{\mathcal{A}}(T)\geq \frac{1}{31} T\sqrt{\frac{(K-1)}{T-1+m-\max_i m_i}},
\end{equation*}
which recovers the $\Omega(\sqrt{TK})$ lower bound for $T$ large. Now, when $J=\left\{ m_K\right\}$, we get
\begin{equation*}
\mathcal{R}_{\mathcal{A}}(T)\geq \frac{1}{31} T\sqrt{\frac{1}{T-1+\min_i m_i}},
\end{equation*}
 which tells us that for $T$ small, the minimum count in the offline data sets a lower limit on the regret. 

We now study the regret of \algucb\ and \alglcb. We start with \algucb, showing that it is minimax optimal. We also provide an instance-dependent bound on the regret.

\begin{theorem}[\algucb's upper bound on the minimax regret]\label{prop:regretminimaxucb}
    For any $T\geq 1$ and any $\theta\in \Theta$, with probability at least $1-2T^2\delta$,
    \[
R_\algucb(T)\leq \sum_{i=1}^K \Delta_i\left( \frac{2}{\Delta_i^2} \log(K/\delta)-m_i\right)_++\sum_{i=1}^K \Delta_i.
    \]
    Also, we have the following instance-independent bound: 
    \[
\mathcal{R}_{\algucb}(T)\leq \min\left( \max_{J\subseteq [K]}2T\sqrt{\frac{2|J|\log(K/\delta)}{T+\sum_{j\in J}m_j}}+|J|;T\sqrt{\frac{2\log(K/\delta)}{\min_i m_i}}\right)+2T^2\delta.
    \]
 \end{theorem}
\textit{Sketch of Proof}: The first bound is obtained through the usual techniques of upper bounding the number of pulls from each arm. The second approach demands more intricate algebraic computations. It entails separately bounding the regret for two categories of arms: those whose gap exceeds a "detection threshold" and those whose gap falls below it. The proof can be found in \cref{sec:omittedproofs}. \hfill \(\Box\)

Theorems~\ref{prop:lowerboundminimax} and \ref{prop:regretminimaxucb} illustrate the difficulty of offline-to-online learning measured by the minimax regret and show how the difficulty depends on the composition of offline data. The derived bounds align with those obtained by \cite{pmlr-v235-cheung24a}, although the proof techniques differ. These differing techniques result in distinct implicit expressions for the bounds. However, it can be shown that the solution to the linear program (LP) in \cite{pmlr-v235-cheung24a} and the inverse of the solution to our maximization problem are within multiplicative constants of each other.

When it comes to \alglcb, the following result shows that the algorithm fully depends on the quality of the offline data. This is perhaps unsurprising, as the algorithm has no built-in exploration: its knowledge of the arms may not improve with the online interactions. 

\begin{proposition}[Minimax regret of \alglcb]\label{prop:minimaxregretlcb}
    For $T\geq 1$, we have
    \[
   \min\left(0.07 T, 0.15 T\sqrt{\frac{1}{\min_i m_i}}\right)\leq  \mathcal{R}_\alglcb(T) \leq T\sqrt{\frac{2\log(K/\delta)}{\min_i m_i}}+ 2T^2\delta.
    \]
\end{proposition}
\textit{Sketch of proof}: For the lower bound, we construct an instance where the optimal arm is the arm with the minimum count in offline samples, and all other arms have deterministic rewards. By the structure of \alglcb, if the optimal arm is not picked in the first iteration, it will never be picked. We show that this happens with a constant probability with a proper choice of distribution parameters. The upper bound is derived by bounding the gap between the mean of the chosen arm and $\mu_*$. We defer the technical steps of the proof to \cref{sec:omittedproofs}. \hfill \(\Box\)

\subsection{Regret against the logging policy}

In this section, we present the results shown in Table~\ref{fig:regretlogging}. All proofs are in Section~\ref{sec:omittedproofs}. 

\begin{proposition}[\algucb's regret against the logging policy for $T=1$]\label{prop:loginucbT=1}
We have
\[
0.07\min \left(1;\sqrt{\frac{1}{\min_i m_i}}\right)\leq \mathcal{R}^{\text{log}}_\textsc{UCB}(1) \leq \sqrt{\frac{2\log(\frac{K}{\delta})}{\min_i m_i}}+2\delta.
\]
\end{proposition}

\textit{Sketch of Proof}: For the lower bound, we construct an instance in which all arms have the same mean, except the arm with the minimal count, which has a slightly lower mean. With a proper choice of distribution parameters, \algucb\ picks the worst arm with a constant probability. The upper bound is a consequence of the upper bound on the minimax regret obtained in \cref{prop:regretminimaxucb}. \hfill \( \Box\)

\begin{proposition}[\algucb's regret against the logging policy for general $T$]\label{prop:lowerboundUCBanyT}
For any $T>0$, $\frac{T}{K}\in \mathbb{N}$, we have
    \[\mathcal{R}^{\text{log}}_{\textsc{UCB}}(T)\geq T\sum_{i=1}^K\left(\frac{1}{K}-\frac{m_i}{m}\right)\left[\sqrt{\frac{1}{2(m_i+\frac{T}{K})}}-\sqrt{\frac{1}{2(\max_{j\in [K]}m_j+\frac{T}{K})}}\right],\]
and
\[
\mathcal{R}^{\text{log}}_{\textsc{UCB}}(T)\leq \mathcal{R}_{\textsc{UCB}}(T).
\]
\end{proposition}

\textit{Sketch of Proof}: The upper bound is a consequence of the definitions, as $\mu_*\geq \mu_0$ always holds. For the lower bound, we construct an instance where the mean of each arm decreases with the number of offline samples for that arm, and all rewards are deterministic. Then, using the property that \algucb matches the upper bounds of arms (up to rounding effects, roughly speaking), we show that for a proper choice of mean parameters, each arm is sampled at least $\frac{T}{K}$ times in the online phase.  \hfill \( \Box\)

The above bound may be hard to interpret. It is easier to interpret when instantiated for extreme values of offline arm counts. For instance, if all of the offline samples come from a single arm, i.e., $m_1=m$ and $m_i=0$ for any $i>1$, we obtain
    \[\mathcal{R}^{\text{log}}_{\textsc{UCB}}(T)\geq T\frac{K-1}{K}\left[\sqrt{\frac{K}{2T}}-\sqrt{\frac{1}{2(m+\frac{T}{K})}}\right].\]
For any $T\leq K m$, this entails
    \[\mathcal{R}^{\text{log}}_{\textsc{UCB}}(T)\geq \frac{1}{10}\sqrt{KT}.\]
Similarly, if we have balanced samples for half of the arms ($m_i=\frac{2m}{K}$, for each $i \leq \frac{K}{2}$, and $m_i=0$ for $i\geq \frac{K}{2}$), then, for any $T\leq m$, we have
        $\mathcal{R}^{\text{log}}_{\textsc{UCB}}(T)\geq \frac{1}{10}\sqrt{KT}.$
On the other hand, if $m_i=\frac{m}{K}$ for all $i \in [K]$, we get $\mathcal{R}^{\text{log}}_{\textsc{UCB}}(T)\geq 0$, which may not be a strong lower bound, but it aligns with the well-known observation that a uniform offline dataset is favorable for \algucb's theoretical guarantee. 

\begin{remark}
    The upper bound obtained here does not match the lower bound for all offline count repartitions. For instance, in the case where the offline samples are uniformly spread, the former is much larger than the latter. It remains an open question to find matching bounds in all regimes. 
\end{remark}

\begin{proposition}[ \alglcb's regret against the logging policy for general $T$]\label{prop:regretlogginglcb}
We have
\[
\mathcal{R}^{\text{log}}_\textsc{LCB}(T) \leq T\frac{\sum_i \sqrt{m_i}}{\sum_i m_i}\sqrt{2\log\left(\frac{K}{\delta}\right)}+2T^2\delta.
\]
Moreover, assuming  $m_1\geq m_2\geq \ldots\geq m_K$,
\[
\mathcal{R}^{\text{log}}_\textsc{LCB}(T)\geq 0.15 T\frac{\sqrt{m_2}}{m}.
\]
\end{proposition}

It is again informative to instantiate the above bound in extreme regimes of the offline count repartition. If $m_1=m$ and $m_i=0$ for any $i>1$, we obtain
    \[\mathcal{R}^{\text{log}}_\textsc{LCB}(T) \leq T\sqrt{\frac{2\log\left(\frac{K}{\delta}\right)}{m}}+2T^2\delta.\]  
On the other hand, if $m_i=\frac{m}{K}$ for all $i \in [K]$, we get:      \[\mathcal{R}^{\text{log}}_\textsc{LCB}(T) \leq T\sqrt{\frac{2K\log\left(\frac{K}{\delta}\right)}{m}}+2T^2\delta.\]

%% file: plots/pareto_same_sample.tex
\begin{tikzpicture}[xscale=0.8,yscale=0.8] 
  \draw [->,line width=0.1pt](0.,0.) -- (0.,5.); 
  \draw [->,line width=0.1pt](0.,0.) -- (5.,0.);
  
  \draw[anchor=north] (-0.1,-0.1) node { $0$};
  \draw[anchor=north] (5.5,-0.2) node { $\mathcal{R}(1)$};
  \draw[anchor=north] (-0.9,5.5) node { $\mathcal{R}^{\text{log}}(1)$};
  \draw[anchor=north] (3.7,-0.1) node { $\sqrt{\frac{K}{m}}$};
  \draw[thin] (3.7,0.1)--(3.7,-0.1);
  \draw[thin] (0.1,3.7)--(-0.1,3.7);
  \draw[anchor=east] (-0.1,3.7) node { $\sqrt{\frac{K}{m}}$};

  \fill[blue,fill opacity=0.3] (3.7,3.7) circle (2pt);
  \fill[orange,fill opacity=0.3] (3.7,3.7) circle (2pt);
  \fill[red,fill opacity=0.3] (3.7,3.7) circle (2pt);

  \draw[anchor=north] (3.7,3.6) node {\parbox{2.5cm}{\centering \textsc{LCB},  \textsc{UCB} \\ and \algoname}};

  \draw [->,line width=0.1pt](7.,0.) -- (7.,5.); 
  \draw [->,line width=0.1pt](7.,0.) -- (7.+5.,0.);
  
  \draw[anchor=north] (7.+-0.1,-0.1) node { $0$};
  \draw[anchor=north] (7.+5.5,-0.2) node { $\mathcal{R}(m)$};
  \draw[anchor=north] (7.+-0.9,5.5) node { $\mathcal{R}^{\text{log}}(m)$};
  \draw[anchor=north] (7.+3.7,-0.1) node { $\sqrt{Km}$};
  \draw[thin] (7.+3.7,0.1)--(7.+3.7,-0.1);
  \draw[thin] (7.+0.1,3.7)--(7.+-0.1,3.7);
  \draw[anchor=east] (7.+-0.1,3.7) node { $\sqrt{Km}$};

  \fill[blue,fill opacity=0.3] (7.+3.7,3.7) circle (2pt);
  \fill[orange,fill opacity=0.3] (7.+3.7,3.7) circle (2pt);
  \fill[red,fill opacity=0.3] (7.+3.7,3.7) circle (2pt);

  \draw[anchor=north] (7.+3.7,3.6) node {\parbox{2.5cm}{\centering \textsc{LCB},  \textsc{UCB}  and \algoname}};

  \draw [->,line width=0.1pt](14.,0.) -- (14.,5.); 
  \draw [->,line width=0.1pt](14.,0.) -- (14.+7.,0.);
  
  \draw[anchor=north] (14.+-0.1,-0.1) node { $0$};
  \draw[anchor=north] (14.+7,-0.2) node { $\mathcal{R}(T)$};
  \draw[anchor=north] (14.+-0.9,5.5) node { $\mathcal{R}^{\text{log}}(T)$};
  \draw[anchor=north] (14.+2.,-0.1) node { $\sqrt{K  T}$};
  \draw[anchor=north] (14.+5.,-0.1) node { $T\sqrt{  \frac{K}{m}}$};

  \draw[thin] (14.+2.,0.1)--(14.+2.,-0.1);
  \draw[thin] (14.+5.,0.1)--(14.+5.,-0.1);

  \draw[thin] (14.+0.1,3.7)--(14.+-0.1,3.7);
  \draw[anchor=east] (14.+-0.1,3.7) node { $\sqrt{K  T}$};

  \fill[blue,fill opacity=0.5] (14.+2.,3.7) circle (2pt);
  \fill[red,fill opacity=0.5] (14.+2.,3.7) circle (2pt);

  \draw[anchor=north] (14.+2.,3.6) node {  \textsc{UCB} and \algoname};

  \fill[orange] (14.+5,3.7) circle (2pt);
  \draw[anchor=north] (14.+5,3.6) node {  \textsc{LCB}};

\end{tikzpicture}

%% file: plots/pareto_one_sample.tex
\begin{tikzpicture}
  \draw [->,line width=0.1pt](0.,0.) -- (0.,5.);
  \draw [->,line width=0.1pt](0.,0.) -- (5.,0.);
  
  \draw[anchor=north] (-0.1,-0.1) node { $0$};
  \draw[anchor=north] (5.5,-0.2) node { $\mathcal{R}(1)$};
    \draw[anchor=north] (-0.9,5.5) node { $\mathcal{R}^{\text{log}}(1)$};
 \draw[anchor=north] (3.7,-0.1) node { $1$};
 \draw[thin] (3.7,0.1)--(3.7,-0.1);
 \draw[thin] (0.1,3.7)--(-0.1,3.7);
 \draw[anchor=east] (-0.1,3.7) node { $1$};
  \draw[anchor=east] (-0.1,0.3) node { $\sqrt{\frac{1}{m}}$};
   \draw[thin] (-0.1,0.3)--(0.1,0.3);
  \draw[anchor=east] (-0.1,1.4) node { $\sqrt{\frac{K}{m}}$};
   \draw[thin] (-0.1,1.4)--(0.1,1.4);
  \fill[orange,fill opacity=0.5] (3.7,0.3) circle (2pt);
    \fill[red,fill opacity=0.5] (3.7,0.3) circle (2pt);
 \fill[red,fill opacity=0.5] (3.7,1.4) circle (2pt);

  \draw[anchor=south] (3.7,1.5) node {  \algoname ($\alpha=\sqrt{K}$)};
  
    \draw[anchor=south] (3.7,0.3) node { \textsc{LCB} and \algoname ($\alpha=1$)};
  \fill[blue,fill opacity=0.5] (3.7,3.7) circle (2pt);
  \draw[anchor=north] (3.7,3.6) node { \textsc{UCB}};

 \draw [->,line width=0.1pt](7.,0.) -- (7.,5.);
  \draw [->,line width=0.1pt](7.,0.) -- (7.+10.,0.);
  
  \draw[anchor=north] (7.+-0.1,-0.1) node { $0$};
  \draw[anchor=north] (7.+10,-0.2) node { $\mathcal{R}_\pi(m)$};
    \draw[anchor=north] (7.+-0.9,5.5) node { $\mathcal{R}^{\text{log}}(m)$};

 \draw[thin] (7.+3.7,0.1)--(7.+3.7,-0.1);
 \draw[anchor=north] (7.+3.7,-0.1) node { $\sqrt{Km}$};
 \draw[thin] (7.+8.7,0.1)--(7.+8.7,-0.1);
 \draw[anchor=north] (7.+8.7,-0.1) node { $m$};

  \draw[thin] (7.+6.,0.1)--(7.+6.,-0.1);
 \draw[anchor=north] (7.+6.,-0.1) node { $K\sqrt{m}$};

 \draw[anchor=east] (7.+-0.1,3.7) node { $\sqrt{Km}$};
 \draw[thin] (7.+0.1,3.7)--(7.+-0.1,3.7);

 \draw[anchor=east] (7.+-0.1,1.7) node { $\sqrt{m}$};
  \draw[thin] (7.+0.1,1.7)--(7.+-0.1,1.7);

   \fill[orange,fill opacity=0.5] (7.+8.7,1.7) circle (2pt);
    \draw[anchor=east] (7.+8.7,1.9) node { \textsc{LCB}};
       \fill[red,fill opacity=1.] (7.+6.,1.7) circle (2pt);
    \draw[anchor=south] (7.+6.,1.9) node { \algoname\ ($\alpha=1$)};
  \fill[blue,fill opacity=0.5] (7+3.7,3.7) circle (2pt);
  \draw[anchor=north] (7+3.7,3.6) node { \parbox{2.5cm}{\centering\textsc{UCB} and \algoname\  ($\alpha=\sqrt{K}$)}};

\end{tikzpicture}

%% file: experiments.tex
\section{Numerical Illustration}

The source code for reproducing all experiments is publicly available at: \url{https://github.com/FloreSentenac/offlinetoonline}.
\subsection{Comparison on Synthetic Data}
\begin{figure}[htb]
\centering
\begin{subfigure}{.48\textwidth}
  \includegraphics[width=\linewidth,keepaspectratio]{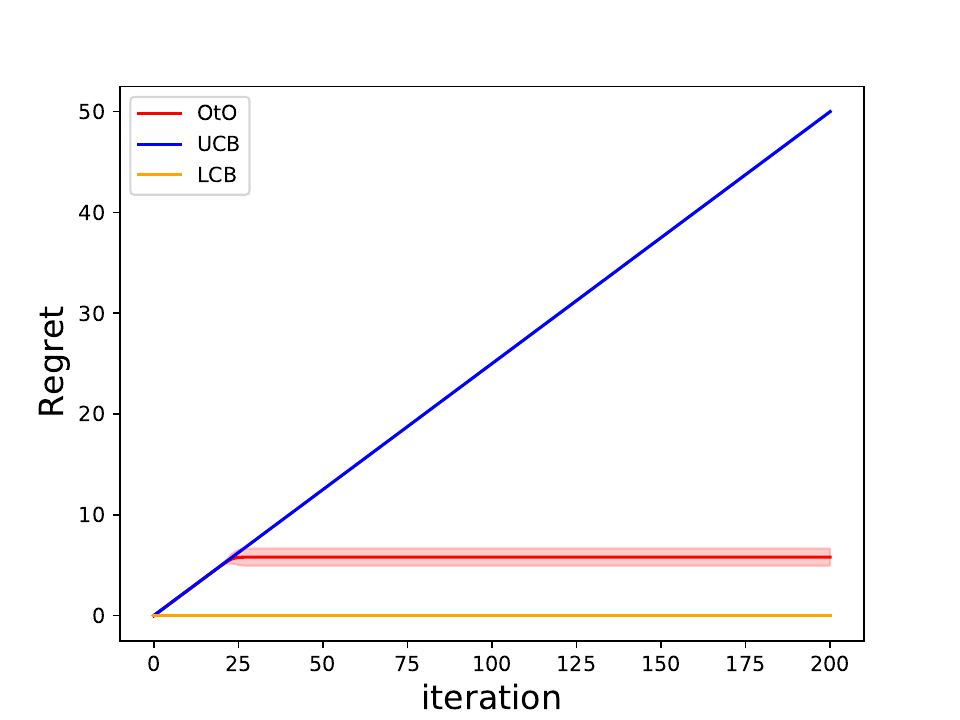}
  \caption{Instance $1$, $T=200$}
  \label{fig:smallhorizonoptsampled}
\end{subfigure}\hfill
\begin{subfigure}{.48\textwidth}
  \includegraphics[width=\linewidth]{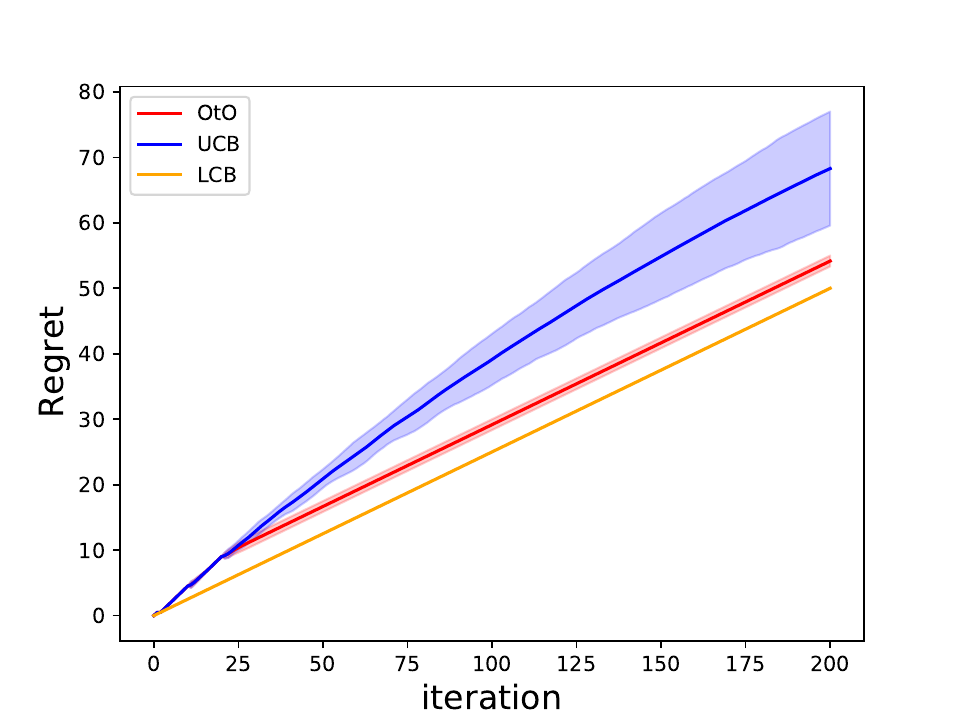}
  \caption{Instance $2$, $T=200$}
  \label{fig:smallhorizonoptnotsampled}
\end{subfigure}
\begin{subfigure}{.48\textwidth}
  \includegraphics[width=\linewidth]{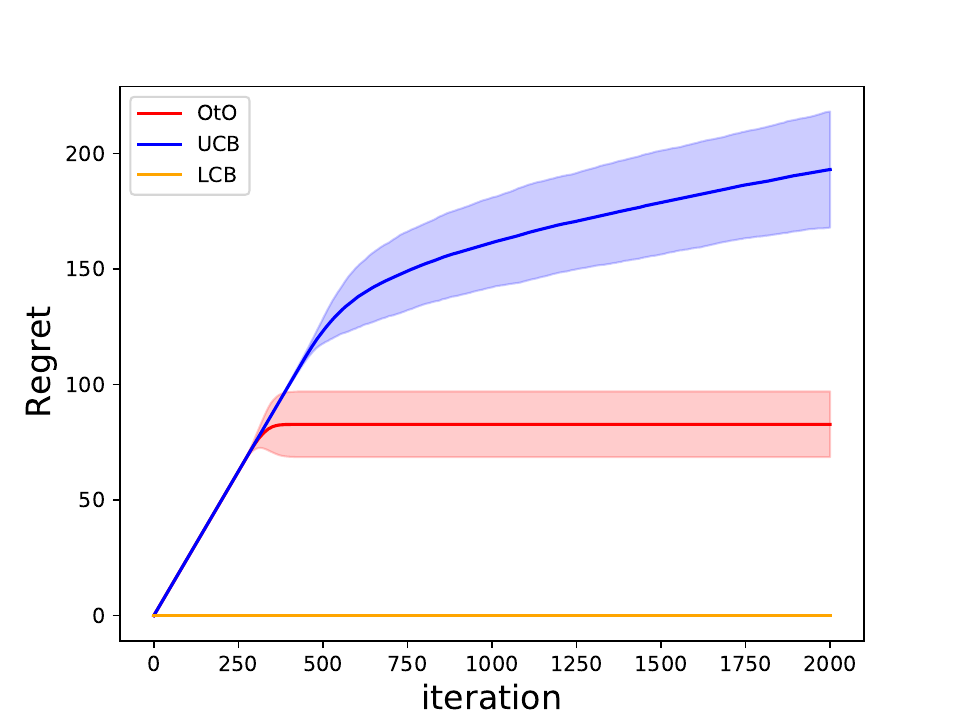}
  \caption{Instance $1$, $T=2000$}
  \label{fig:Largehorizonoptsampled}
\end{subfigure}\hfill
\begin{subfigure}{.48\textwidth}
  \includegraphics[width=\linewidth]{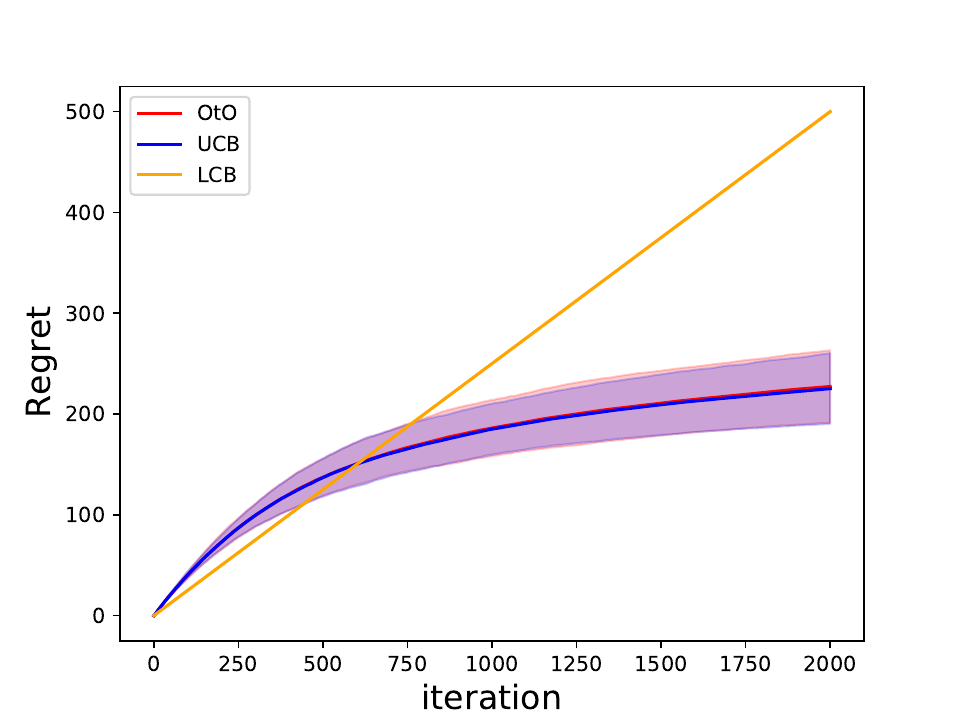}
  \caption{Instance $2$, $T=2000$}
  \label{fig:LargehorizonoptnotsampledTknown}
\end{subfigure}
\caption{Regret of the three algorithms for different instances and $T$ values, when the horizon $T$ is given to \algoname}
\label{fig:firstsetexp}
\end{figure}

We now illustrate the discussion of the previous section with some numerical experiments on synthetic data.  We consider two bandits instances in this experiment section. Both have $K=20$ arms, a total of $m=2000$ offline samples, and the first $10$ arms are each sampled $200$ times in the offline data set.  The remaining arms are not sampled at all. The reward distribution for each arm $i\in [K]$ is Bernoulli with mean $\mu_i$. In instance $1$, each arm $1\leq i\leq 10$ has mean $\mu_i=0.5$, while for each $i >10$, $\mu_i=0.25$. Note that on that instance, the logging policy is optimal. Based on the results of the previous sections, it is expected that \alglcb\ should outperform \algucb. In instance $2$, the means are the same as in instance $1$ for all arms $i<20$, while $\mu_{20}=0.75$. Note that in instance $2$, the optimal arm is not sampled in the offline dataset. Again, based on the results of the previous sections, we expect \algucb\ to outperform \alglcb\ for large horizons $T$. Also, according to the previous section, \algoname\ should perform in between \alglcb\ and \algucb\ on both instances.

\begin{figure}[htb]
\begin{centering}
\begin{subfigure}{.48\textwidth}
  \centering
  \includegraphics[width=\linewidth]{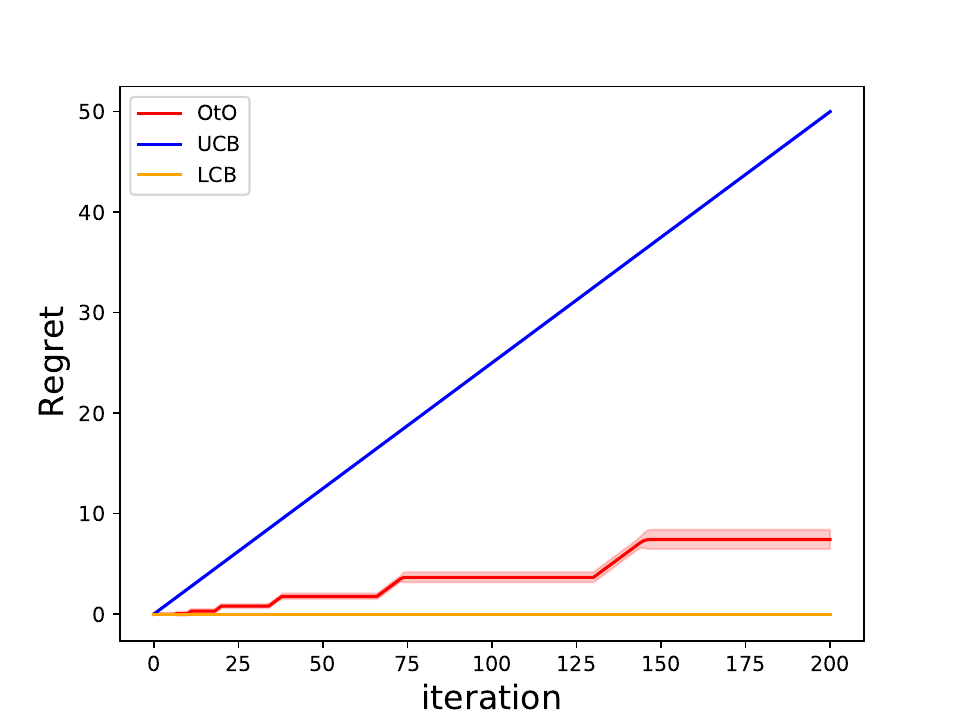}
  \caption{Instance $1$, $T=200$}
\end{subfigure}%
\begin{subfigure}{.48\textwidth}
  \centering
  \includegraphics[width=\linewidth]{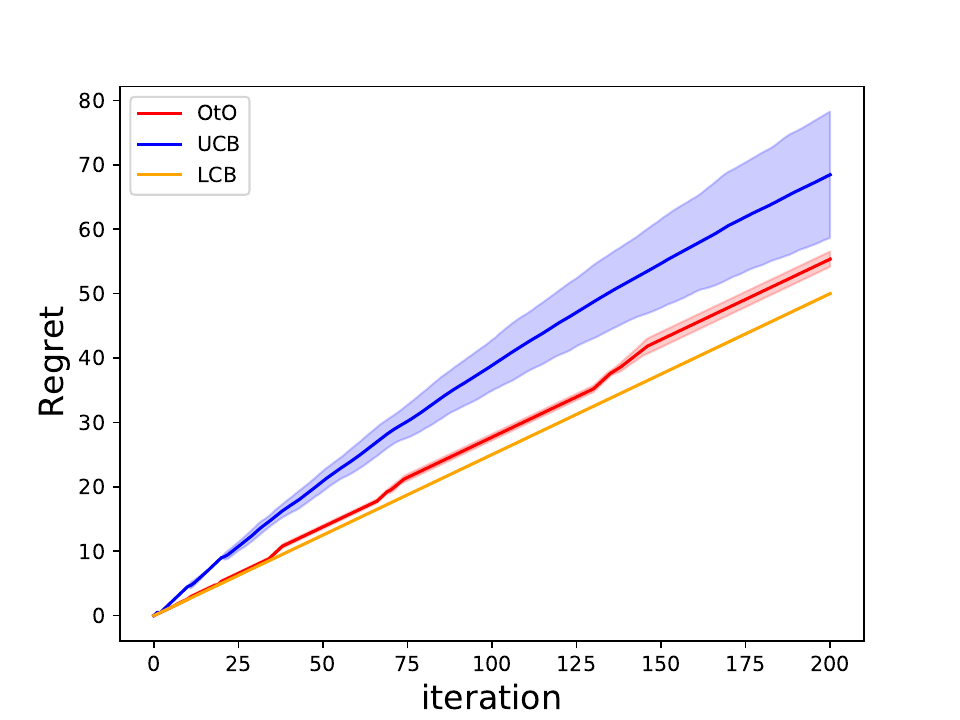}
  \caption{Instance $2$, $T=200$}
\end{subfigure}
\begin{subfigure}{.48\textwidth}
  \centering
  \includegraphics[width=\linewidth]{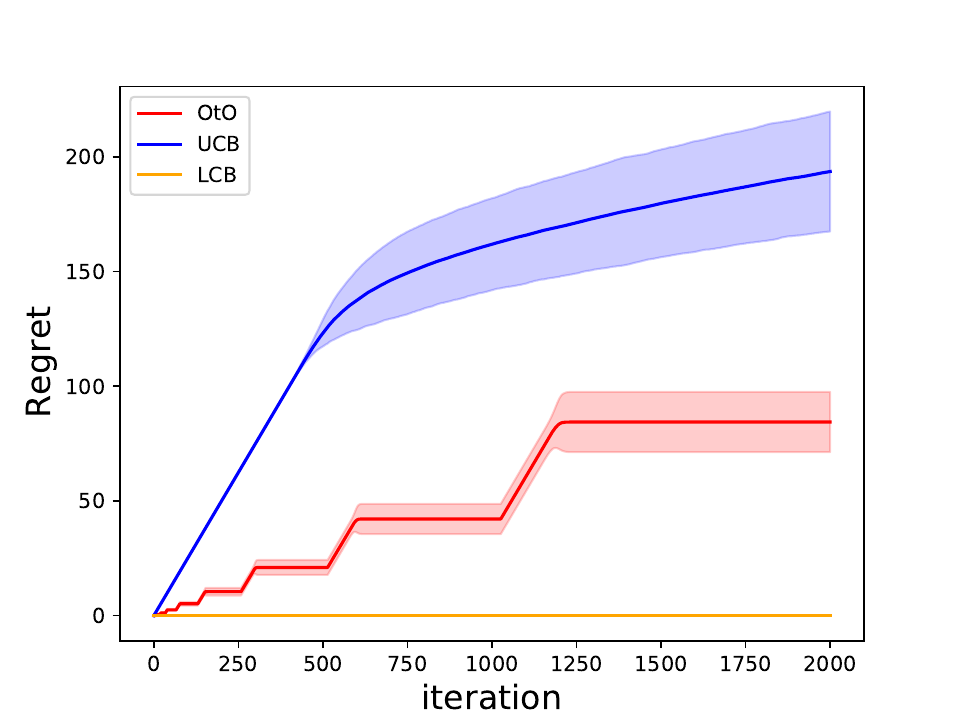}
  \caption{Instance $1$, $T=2000$}
\end{subfigure}%
\begin{subfigure}{.48\textwidth}
  \centering
  \includegraphics[width=\linewidth]{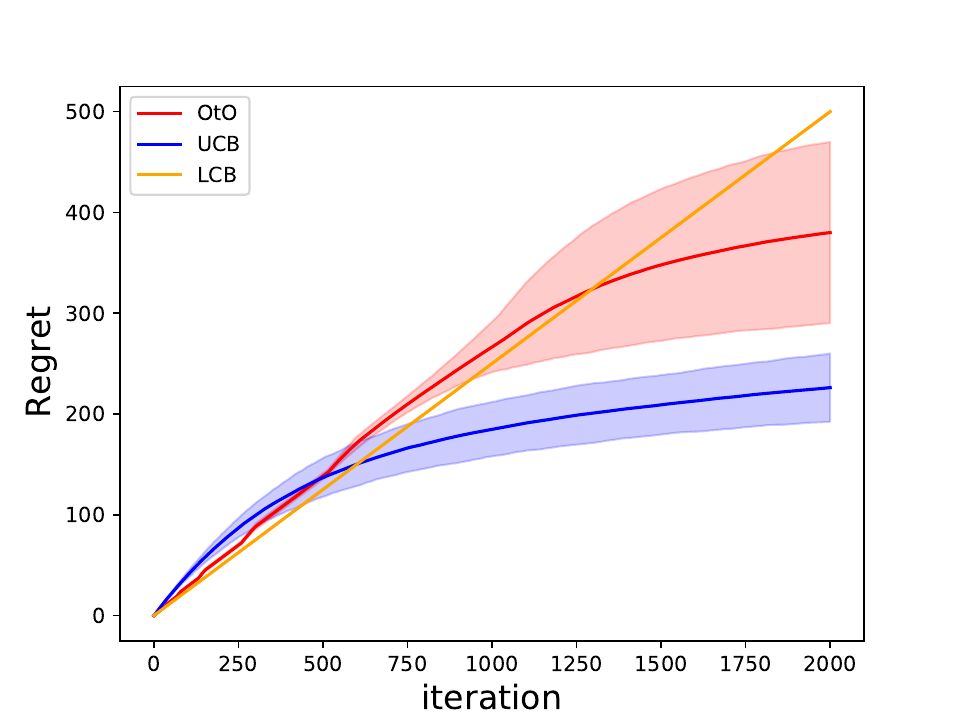}
  \caption{Instance $2$, $T=2000$}\label{fig:largehorizonTunknown}
\end{subfigure}
\caption{Regret of the three algorithms for $T=m=2000$, when the horizon $T$ is unknown.}
\label{fig:secondsetexp}
\end{centering}
\end{figure}

We compare the three algorithms, \alglcb, \algucb\, and \algoname\, on the two instances for small horizons ($T=200$) and larger horizons ($T=2000$). In a first set of experiments, the horizon $T$ is known, and we set $\delta=\frac{1}{T^2}$ for all algorithms and $\alpha=0.2$. In the second one, the horizon is unknown, and we set $\delta_t= \frac{0.01}{t^2}$ for all algorithms and $\alpha=0.6$. For each combination of instance, horizon and parameter setting, the experiment is run $200$ times. The results for the first set of experiments are displayed in \cref{fig:firstsetexp}, and for the second set in \cref{fig:secondsetexp}. The bold lines show the average regrets of each algorithm. The shaded areas represent the means plus or minus two standard deviations.

In the unknown horizon case, we opted for a slightly higher value of $\alpha$,  for which the algorithm's behavior remains comparable between the known and unknown horizon settings. Note that while the algorithm tends to perform better with carefully chosen values of $\alpha$, the bounds in \cref{thm:boundofftoon} are valid for any choice of $\alpha$. Moreover, we expect the algorithm not to be overly sensitive to the parameter, as is explored experimentally in the next section.

In the results displayed in Figure~\ref{fig:firstsetexp}, we first compare \algucb and \alglcb. We observe that \alglcb\ outperforms \algucb\ when the offline data contains the optimal arms, e.g., generated by an expert policy (\cref{fig:smallhorizonoptsampled,fig:Largehorizonoptsampled}). When the optimal arm is not sampled, \alglcb's regret grows linearly but is still lower than that of \algucb when the horizon is short because \algucb did not have enough time to explore and \alglcb made a safe choice based on the offline data (Figure~\ref{fig:smallhorizonoptnotsampled}). However, UCB performs better when the offline data does not contain the optimal arm and the horizon is long enough (Figure~\ref{fig:LargehorizonoptnotsampledTknown}). 

We note that \algoname\ smoothly interpolates between \alglcb\ and \algucb. In \cref{fig:smallhorizonoptsampled,fig:Largehorizonoptsampled}, it starts by following the footsteps of \algucb, then, as the budget is detected to be too low, it switches and sticks to \alglcb until the end of the run. In \cref{fig:smallhorizonoptnotsampled,fig:LargehorizonoptnotsampledTknown}, it also starts by following the footsteps of \algucb. Then, it exhibits two different behavior for the small and large horizons. In the case of the small horizon, \algucb\ is still in the exploratory phase, where the regret grows faster than that of \alglcb. Here, as in the two previous cases, the lack of budget causes a switch to \alglcb. On the other hand, for the larger horizon, \algucb enters the logarithmic growth phase of regret and outperforms \alglcb vastly. The budget is now always in excess, and our algorithm never switches to \alglcb.

By design, the budget ensures that \algoname\ focuses on exploration at the beginning of the experiment window. A natural question arises: does this behavior persist when the horizon is unknown? The plots in \cref{fig:secondsetexp} indicate that the behavior remains relatively similar, though with some important distinctions. First, the algorithm now explores in phases, with each pseudo-horizon update triggering a brief exploration period. Second, there are notable differences between \cref{fig:largehorizonTunknown} and \cref{fig:LargehorizonoptnotsampledTknown}. In the known horizon case (\cref{fig:LargehorizonoptnotsampledTknown}), \algoname\ never switches back to \alglcb, and there is no apparent cost to enforcing the budget. In contrast, in the unknown horizon case (\cref{fig:largehorizonTunknown}), while \algoname\ ultimately behaves like \algucb, there is an additional cost due to the budget constraint, as the algorithm reverts to the safe option, \alglcb, at shorter horizons.
This is to be expected in general: in the known horizon case, when \algucb is the better option, the cost of enforcing the budget constraint sometimes disappears. In the unknown horizon case, even for very large values of $\alpha$, an additional cost is always expected.

\subsection{Ad Click Through Rate Data}\label{sec:exprealdata}
\newcommand{\oto}{\algoname}

 While our synthetic data experiments should illustrate well the gain from using \oto, it remains to be seen whether these gains would also manifest themselves in more realistic scenarios. 
To get some insight into this, we will explore the performance of \oto 
in the context of ad recommendation. Ads recommendation
is the process of selecting and delivering advertisements personalized to internet users so as to maximize the relevance of the advertisements to the users. The main tool for developing such personalized ad recommendation systems is to use machine learning algorithms to predict the probability that the viewer of an ad will click on the ad. Given accurately predicted \emph{click-through rates} (CTRs), the system can select the ad with the highest predicted CTR. Companies working on such systems continuously collect data to improve the accuracy of the CTR prediction methods. 
Oftentimes, over the course of years, companies will develop multiple, competing CTR prediction methods. Typically, companies cannot reliably evaluate the performance of trained models without having them deployed into the systems. Indeed, the data available for evaluation rarely is a uniform sample of the true data that would appear in the real system: on top of other issues, such as temporal trends, ads collected are chosen by the system in place, and as such are selected with a bias. New methods then have to be put into production to be assessed accurately. The problem of choosing between different prediction methods in production can then be viewed as a multi-armed bandit problem, where the arms correspond to the individual prediction methods and the reward is, say, $+1$, when the ad chosen by the prediction method is clicked by a user and is zero otherwise. 

In this section we use publicly available data to emulate this situation and generate data that illustrate different scenarios. The data we use was collected by Avazu, a mobile internet advertising and performance marketing company
 and was made available in 2014 for a publicly held competition \citep{avazu-ctr-prediction}.
The data consists of over 11 million records of mobile ad display. Each record captures detailed information about an ad impression. 
 
 The experiment was conducted in two phases. First, a \textit{preparatory} phase, during which models were trained and a reward table was generated. Then, a \textit{bandit} phase followed, where we simulated the bandit-algorithm-assisted deployment of the trained models in an environment based on the reward table.  
 
\textbf{Preparatory Phase}: 
The initial step of the preparatory phase involved training a predictive model (Model 1) using a dataset of 30,000 CTR entries, uniformly sampled without replacement from the Avazu dataset, referred to as Dataset 1. 10\% of Dataset 1 was withheld from training and set aside as a test set for later use. This portion is referred to as Test Set 1. We then simulated the deployment of Model 1. In each of 30,000 iterations, the model was tasked with selecting an advertisement from 10 options presented at each iteration, choosing the one with the highest predicted click probability. The 10 options were uniformly sampled without replacement from the Avazu dataset, excluding entries already included in Dataset 1. The selected advertisements and their observed outcomes (clicked or not clicked) were recorded to construct a new dataset, referred to as Dataset 2.

This process reflects how companies deploy predictive models in practice to select advertisements for users. Dataset 2 also mirrors the type of data a company would collect, shaped by the biases inherent in the deployed model’s predictions.

Next, six new models (Models 2–7) are trained on the data generated in the simulated production environment, Dataset 2, under the assumption that the original dataset, Dataset 1, is no longer available. This reflects a realistic scenario, as companies rarely have access to datasets that are sampled unbiasedly from the true ad distribution. Furthermore, they are often obligated by regulators to delete old data.

A natural but naive approach to selecting the best model for production is to set aside a portion of Dataset 2 (10\% in our case), evaluate all models on this test data, and retain the one with the best performance. We refer to this set-aside portion as Test Set 2. However, in our scenario, model performance on Test Set 2 does not accurately reflect how the models would perform in the production environment due to the distribution mismatch. This is illustrated in \cref{fig:crossentropyloss}, where the first line shows the cross-entropy loss of all models on Test Set 2, and the second line shows the cross-entropy loss on Test Set 1, which has the same distribution as the production data. Note that, unless models are actually deployed in production, the company does not have access to the information in the second row of the table. We refer to the model with the smallest cross-entropy log-loss on Test Set 2 as the empirical best arm on the test set, which, in this specific case, is Model 3.
\begin{table}[bth] 
\begin{center}
\renewcommand{\arraystretch}{1.4}
\begin{tabular}{>{\centering\arraybackslash}m{3cm} | |>{\centering\arraybackslash}m{1.cm} |>{\centering\arraybackslash}m{1.cm} |>{\centering\arraybackslash}m{1.cm} |>{\centering\arraybackslash}m{1.cm}|>{\centering\arraybackslash}m{1.cm} |>{\centering\arraybackslash}m{1.cm} |>{\centering\arraybackslash}m{1.cm} |>{\centering\arraybackslash}m{1.cm}  }
\toprule
Model \#&  1 &  2 &  3& 4& 5& 6& 7& 8 \\
 \hline\hline
 \addlinespace

Test Set 2 & $0.666$&$0.576$ &$ 0.566$ &$0.576$ &$0.576$ &$0.572$ &$0.579$ & $0.648$\\
\addlinespace
\hline
\addlinespace

Test Set 1 &$0.394$ &$0.896$&$0.640$&$0.714$&$1.063$&$0.690$&$0.752$& $0.388$ \\
\bottomrule
\end{tabular}
\vspace{0.2cm}
\caption{Binary cross entropy loss (rounded to the third digit) for each model on the two test sets}
\label{fig:crossentropyloss}
\end{center}
\end{table}

All models trained on Dataset 2 perform worse than Model 1 on Test Set 1. To illustrate different scenarios regarding the quality of the arm generating offline data, 
we needed at least one model that outperforms Model 1. To achieve this, we trained another model, Model 8, on Dataset 1, similar to Model 1. Although training on the original dataset does not align with the constraints companies typically face, it provides a reliable method for constructing a strong predictor. In practice, companies do manage to improve upon their existing models, especially since our baseline model has the unfair advantage of being trained on perfect data. Regardless of how it was trained, Model 8 is evaluated on Test Set 2. While it does outperform Model 1, it is still outperformed by Models 2–7.

The final step of the preparatory phase involves generating a reward table, which serves as the foundation for the reward generation process in the next phase. This table is constructed by evaluating the models iteratively over multiple events. For each event, all models are presented with a set of 10 advertisements, sampled uniformly at random without replacement from the unused ads. Each model selects the advertisement with the highest predicted Click-Through Rate (CTR). A reward of 1 is recorded if the selected advertisement is clicked, and 0 otherwise. The reward table entry $(i,j)$ represents the reward for model j during presentation event i, with 1 indicating a click and 0 indicating no click. A total of $300,000$ events are collected.
Details on the models used can be found in \cref{app:simulationrealdata}.

\textbf{Bandit Phase}: This phase is the core offline-to-online experiment, simulating the transition to production for new models after the baseline model (Model 1) has been in operation for some time. In this experiment, each prediction model is treated as an arm. Dataset 2 is the offline dataset. Generating a reward for arm $j$ is done by randomly selecting a row $i$ from the reward table and outputting the entry $(i,j)$. The goal of the company is to identify the best model using a combination of offline data and online experiments. 

We consider two scenarios regarding the quality of the running model (i.e., the performance of the logging policy): \textit{Setting 1}, where one of the six new models performs better than the running model (we include Models 2-6 and 8), and \textit{Setting 2}, where all new models perform worse than the running model (we include Models 2-7).

We evaluate three strategies—\alglcb, \algucb, and \algoname—for two horizon lengths, $T = 3,000$ and $T = 300,000$, reporting results for \algoname with several values of $\alpha$. For each value of $T$, we ran the experiment 200 times. For comparison, we include two benchmark arms in the results: (1) the best empirical arm in hindsight, which is the arm identified by evaluating each arm over 200 runs and selecting the one with the highest empirical mean, and (2) the arm with the lowest binary cross-entropy loss on Test Set 2.

\begin{figure}[ht]
    \centering
    \begin{subfigure}[b]{0.8\textwidth}
        \centering
        \includegraphics[width=\textwidth]{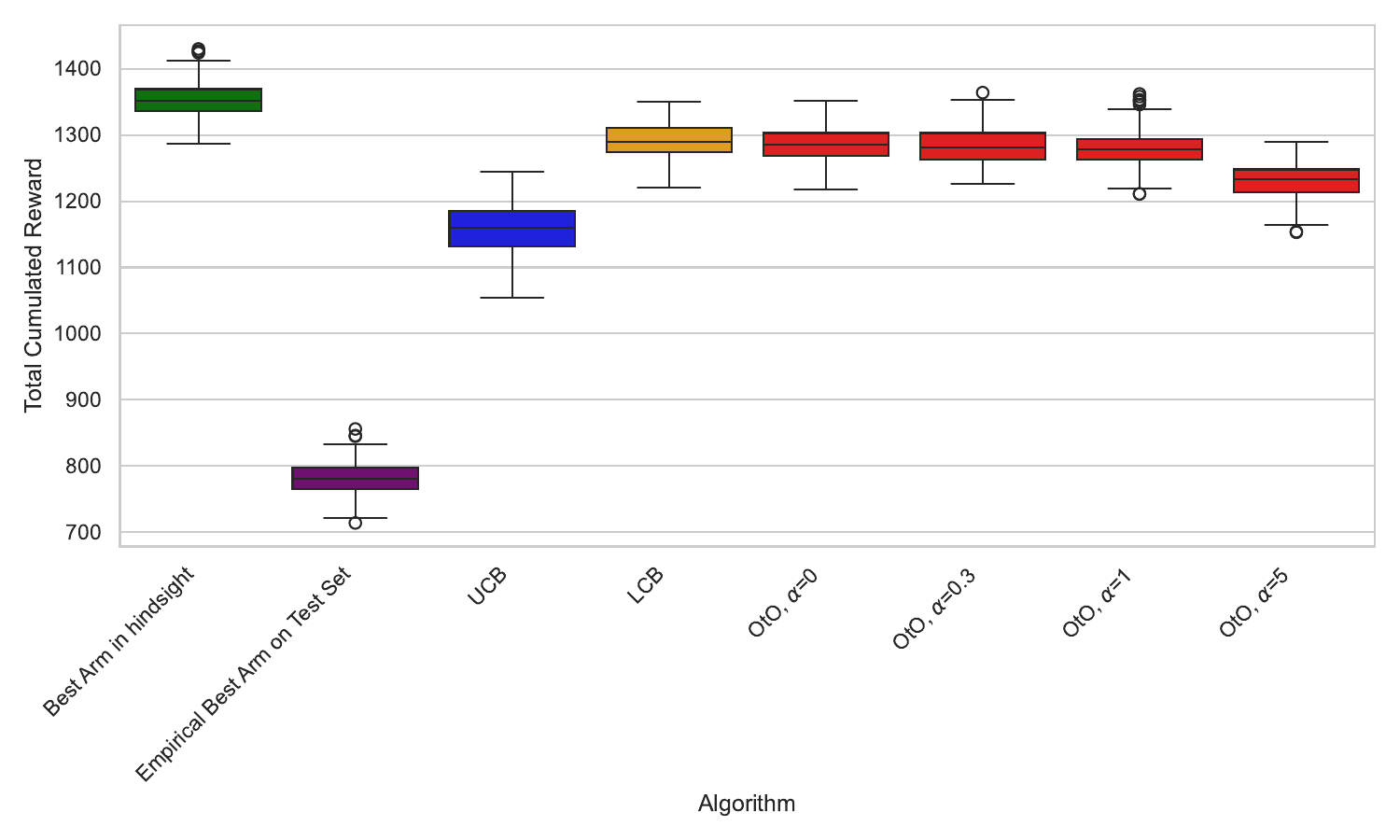} 
        \caption{Short horizon, $T=3,000$}
        \label{fig:fiboxplotonegoodarmshorthorizon}
    \end{subfigure}
    \vspace{1cm} 
    \begin{subfigure}[b]{0.8\textwidth}
        \centering
        \includegraphics[width=\textwidth]{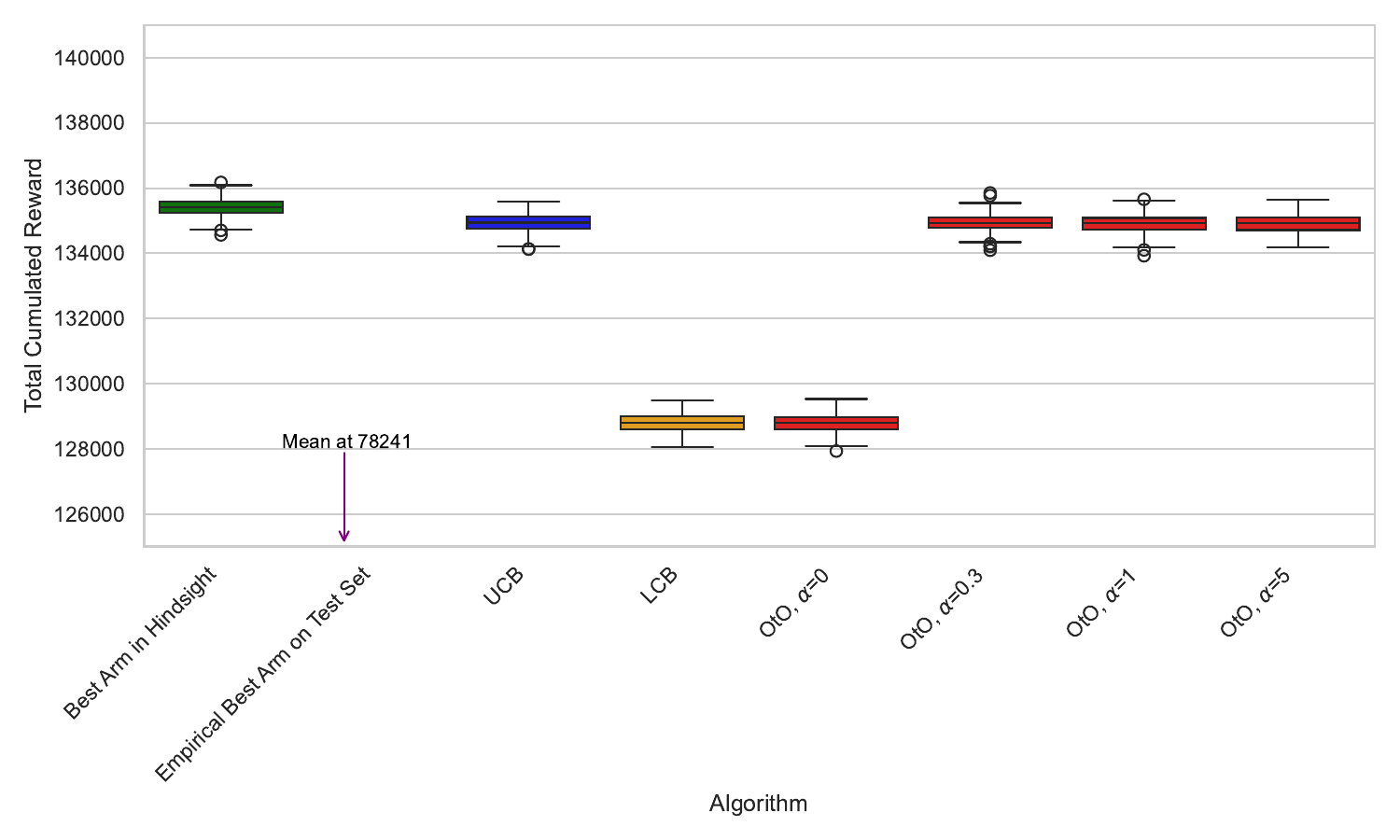} 
        \caption{Long horizon, $T=300,000$}
    \end{subfigure}
    \caption{The total cumulated reward by each algorithm in Setting 1}
    \label{fig:summaryexprealdatagoodarm}
\end{figure}

\begin{figure}[ht]
    \centering
    \begin{subfigure}[b]{0.8\textwidth}
        \centering
        \includegraphics[width=\textwidth]{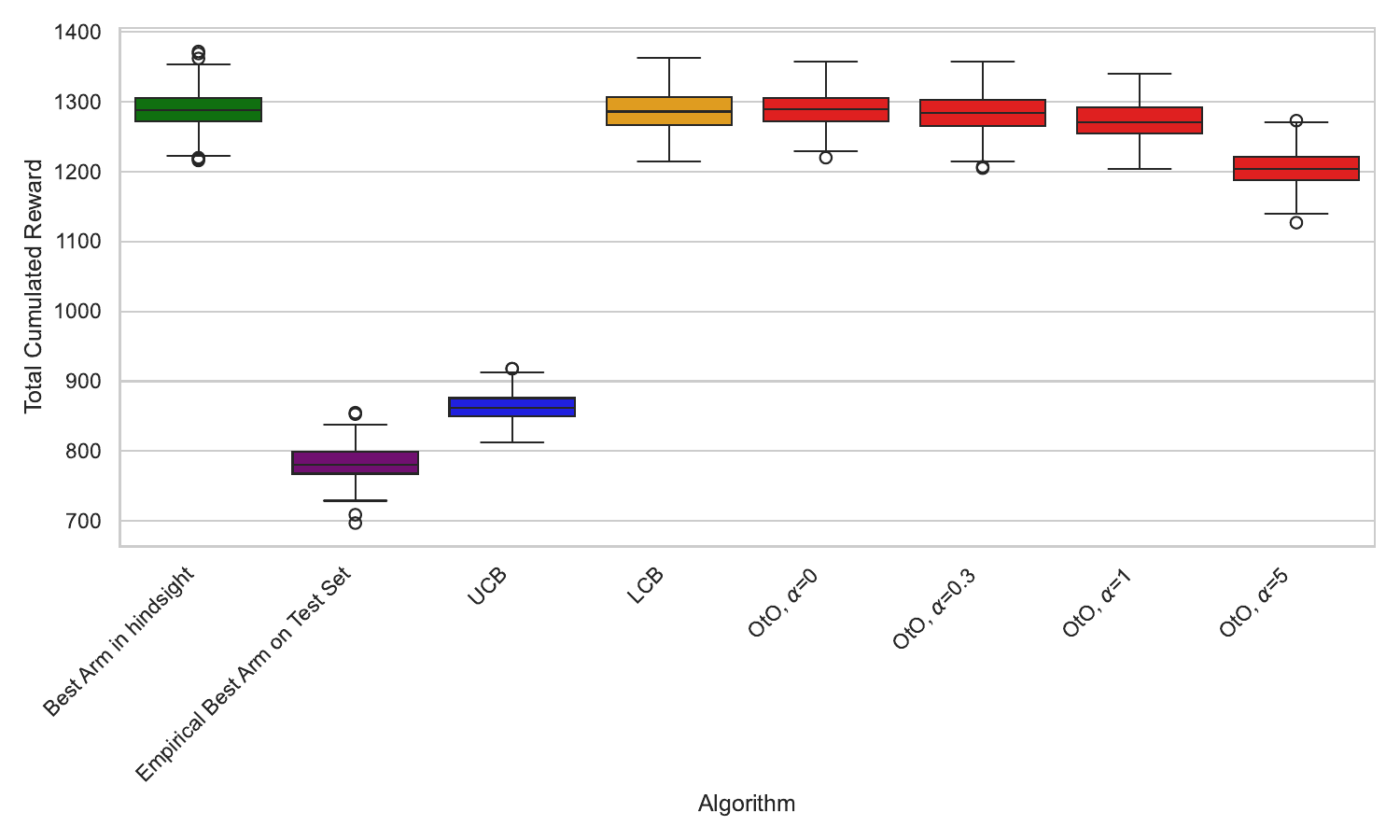} 
        \caption{Short horizon, $T=3,000$}
        \label{fig:fiboxplotbadarmsonlyhorthorizon}
    \end{subfigure}
    \vspace{1cm} 
    \begin{subfigure}[b]{0.8\textwidth}
        \centering
        \includegraphics[width=\textwidth]{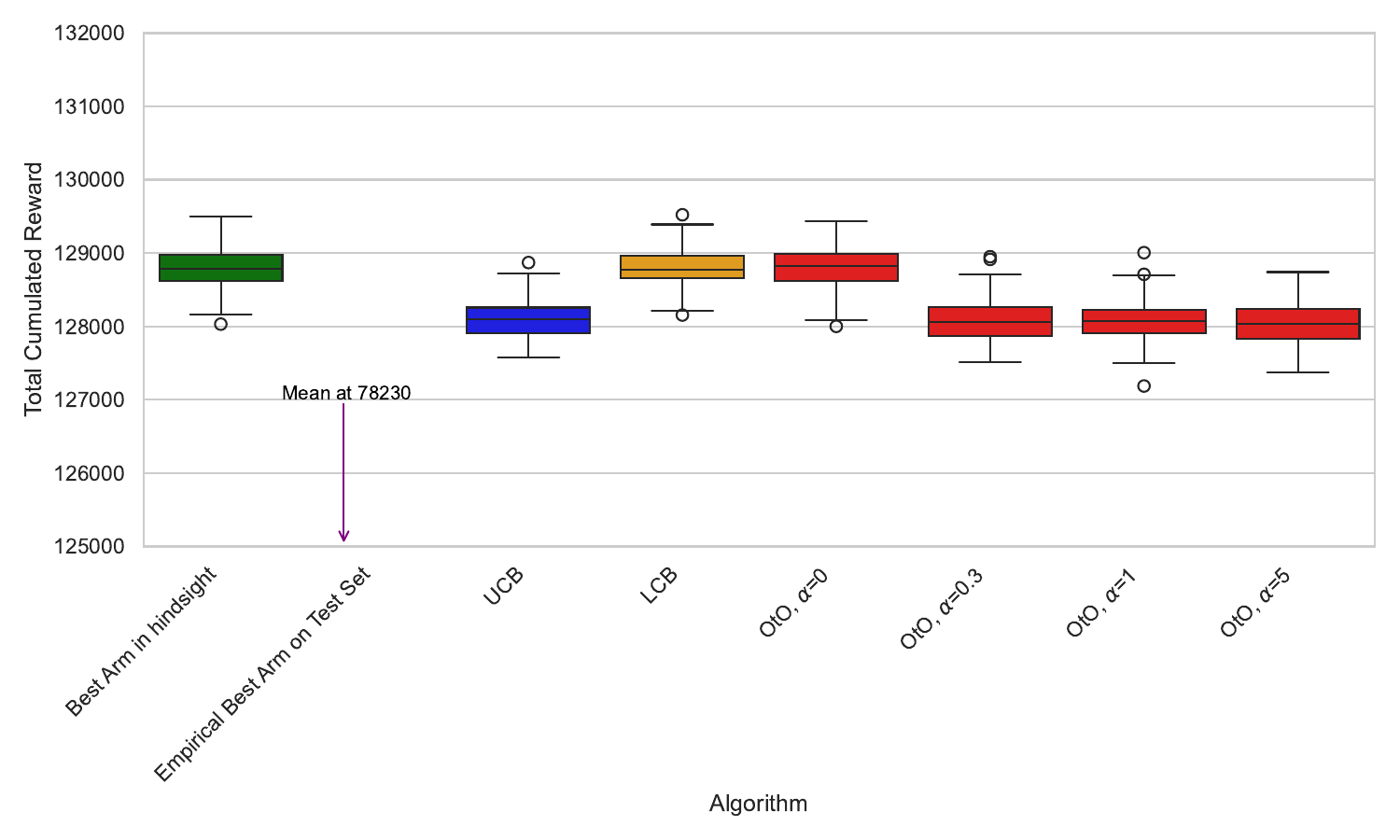} 
        \caption{Long horizon, $T=300,000$}
    \end{subfigure}
    \caption{The total cumulated reward by each algorithm in Setting 2}
    \label{fig:summaryexprealdatabadarm}
\end{figure}

\textbf{Results}: The results are summarized in \cref{fig:summaryexprealdatagoodarm,fig:summaryexprealdatabadarm}, where we provide box plots of the total cumulative reward by each algorithm, for each combination of setting and horizon length. Additional plots showing the average cumulative rewards can be found in \cref{app:simulationrealdata}.

The findings reveal that the model selected based on its performance on Test Set 2 consistently underperforms compared to other methods across all scenarios.  This outcome highlights a critical issue: the performance on a biased production-generated dataset does not necessarily translate to the true performance in production. This reinforces the importance of methods such as the proposed algorithm, which adaptively evaluates models in real-time production environments.

\cref{fig:summaryexprealdatagoodarm} shows the results for Setting 1 with two horizon lengths. For the short horizon, \alglcb outperforms \algucb, as the latter spends a substantial number of rounds exploring suboptimal options. Conversely, for the long horizon, \algucb surpasses \alglcb, as its extended exploration phase identifies the best prediction model. In both cases, \algoname demonstrates performance close to the best of the two algorithms. These results align with our theoretical analysis and the design principles of \algoname.

In Setting 2, the arm that generated offline data is the best arm, making \alglcb more desirable than in Setting 1. We observe this with both horizon lengths in \cref{fig:summaryexprealdatabadarm}. As expected, the performance gap between \algucb and \alglcb decreases as the horizon increases. Note that the $y$-axes in \cref{fig:summaryexprealdatabadarm} have different scales.
We observe that, similar to the previous experiments, \algoname interpolates between the performance of \algucb and \alglcb, close to the best of the two in any case. For low values of $\alpha$ (e.g., $\alpha=0$), it matches the performance of \alglcb, which is expected since a low $\alpha$ value strengthens the budget constraint.  For $\alpha=5$ (corresponding to $\alpha = 2\sqrt{K}$), \algoname behaves more like \algucb, which again aligns with the expected behavior. For $\alpha=1$ and $\alpha=0.3$, \algoname's performance is nearly identical to the best of the other two algorithms. When the performance gap between \alglcb and \algucb is large, \algoname's performance is close to the maximum cumulative reward of the two. This suggests that although $\alpha$ is a free parameter, the algorithm's behavior remains fairly robust to its choice.

%% file: proofs.tex
\section{Proof of the main theorem (\cref{thm:boundofftoon})} \label{sec:boundregretalgo}

Before going into the details of the proof, we start with a simple concentration inequality. By \cref{lem:hoeff} and a union bound, with probability at least $1-2T\delta$, for all $t\leq T$ and $i\in K$ the following inequalities hold:
\[
\underline{\mu}_i(t)\leq \mu_i \leq \overline{\mu}_i(t).
\]
 Assume now and for the rest of the proof that this event holds.  

We start by bounding $R^{\text{log}}(T)$. The bound can be split into two parts: the regret due to the iterations where \algucb\ is played and the regret due to the iterations where \alglcb\ is played, i.e.:
\begin{align}\label{eq:decloggingregret}
    R^{\text{log}}_\algoname(T)=\sum_{i=1}^{K}T_i^{U}(T)\left(\mu_0-\mu_i\right)+\sum_{i=1}^{K}T_i^L(T)\left(\mu_0-\mu_i\right).
\end{align}
We start by focusing on the former and bound the first term of the above equation. Consider the pseudo-budget:
\[
\tilde{B}_{\Tilde{T}}(t):=\sum_{i=1}^{K}T_i^U(t)(\underline{\mu}_i(t)-\gamma)+(T^L(t)+\Tilde{T}-t)\alpha \beta.
\]
Compared with the budget, 
\[
B_{\tilde{T}}(t):=\sum_{i=1}^{K}T_i^U(t-1)(\underline{\mu}_i(t)-\gamma)+\underline{\mu}_{U(t)}(t)-\gamma+(T^L(t-1)+\tilde{T}-t)\alpha \beta,
\]
the pseudo-budget can be seen as the "realized budget", where the actual arm played is considered instead of $U(t)$.
We show by induction on $t$ that the pseudo-budget is positive. 
If the budget is positive at iteration  $t$, then   $\tilde{B}_{\Tilde{T}}(t)$ and $B_{\Tilde{T}}(t)$ are equal and $\tilde{B}_{\Tilde{T}}(t)$ is positive. If the budget is negative, then $T_L(t)=T_L(t-1)+1$ and,  whether or not the proxy horizon has been modified between the two successive iterations, we have:
\[
\tilde{B}_{\Tilde{T}}(t)\geq  \tilde{B}_{\Tilde{T}}(t-1).
\]
Therefore, if $\tilde{B}_{\Tilde{T}}(t-1)$ is positive, then so is $\tilde{B}_{\Tilde{T}}(t)$. As $\tilde{B}_{\Tilde{T}}(0)$, is positive, we conclude, by induction, that the pseudo-budget is positive at every iteration. Notably, this implies that the final pseudo-budget is positive:
\begin{align*}
    \tilde{B}_{\Tilde{T}}(T)\geq 0. 
\end{align*}
Let us now show how this implies a bound on the first term of \cref{eq:decloggingregret}. Re-organizing the terms in $\tilde{B}_{\Tilde{T}}(T)$, and using that $\mu_i\geq \underline{\mu}_i(t)$ for all arms $i \in [K]$, we obtain from the positivity of $\tilde{B}_{\Tilde{T}}(T)$:
\begin{align}
\sum_{i=1}^K T_i^U(T)(\gamma-\mu_i)
    &\leq (T^L(T)+\Tilde{T}-T)\alpha \beta.\label{eq:playucbinter}
\end{align}
By definition, we have $
    \gamma= \underline{\mu}_{L(0)}(0)-\alpha \beta.
$ Also:
\begin{align}
\underline{\mu}_{L(0)}(0)
      \geq &\sum_{i=1}^K\frac{m_i \underline{\mu}_{i}(0)}{m}\notag\\
      \geq &\frac{1}{\sum_i m_i}\sum_i m_i\left(\mu_i-\sqrt{2\frac{\log(\frac{K}{\delta})}{m_i}}\right)\notag\\
      \geq& \mu_0-\beta. \label{eq:lbmul0}
\end{align}

This implies $
   \gamma \geq \mu_0-\left(1+\alpha\right)\beta.
$
Injecting in \cref{eq:playucbinter} and re-organizing the terms,
we get the following bound on the first term of \cref{eq:decloggingregret}:
\begin{align}\label{bound:playsucb}
\sum_{i=1}^K T_i^U(T)(\mu_0-\mu_i)\leq \sum_{i=1}^{K}T_i^U(T)(1+\alpha)\beta+(T^L(T)+\Tilde{T}-T)\alpha \beta.
\end{align}
Let us now bound the second term of \cref{eq:decloggingregret}, $\sum_{i=1}^{K}T_i^L(T)\left(\mu_0-\mu_i\right)$, i.e. the regret w.r.t. the logging policy of iterations where \alglcb is played. For any $i \in [K]$ s.t. $L(t)=i$ for some $i\in [K]$ and some $t\leq T$, we have:
\begin{align}
\mu_i\geq& \underline{\mu}_{L(t)}(t)\notag\\
\geq& \underline{\mu}_{L(0)}(0)\notag\\
      \geq& \mu_0-\beta, \label{eq:lbmult}
\end{align}
where the last line comes from \cref{eq:lbmul0}.
Hence, we obtain:
\begin{equation*}
    \sum_{i=1}^{K}T_i^L(T)\left(\mu_0-\mu_i\right)\leq T^L(T)\beta.
\end{equation*}
Injecting this and \cref{bound:playsucb} in \cref{eq:decloggingregret}, we have:
\begin{align*}
    R^{\text{log}}_\algoname(T)\leq &\sum_{i=1}^{K}T_i^U(T)(1+\alpha)\beta+(T^L(T)+\Tilde{T}-T)\alpha \beta+ T^L(T)\beta\\
    \leq& T\left(1+\alpha\right)\beta +(\Tilde{T}-T)\alpha \beta.
\end{align*}

When the horizon is known, $\Tilde{T}=T$, otherwise, it always holds that $\Tilde{T}\leq 2T$, hence:
\begin{align*}
   R^{\text{log}}_\algoname(T)\leq &T\left(1+(1+\mathds{1}_{\text{T unknown}})\alpha\right)\beta.   
\end{align*}
This is exactly the bound of the theorem.

Let us now turn to the bound on $R_\algoname(T)$. Consider an arm $i$ with $\Delta_i>0$. If $U(t)=i$ for some $t$, we have:
\begin{align*}
    \overline{\mu}_i(t)\geq& \overline{\mu}_*(t).
\end{align*}
This implies:
\begin{align*}
    \mu_i+\sqrt{2\frac{\log(K/\delta)}{(m_i+T_i(t))}}\geq& \mu_*,
\end{align*}
hence:

\begin{align*}
  m_i+T_i(t)\leq& \frac{2\log(K/\delta)}{\Delta_i^2},
\end{align*}
and finally, we obtain:
\begin{align}\label{eq:T_i^T}
T_i^U(T)\leq \left(\frac{2\log(K/\delta)}{\Delta_i^2}-m_i\right)_++1.
\end{align}

We now define
\[
\tau:= \max \left\{t\leq T \text{ s.t. } I(t)\neq U(t) \right\},
\]
the last iteration for which \algucb\ is not played. By algorithm design, this implies that the budget is negative at this iteration:
\begin{align}\label{eq:budgetneg}
B_{\tilde{T}}(\tau)=\sum_{i=1}^{K}T_i^U(\tau-1)(\underline{\mu}_{i}(\tau)-\gamma)+\underline{\mu}_{U(\tau)}(\tau)-\gamma+(T^L(\tau-1)+\tilde{T}-\tau)\alpha \beta<0.
\end{align}
Because $\underline{\mu}_{U(\tau)}(\tau)\ge \mu_{U(\tau)}-\sqrt{2\log(K/\delta)}$ and $\alpha \beta >0$, we have
 \begin{align*}
 \gamma-\underline{\mu}_{U(\tau)}(\tau)&=\underline{\mu}_{L(0)}(0)-\alpha\beta - \underline{\mu}_{U(\tau)}(\tau)\\
 &\le \mu_{L_0} -\mu_{U(\tau)} + \sqrt{2\log(K/\delta)}\\
 &\le 1+\sqrt{2\log(K/\delta)}.
 \end{align*}
 Using the above and again the fact that $\alpha \beta >0$, \eqref{eq:budgetneg} implies: 
\begin{align*}
 \sum_{i=1}^{K}T_i^U(\tau-1)(\underline{\mu}_{i}(\tau)-\gamma)-1-\sqrt{2\log(K/\delta)}+T^L(\tau-1)\alpha \beta<0.
\end{align*}
Rearranging the terms, we obtain:
\begin{align}
T^L(\tau-1) \alpha\beta< &1+\sqrt{2\log(K/\delta)}+\sum_{i=1}^K T^U_i(\tau-1)\left(\gamma-\underline{\mu}_{i}(\tau)\right)\notag\\
\leq &1+\sqrt{2\log(K/\delta)}+\sum_{i=1}^K T^U_i(\tau-1)\left(\gamma-\mu_i+\sqrt{2\frac{\log(K/\delta)}{(m_i+T_i(t))}}\right)\notag\\
\leq& 1+\sqrt{2\log(K/\delta)}+\sum_{i=1}^K \underbrace{T^U_i(\tau-1)\left(\Delta_i-(\mu_*-\gamma)\right)+\sqrt{2 T^U_i(\tau-1)\log(K/\delta)}}_{=:S_i}.\label{eq:boundT0}
\end{align}

Denote $a_i:=\Delta_i-(\mu_*-\gamma)$. Note that by definition of $\gamma$, $a_i<\Delta_i$ always holds. Then,  if $a_i>0$, \cref{eq:T_i^T} gives:
\begin{align}
S_i\leq& \frac{2 \log(K/\delta)}{\Delta_i}+\Delta_i+\sqrt{\frac{4 \log^2(K/\delta)}{\Delta_i^2}+2\log(K/\delta)}\notag\\
\leq&\frac{4 \log(K/\delta)}{\Delta_i}+2\log(K/\delta)+1\notag\\
\leq &\frac{4 \log(K/\delta)}{\mu_*-\gamma}+2\log(K/\delta)+1.\label{eq:boundSiposai}
\end{align}
If $a_i<0$, we have:
\[
S_i\leq \sqrt{T^U_i(\tau-1)\log(K/\delta)}\leq \frac{2 \log(K/\delta)}{\Delta_i}+2\log(K/\delta)+1,
\]
and by using $a x^2+b x \leq-b^2 / 4 a$ for $a<0$ we have
\[
S_i \leq-\frac{\log(K/\delta)}{ a_i}=\frac{\log(K/\delta)}{\left((\mu_*-\gamma)-\Delta_i\right)} .
\]

Summarizing the two bounds gives
\begin{align}
S_i \leq \frac{4 \log(K/\delta)}{\max \left\{\Delta_i, (\mu_*-\gamma)-\Delta_i\right\}}+2\log(K/\delta)+1 .\label{eq:boundSi}
\end{align}
We also have:
\begin{align*}
    \max \left\{\Delta_i, (\mu_*-\gamma)-\Delta_i\right\}\geq &\frac{\mu^*-\gamma}{2}.
\end{align*}
Combining with \cref{eq:boundSiposai}, we obtain that for any $i\in [K]$,
\[
S_i\leq \frac{8 \log(K/\delta)}{\mu^*-\gamma}+2\log(K/\delta)+1
\]
Injecting in \cref{eq:boundT0}, we obtain:

\begin{align}\label{eq:T^L}
T^L(\tau-1) \leq &\frac{1}{\alpha \beta}\left(\frac{8 K\log(K/\delta)}{\mu^*-\gamma}+2K\log(K/\delta)+K+1+\sqrt{2\log(K/\delta)}\right)
\end{align}

Now, we have:
\begin{align*}
    \mu_{L(t)}\geq& \underline{\mu}_{L(0)}(0)\\
    \geq& \gamma,
\end{align*}
hence:
\begin{align*}
   R_\algoname(T)\leq \sum_{i \in [K]}\Delta_i T^U_i(T)+T^L(\tau-1) \min\left(1;\mu^*-\gamma\right).
\end{align*}
Injecting \cref{eq:T^L} and \cref{eq:T_i^T}, we get:
\begin{align*}
   R_\algoname(T)\leq&\sum_{i=1}^K\Delta_i\left(\frac{4\log(K/\delta)}{\Delta_i^2}-m_i\right)_++\frac{1}{\alpha \beta}\left(10 K\log(K/\delta)+K+1+\sqrt{2\log(K/\delta)}\right)+K\\
   \leq&\sum_{i=1}^K\Delta_i\left(\frac{4\log(K/\delta)}{\Delta_i^2}-m_i\right)_++\frac{(12 K+2)\log(K/\delta)}{\alpha \beta}+K,
\end{align*}
by using $\log(K/\delta)>1/2$.

With some computations to bound $\sum_{i \in [K]}\Delta_i T^U_i(T)$, that are detailed in  the proof of \cref{prop:regretminimaxucb},  we also have:
\begin{align*}
   \mathcal{R}_\algoname(T)\leq&
   \max_{J\subseteq [K]}2T\sqrt{\frac{2|J|\log(K/\delta)}{T+\sum_{j\in J}m_j}}+|J|+\frac{12 K\log(K/\delta)}{\alpha \beta}+2T^2\delta.
\end{align*}

\hfill \( \Box\)

%% file: conc.tex
\section{Conclusions}

We explored the offline-to-online learning problem within the multi-armed bandit framework. This problem involves starting with historical, offline data and then improving performance through online interactions. We proposed that a natural way to evaluate algorithm performance in this setting was to compare against the logging policy in short-horizon scenarios, where there was limited opportunity for effective exploration, and against the optimal arm in long-horizon settings, where accumulated data allowed for more informed decision-making. These two objectives are inherently competing, and the distinction between what constituted a short or long horizon depended on the specific instance, which is the central challenge we addressed.

To address this, we introduced a novel algorithm, \algoname, designed to dynamically balance the benefits of the Lower Confidence Bound (\alglcb) and Upper Confidence Bound (\algucb) algorithms. \algoname was shown to adapt seamlessly across different conditions without prior knowledge of whether to prioritize exploration or exploitation, maintaining robust performance across a range of scenarios. 

Our experimental results further supported these findings. Through evaluations on both synthetic and real-world datasets, \algoname consistently demonstrated strong performance across different horizon lengths and problem instances. The experiments highlighted how \algoname effectively interpolated between the strengths of \alglcb and \algucb, confirming its robustness and adaptability in practice.

Overall, our work bridges a critical gap in offline-to-online learning and offers a robust, adaptive approach that we hope will inspire continued exploration in this evolving field. In particular, we believe that the ideas underlying \algoname can extend naturally to more complex settings, such as contextual bandits, reinforcement learning, and nonstationary environments, which reflect more practical scenarios closer to real-world applications.

%% file: acknowledgements.tex
\section*{Acknowledgements}

Csaba Szepesvari acknowledges support from NSERC Discovery and CIFAR AI Chair, and Ilbin Lee from NSERC Discovery.

%% file: appendix.tex
\begin{APPENDICES}
\begin{table}[h]
\caption{Notation Summary}
    \centering
    \renewcommand{\arraystretch}{1.5}
    \begin{tabular}{|p{2.5cm}|p{5cm}|p{2.5cm}|p{5cm}|}
        \hline
        \textbf{Notation} & \textbf{Description} & \textbf{Notation} & \textbf{Description} \\ \hline
        $K$ & \small Number of arms & $m_i$ & \small Offline samples from arm $i$ \\ \hline
        $\mathcal{P}_i$ & \small Distribution of arm $i$ (1-subgaussian) & $\mathbf{m}$ & \small Vector of samples for each arm \\ \hline
        $\mu_i$ & \small Mean reward of arm $i$, $\mu_i \in [0,1]$ & $\Theta_{\mathbf{m}}$ & \small Set of instances with $\mathbf{m}$ fixed \\ \hline
        $\mu^*$ & \small Mean reward of the optimal arm & $\hat{\mu}_i(0)$ & \small Empirical mean of $m_i$ samples from arm $i$ \\ \hline
        $\Delta_i$ & \small Suboptimality gap of arm $i$, $\Delta_i = \mu^* - \mu_i$ & $m$ & \small Total number of offline samples, $m = \sum_{i \in [K]} m_i$ \\ \hline
        $\Theta$ & \small Set of all instances & $\hat{\mu}_0$ & \small Weighted empirical mean, $\hat{\mu}_0 = \frac{\sum_{i \in [K]} m_i \hat{\mu}_i(0)}{m}$ \\ \hline
        $\mu_0$ & \small Weighted true mean, $\mu_0 = \frac{\sum_{i \in [K]} m_i \mu_i}{m}$ & $\Delta_0$ & \small Gap between $\mu^*$ and $\mu_0$, $\Delta_0 = \mu^* - \mu_0$ \\ \hline
        $t$ & \small Time step in online phase &  $I(t)$& \small Arm chosen at time $t$ \\ \hline
        $x^t_{I(t)}$ & \small Reward from pulling $I(t)$ at time $t$ & $T_i(t)$ & \small Times arm $i$ has been pulled by time $t$ \\ \hline
        $ \hat{\mu}_i(t)$ & \small Empirical mean of arm $i$ at time $t$ &  $\overline{\mu}_i(t)$& \small Upper bound for arm $i$ at $t$ \\ \hline
        $\underline{\mu_i}(t)$ & \small Lower bound for arm $i$ at $t$ & $R(T)$ & \small Pseudo-regret over horizon $T$, $R(T) = T \mu^* - \sum_{t=1}^T \mu_{I(t)}$ \\ \hline
        $R^{\text{log}}(T)$ & \small Logging pseudo-regret over $T$, $R^{\text{log}}(T) = T \mu_0 - \sum_{t=1}^T \mu_{\pi(t)}$ & $\mathcal{R}_\pi(T)$ & \small Minimax pseudo-regret, $\mathcal{R}_\pi(T) = \max_{\theta \in \Theta_{\mathbf{m}}} \mathbb{E}_{\theta,\pi}[R_\pi(T)]$ \\ \hline
        $\mathcal{R}^{\text{log}}_\pi(T)$ & \small Minimax logging pseudo-regret, $\mathcal{R}^{\text{log}}_\pi(T) = \max_{\theta \in \Theta_{\mathbf{m}}} \mathbb{E}_{\theta,\pi}[R^{\text{log}}(T)]$ & $L(t)$& \small Arm chosen by \alglcb\ at time $t$\\ \hline
        $U(t)$& \small Arm chosen by \algucb\ at time $t$ & $\beta$ & \small Constant for exploration budget, $\beta = \frac{\sum_i \sqrt{m_i}}{m} \sqrt{2\log(\frac{1}{\delta})}$ \\ \hline
        $\gamma$ & \small Safety threshold, $\gamma = \underline{\mu_{L(0)}(0)} - \alpha \beta$ & $\alpha$ & \small Parameter determining budget stringency \\ \hline
        $T_i^U(t)$ & \small Times arm $i$ was played as \algucb\ arm by time $t$ & $T^L_i(t)$ & \small Times arm $i$ was played as \alglcb\ arm by time $t$ \\ \hline
         $T^L(t)$& \small Total times \alglcb\ was played by time $t$,  &  $B_{T}(t)$& \small Exploration budget at time $t$, see definition in the text \\ \hline
        $\tilde{T}$ & \small Proxy horizon, initially 2, doubled each time $t > \tilde{T}$ in the unknown horizon case &  &  \\ \hline
    \end{tabular}
\end{table}

\section{Additional Experiments}\label{app:extrasim}

    As mentioned in \cref{sec:algorithm}, there are multiple ways to set the budget, which could potentially result in similar theoretical guarantees.  The chosen version of the budget:

    \[
B_{T}(t)=\sum_{i=1}^{K}T_i^U(t-1)(\underline{\mu_i}(t)-\gamma)+\underline{\mu_{U(t)}}(t)-\gamma+(T^L(t-1)+T-t)\alpha \beta,
\]
does not update in the same way when the chosen policy is \algucb\ or \alglcb. It is natural to wonder if unifying the update as:
        \[
B^{'}_{T}(t)=\sum_{i=1}^{K}T_i(t-1)(\underline{\mu_i}(t)-\gamma)+\underline{\mu_{U(t)}}(t)-\gamma+(T-t)\alpha \beta,
\]
would also work.
\begin{figure}[h!]
\begin{centering}
\begin{subfigure}{.48\textwidth}
  \centering
  \includegraphics[width=\linewidth]{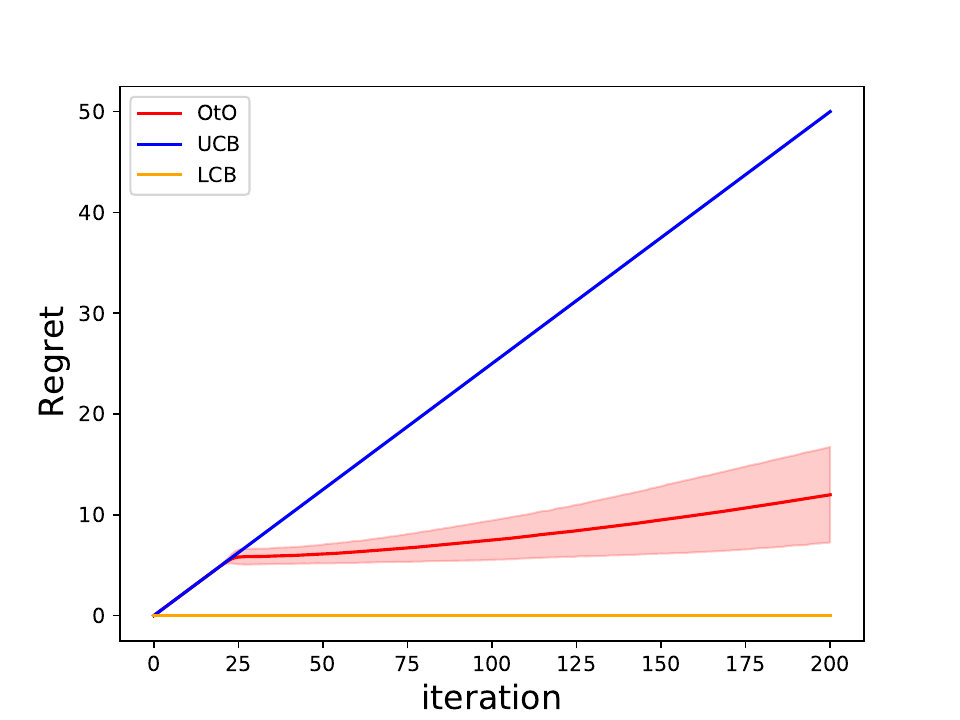}
  \caption{Instance $1$, $T=2000$}
\end{subfigure}%
\begin{subfigure}{.48\textwidth}
  \centering
  \includegraphics[width=\linewidth]{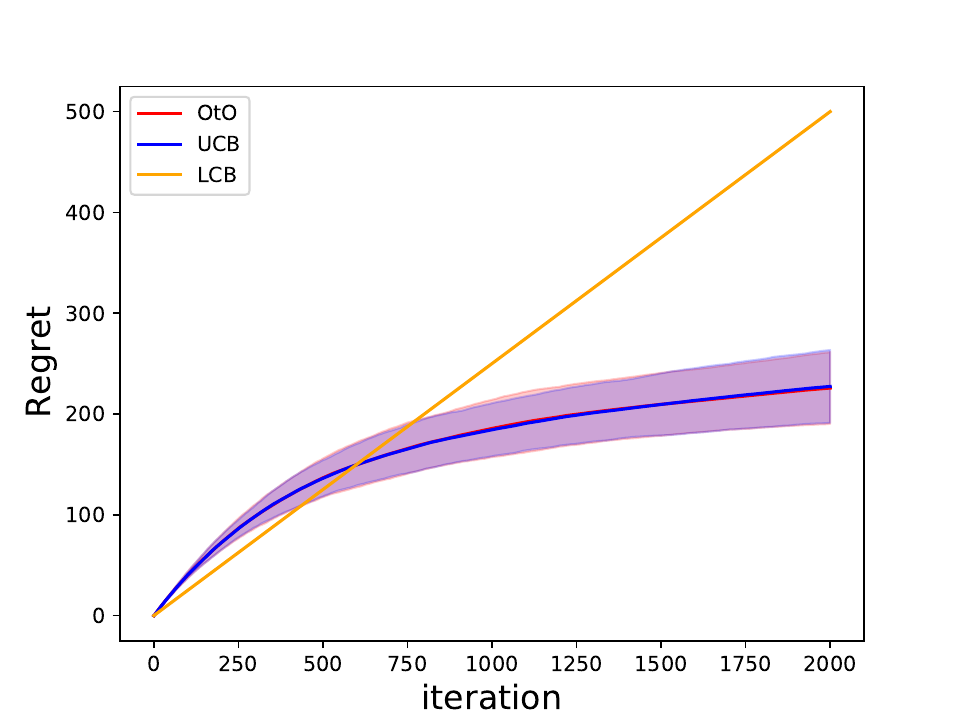}
  \caption{Instance $2$, $T=2000$}
\end{subfigure}
\begin{subfigure}{.48\textwidth}
  \centering
  \includegraphics[width=\linewidth]{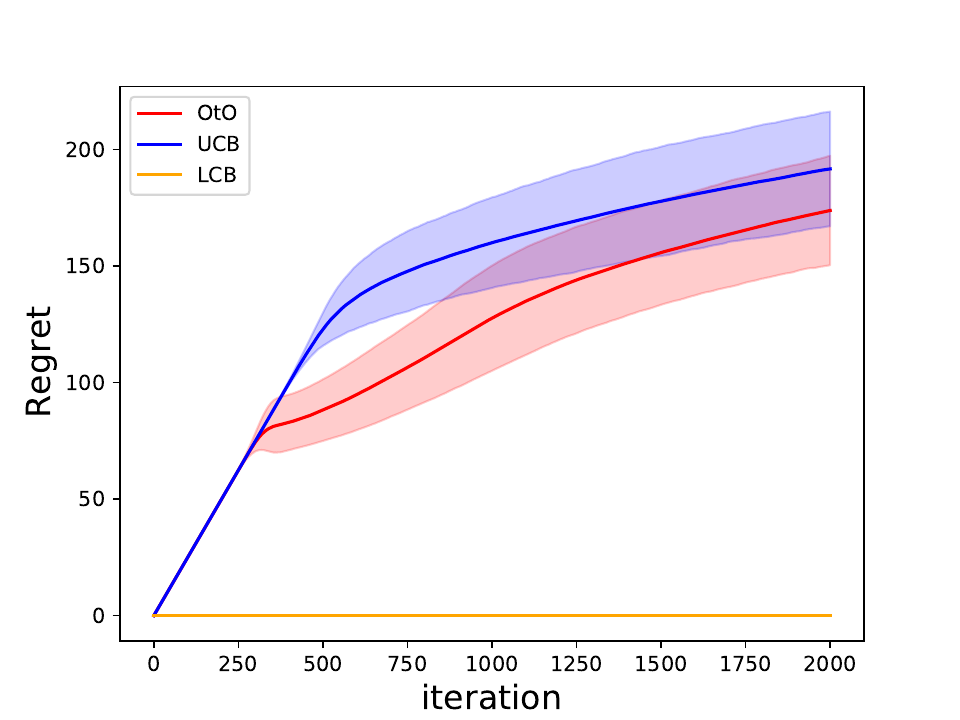}
  \caption{Instance $1$, $T=2000$}
\end{subfigure}%
\begin{subfigure}{.48\textwidth}
  \centering
  \includegraphics[width=\linewidth]{plots/plots_appendices/partialupdateparamTlargeoptnotsampled.pdf}
  \caption{Instance $2$, $T=2000$}
\end{subfigure}
\caption{Regret of the three algorithms for different instances and $T$ values, when the horizon $T$ is given to \algoname, with \algoname implemented with the first alternative version of the budget and $\alpha=0.2$.}
\label{fig:badversion}
\end{centering}
\end{figure}

 Regarding theoretical guarantees, we suspect that a similar result can be proven. However, as can be seen in \cref{fig:badversion}, the algorithm with that alternative budget does not perform as well as the chosen one. We observe in \cref{fig:badversion} that the algorithm switches to \alglcb, then gradually starts playing \algucb again. The reason for that behavior is simple: as the arm $L(0)$ is played more and more, $\underline{\mu_{L(0)}}(t)$ increases, while $\gamma$ remains unchanged. Thus, the projected budget is underestimated, and the algorithm returns to exploration towards the end of the horizon.
\begin{figure}[htb]
\begin{centering}
\begin{subfigure}{.48\textwidth}
  \centering
  \includegraphics[width=\linewidth]{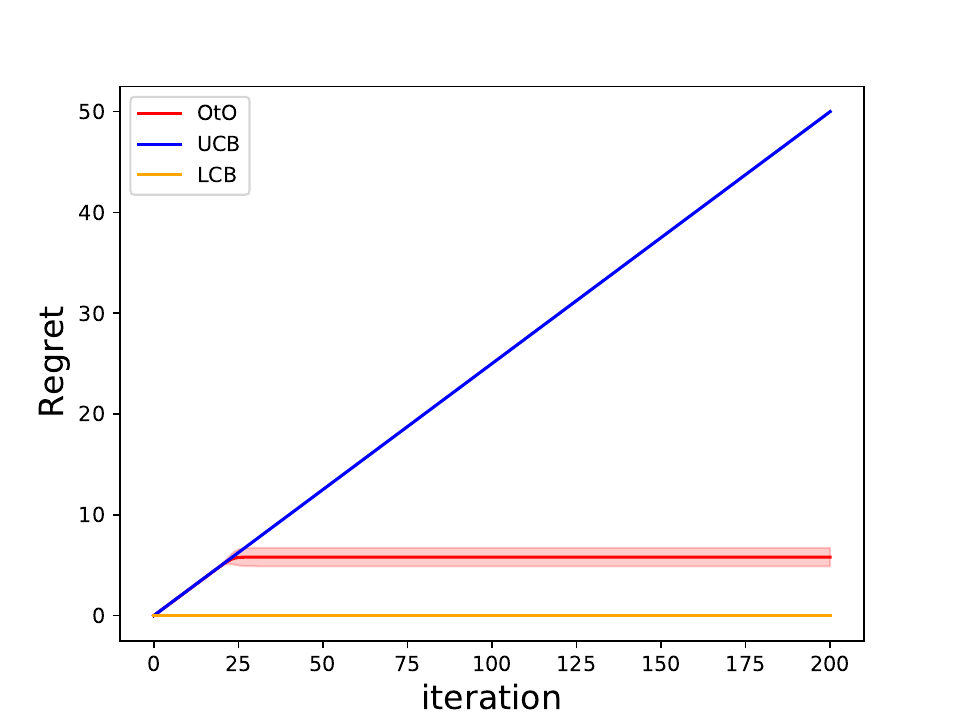}
  \caption{Instance $1$, $T=2000$}
\end{subfigure}%
\begin{subfigure}{.48\textwidth}
  \centering
  \includegraphics[width=\linewidth]{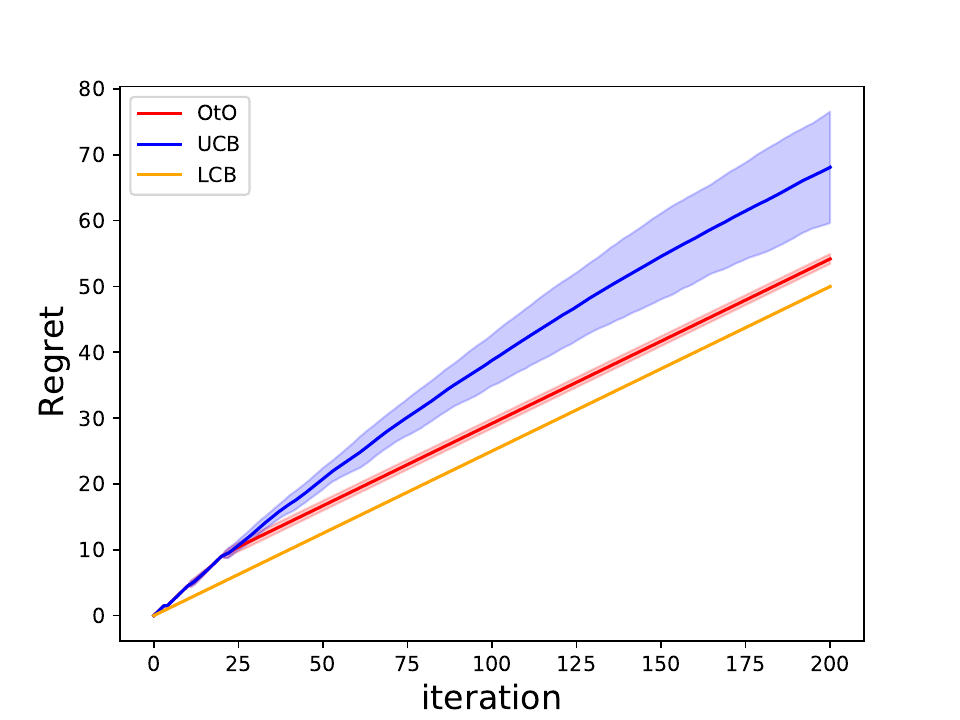}
  \caption{Instance $2$, $T=2000$}
\end{subfigure}
\begin{subfigure}{.48\textwidth}
  \centering
  \includegraphics[width=\linewidth]{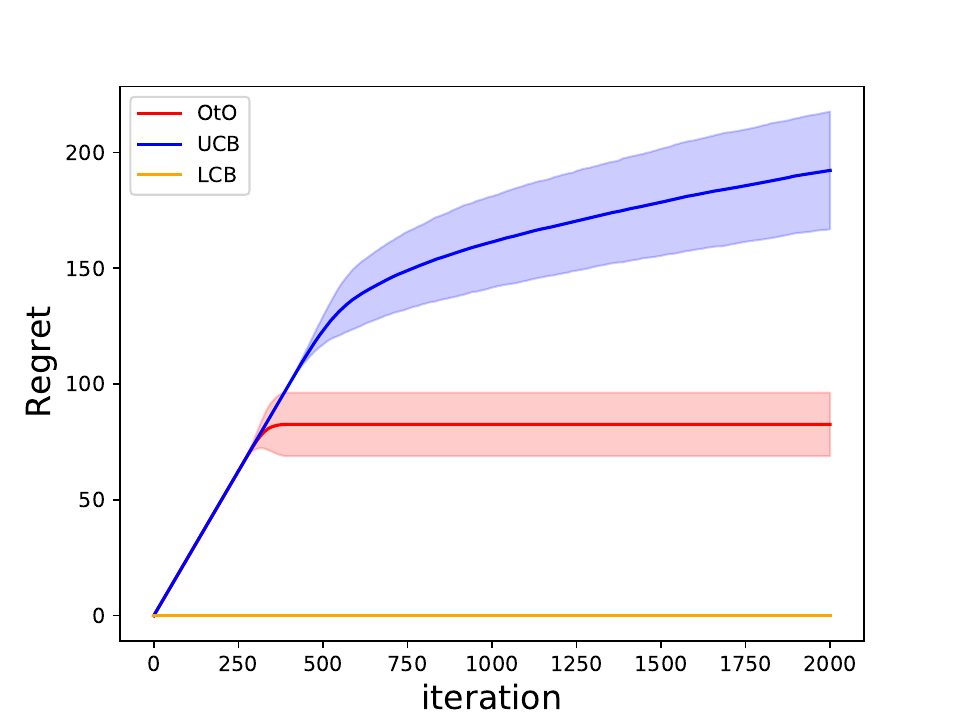}
  \caption{Instance $1$, $T=2000$}
\end{subfigure}%
\begin{subfigure}{.48\textwidth}
  \centering
  \includegraphics[width=\linewidth]{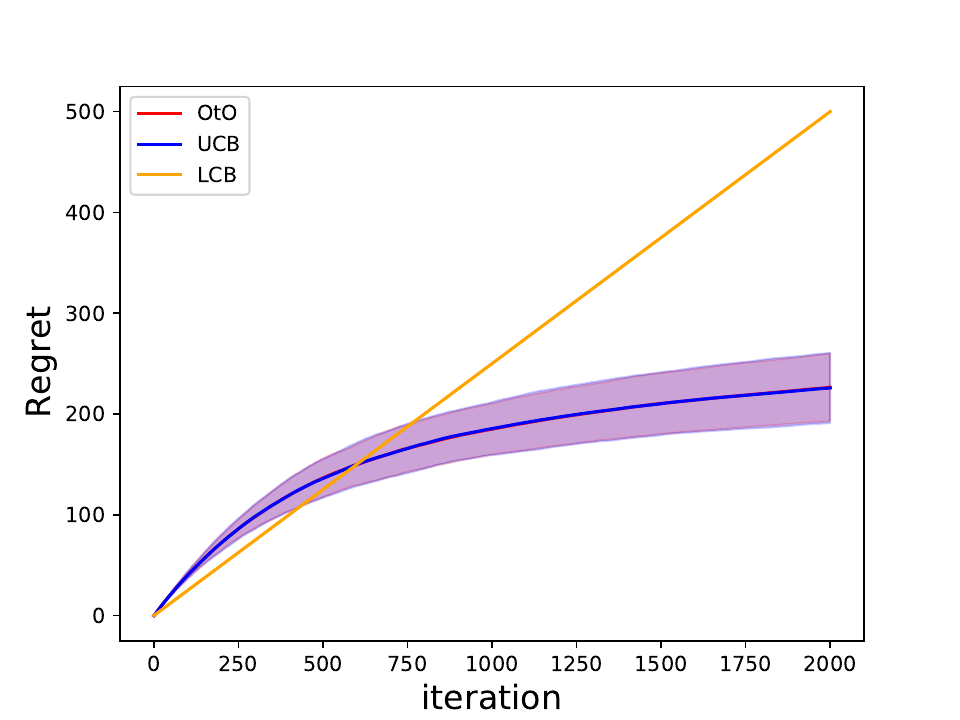}
  \caption{Instance $2$, $T=2000$}
\end{subfigure}
\caption{Regret of the three algorithms for different instances and $T$ values, when the horizon $T$ is given to \algoname, with \algoname implemented with the second alternative version of the budget and $\alpha=0.2$.}
\label{fig:fullupdateparam}
\end{centering}
\end{figure}

A second natural option is to then also update $\gamma$ as 
\[
\gamma_t= \underline{\mu_{L(0)}}(t)-\alpha\beta,
\]
 and have:

  \[
B^{''}_{T}(t)=\sum_{i=1}^{K}T_i(t-1)(\underline{\mu_i}(t)-\gamma_t)+\underline{\mu_{U(t)}}(t)-\gamma_t+(T-t)\alpha \beta,
\]
As can be seen in \cref{fig:fullupdateparam}, this version performs comparably to the one chosen. However, we did not analyze this one further, as it seems harder to obtain theoretical guarantees for this version. The main challenge is that an arm $i$ that is chosen as $L(t)$ at some iteration may no longer be chosen again. Its lower bound $\underline{\mu}_i(t)$ would then no longer update, and it is not straightforward to see if $\underline{\mu}_i(t')>\gamma_{t'}$ would hold for subsequent iterations. This implies in turn that it is not easy to guarantee the budget remains positive. Also, this version did not show any advantage in empirical performance, compared to the presented version. 

\section{Simulation using Ad Click Through Rate data}\label{app:simulationrealdata}

The prediction models were constructed based on methodologies from a top-performing submission in the Kaggle data challenge for the Avazu dataset \citep{kaggle_deepctr_difm}. Three state-of-the-art methods were selected: the Factorization-Machine-based Neural Network (DeepFM, \citealt{guo2017deepfmfactorizationmachinebasedneural}), the Deep \& Cross Network (DCN, \citealt{wang2017deepcrossnetwork}), and the Dual Input-aware Factorization Machine (DIFM) for CTR Prediction \citep{Lu2020ADI}. All methods were implemented using the {\tt DeepCTR-Torch} Python package \citep{deepctr_torch}. Feature engineering included transforming sparse categorical features through label encoding or one-hot encoding and normalizing dense features, following the exact methodology described in the selected Kaggle notebook.

To create six distinct models, two different configurations were used for the number of neurons in the neural network layers underlying each method. DCN and DeepFM, configured with the default parameters provided in the notebook, were used to construct Model 1 and Model 8, respectively. The remaining models (Models 2–7) were generated by applying each method—DeepFM, DIFM, and DCN—with the two distinct parameter configurations.

The results of these experiments are summarized in \cref{sec:exprealdata}. Additional plots of cumulative rewards are provided in \cref{fig:plotavcumrewardsetting1,fig:plotavcumrewardsetting2} to further illustrate these findings.

\begin{figure}[htb]
\begin{centering}
\begin{subfigure}{.48\textwidth}
  \centering
  \includegraphics[width=\linewidth]{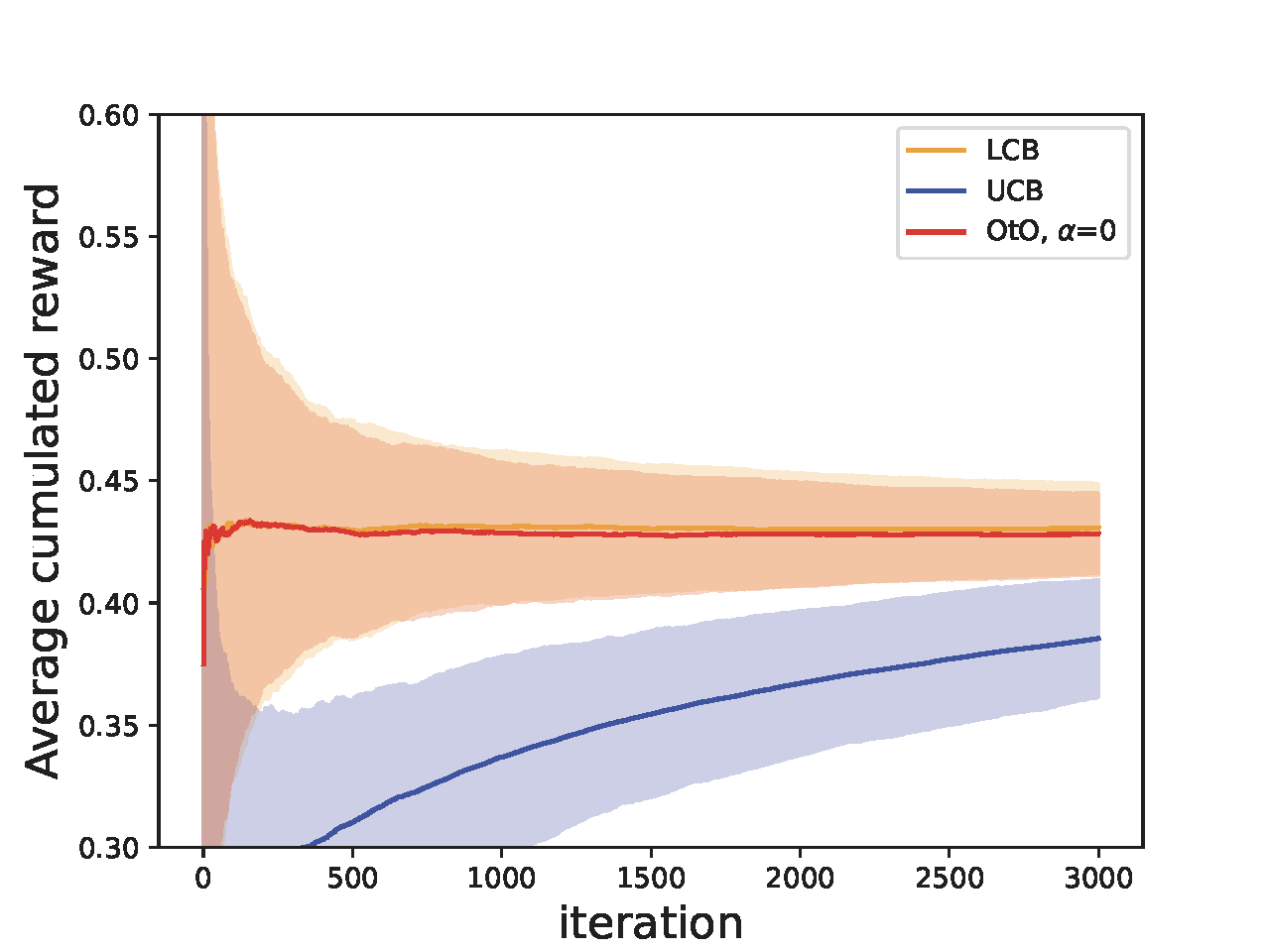}
\end{subfigure}%
\begin{subfigure}{.48\textwidth}
  \centering
  \includegraphics[width=\linewidth]{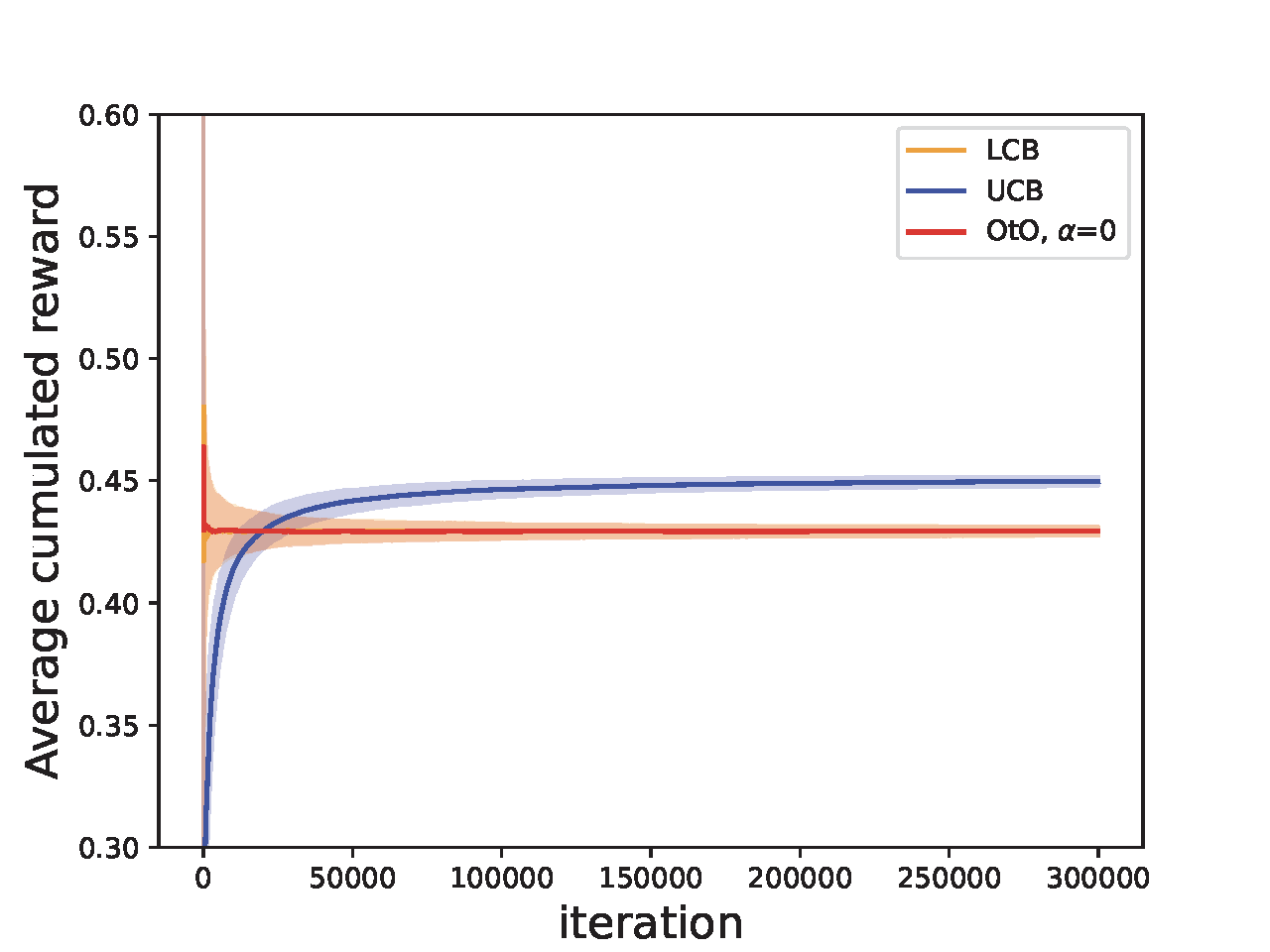}
\end{subfigure}
\begin{subfigure}{.48\textwidth}
  \centering
  \includegraphics[width=\linewidth]{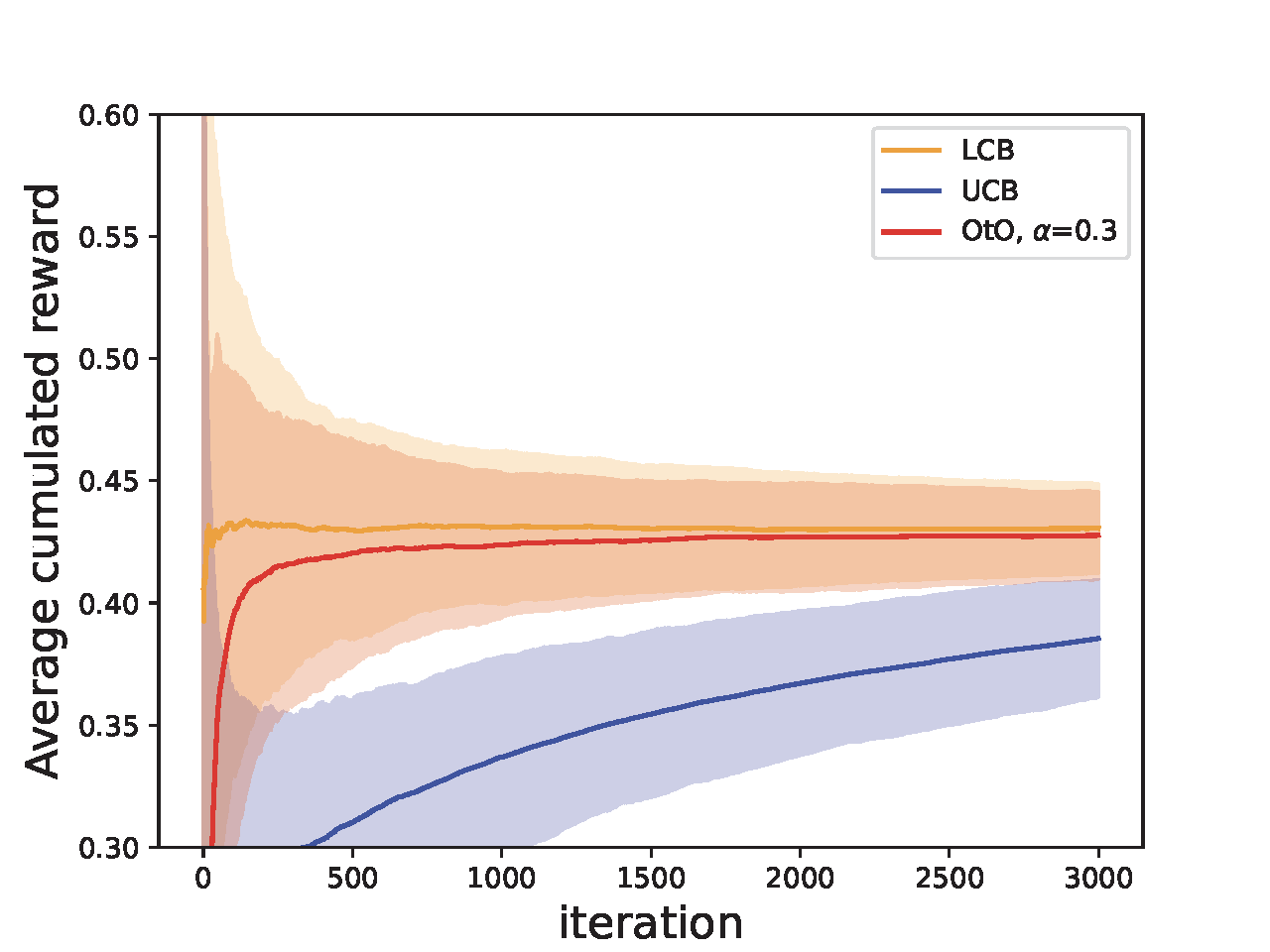}
\end{subfigure}%
\begin{subfigure}{.48\textwidth}
  \centering
  \includegraphics[width=\linewidth]{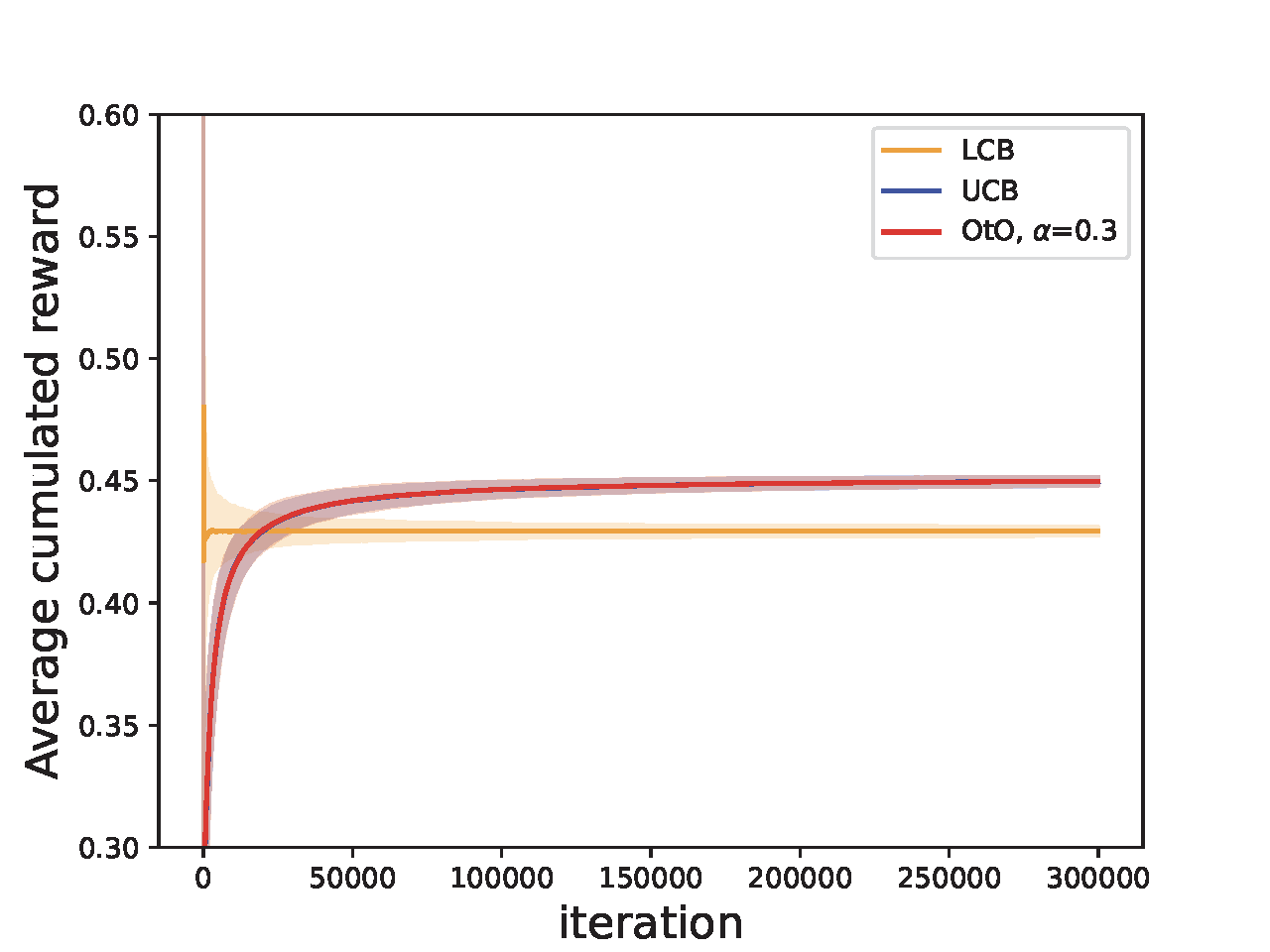}
\end{subfigure}
\begin{subfigure}{.48\textwidth}
  \centering
  \includegraphics[width=\linewidth]{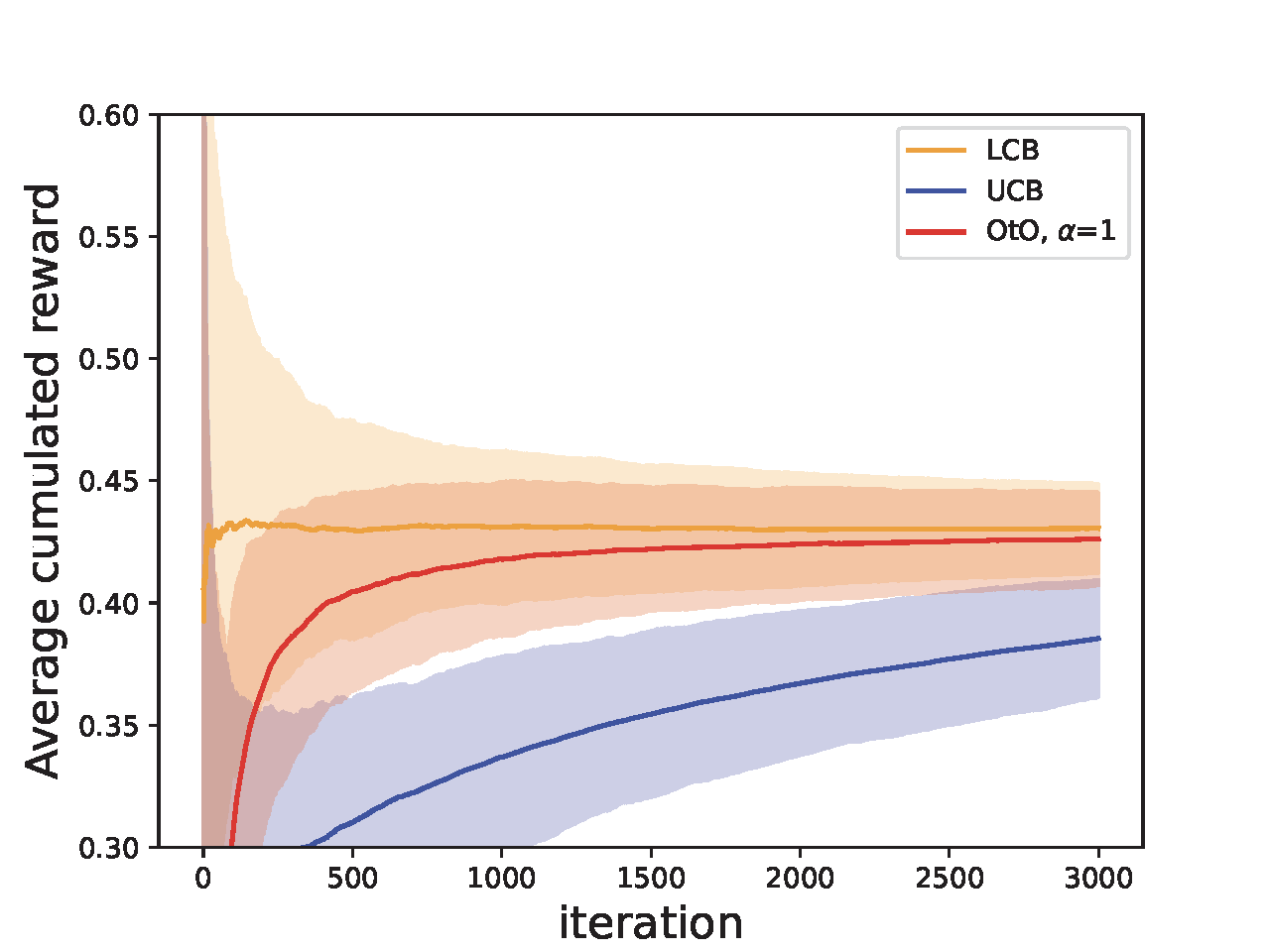}
\end{subfigure}%
\begin{subfigure}{.48\textwidth}
  \centering
  \includegraphics[width=\linewidth]{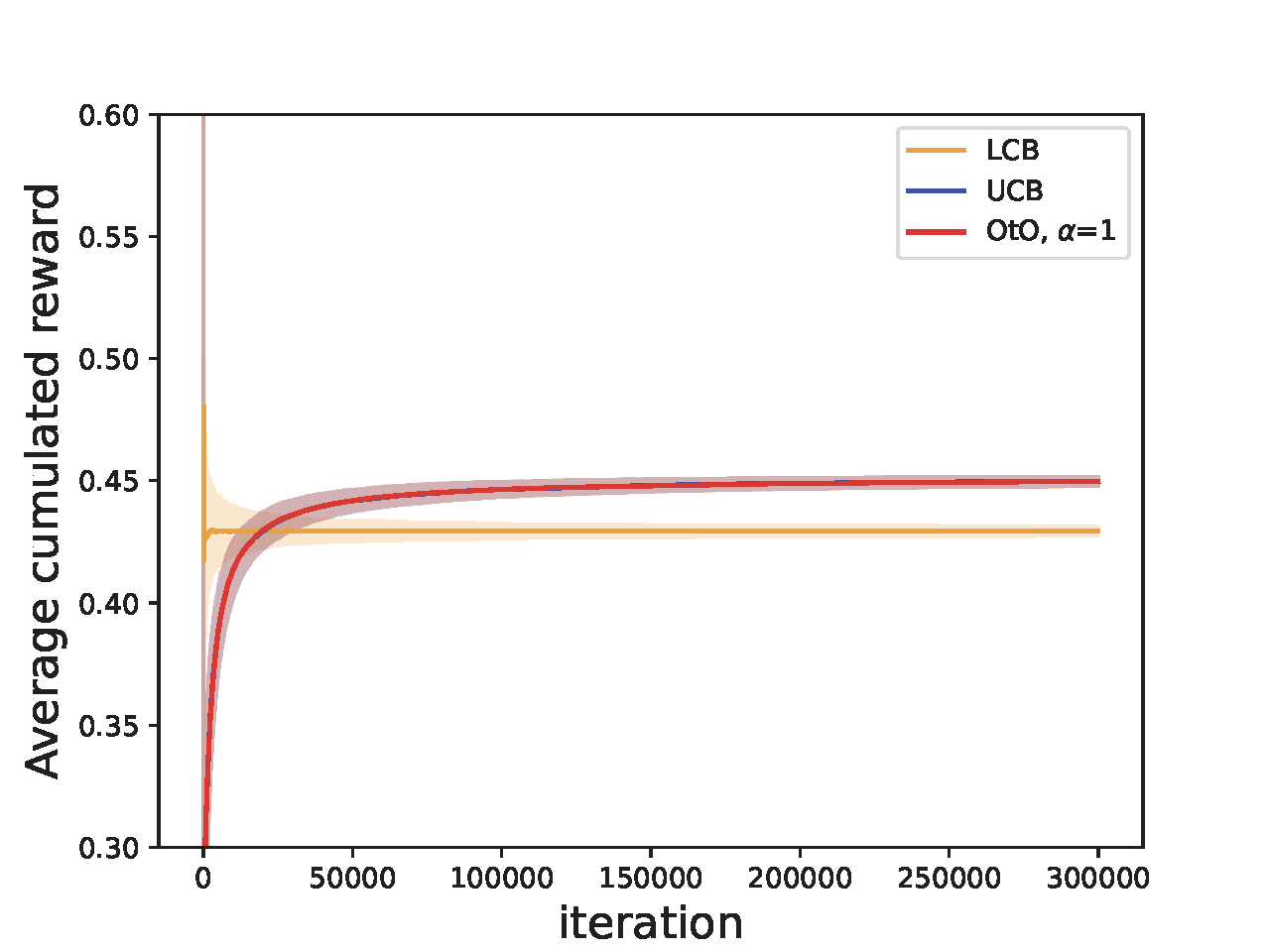}
\end{subfigure}
\begin{subfigure}{.48\textwidth}
  \centering
  \includegraphics[width=\linewidth]{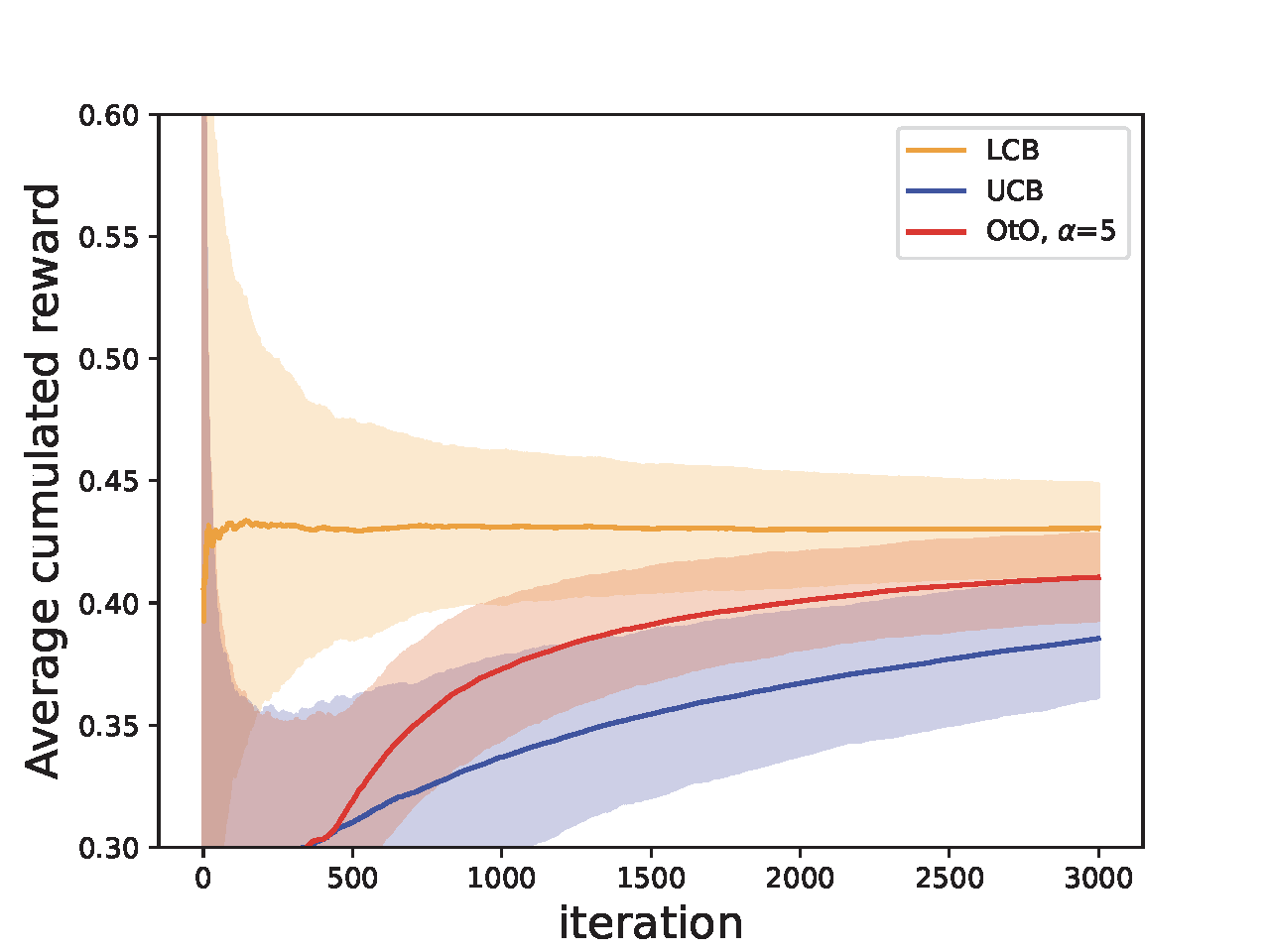}
\end{subfigure}%
\begin{subfigure}{.48\textwidth}
  \centering
  \includegraphics[width=\linewidth]{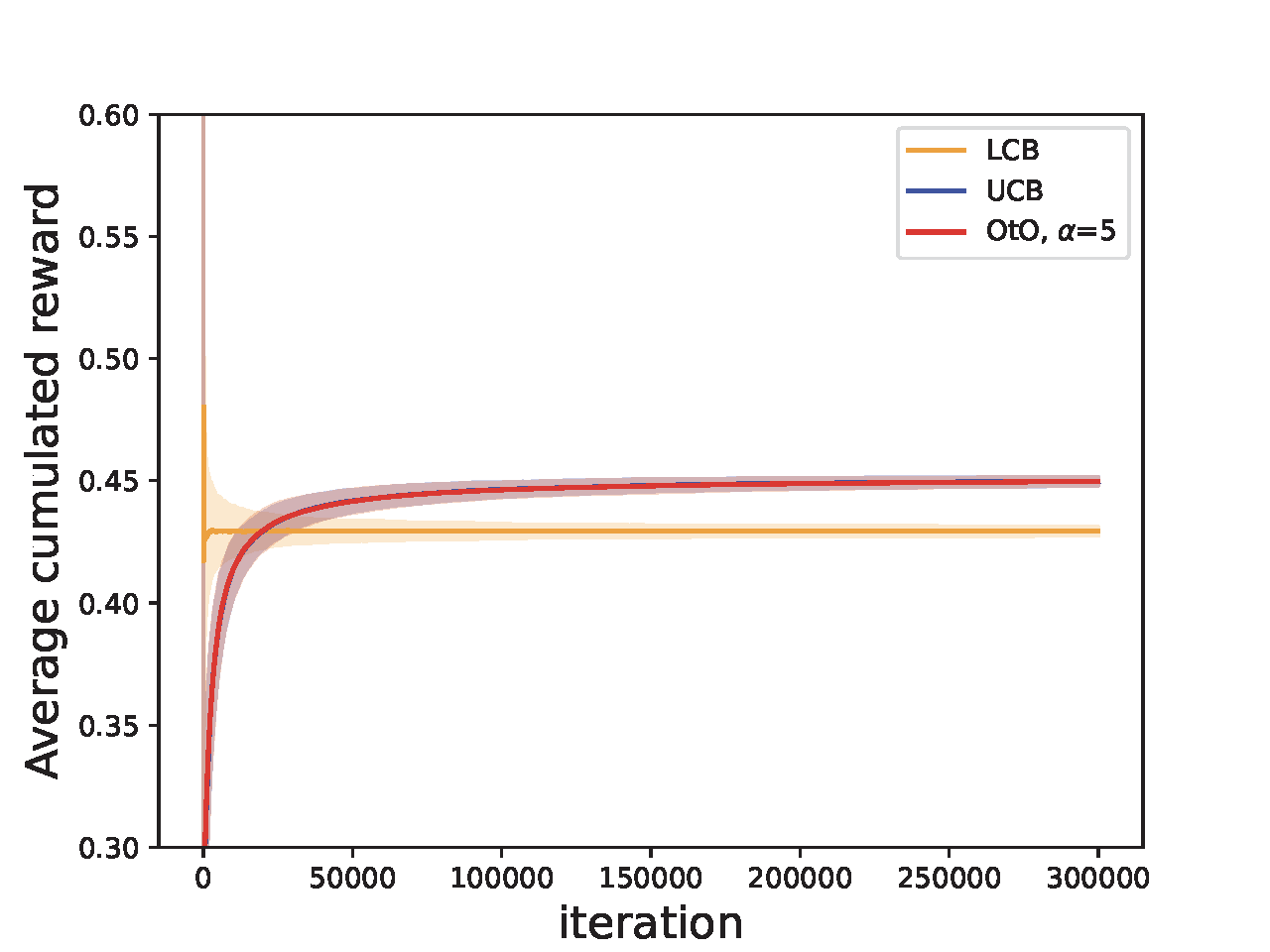}
\end{subfigure}

\caption{Cumulated reward of the three algorithms on the CTR prediction data in Setting 1, for various values of parameter $\alpha$, horizon $T=3000$ on the left, $T=300000$ on the right}

\label{fig:plotavcumrewardsetting1}
\end{centering}
\end{figure}

\begin{figure}[htb]
\begin{centering}
\begin{subfigure}{.48\textwidth}
  \centering
  \includegraphics[width=\linewidth]{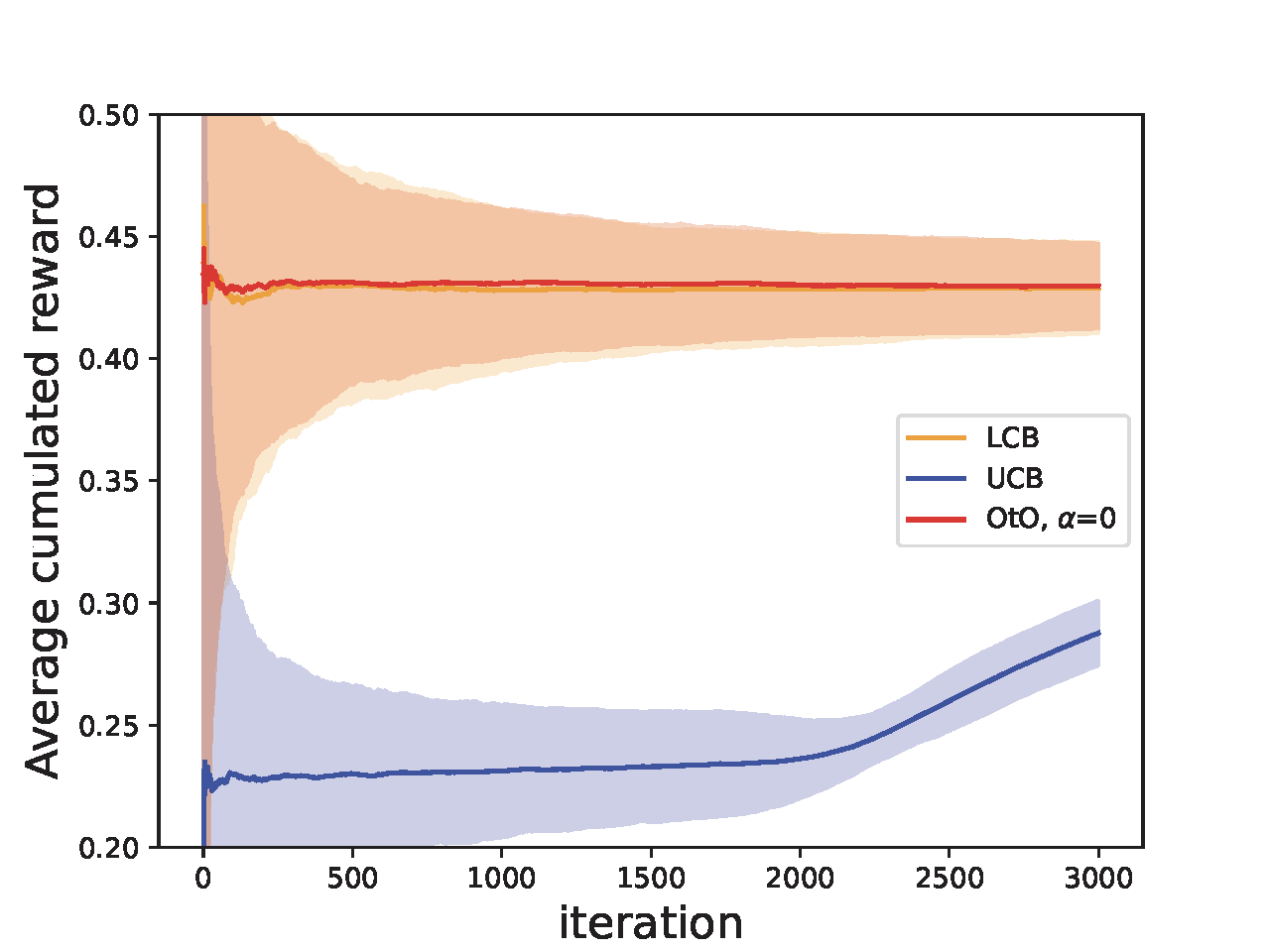}
\end{subfigure}%
\begin{subfigure}{.48\textwidth}
  \centering
  \includegraphics[width=\linewidth]{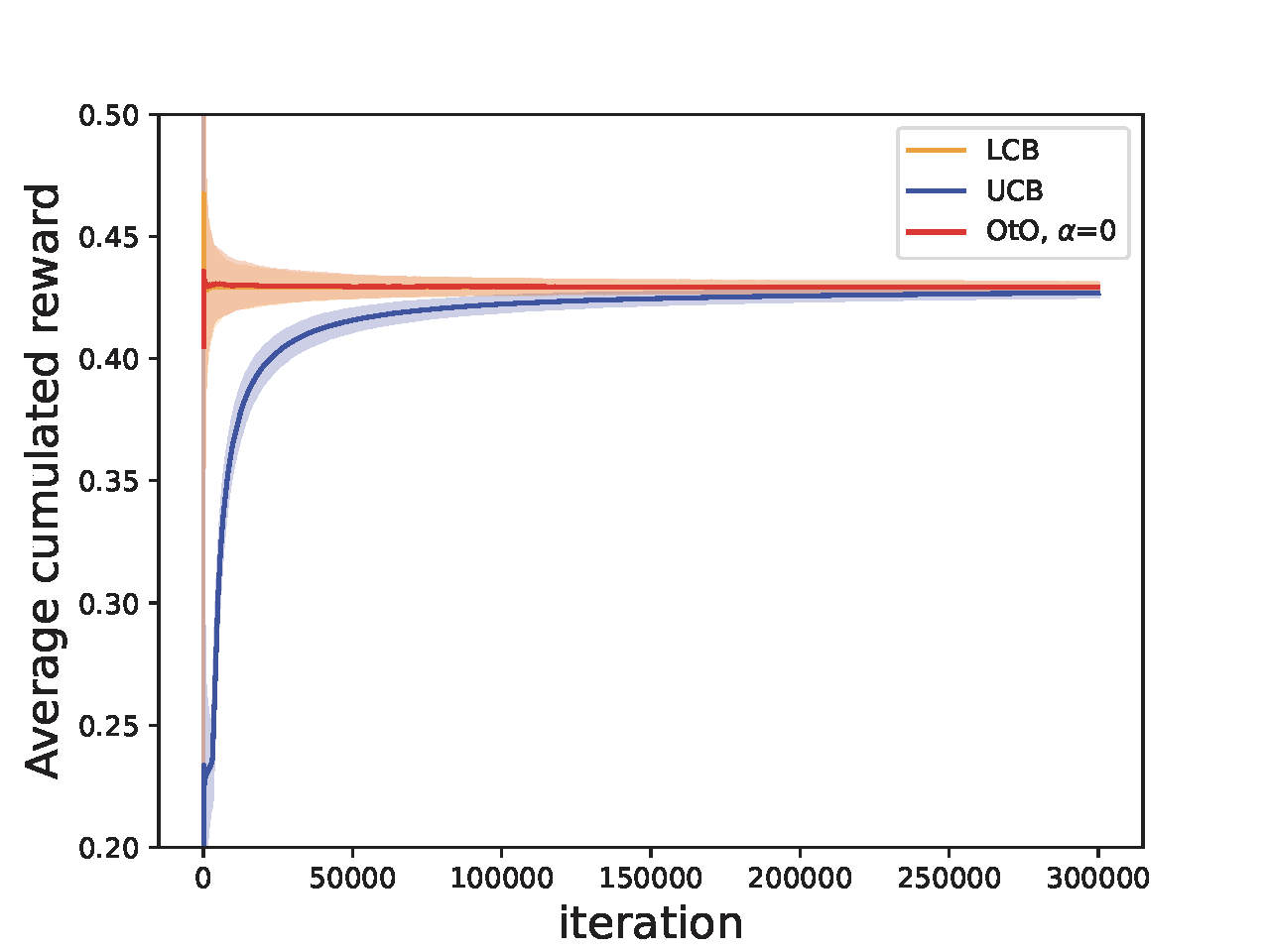}
\end{subfigure}
\begin{subfigure}{.48\textwidth}
  \centering
  \includegraphics[width=\linewidth]{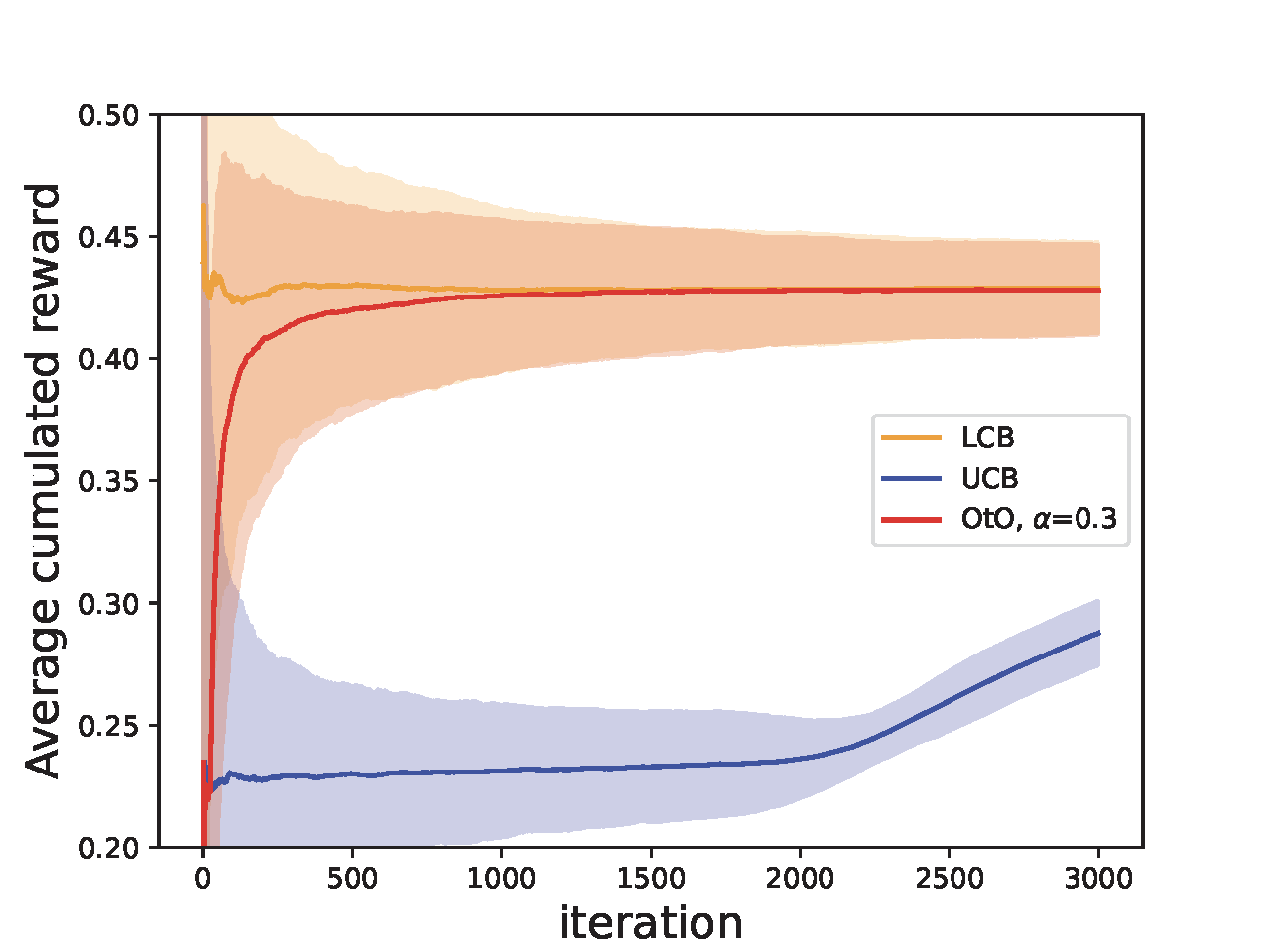}
\end{subfigure}%
\begin{subfigure}{.48\textwidth}
  \centering
  \includegraphics[width=\linewidth]{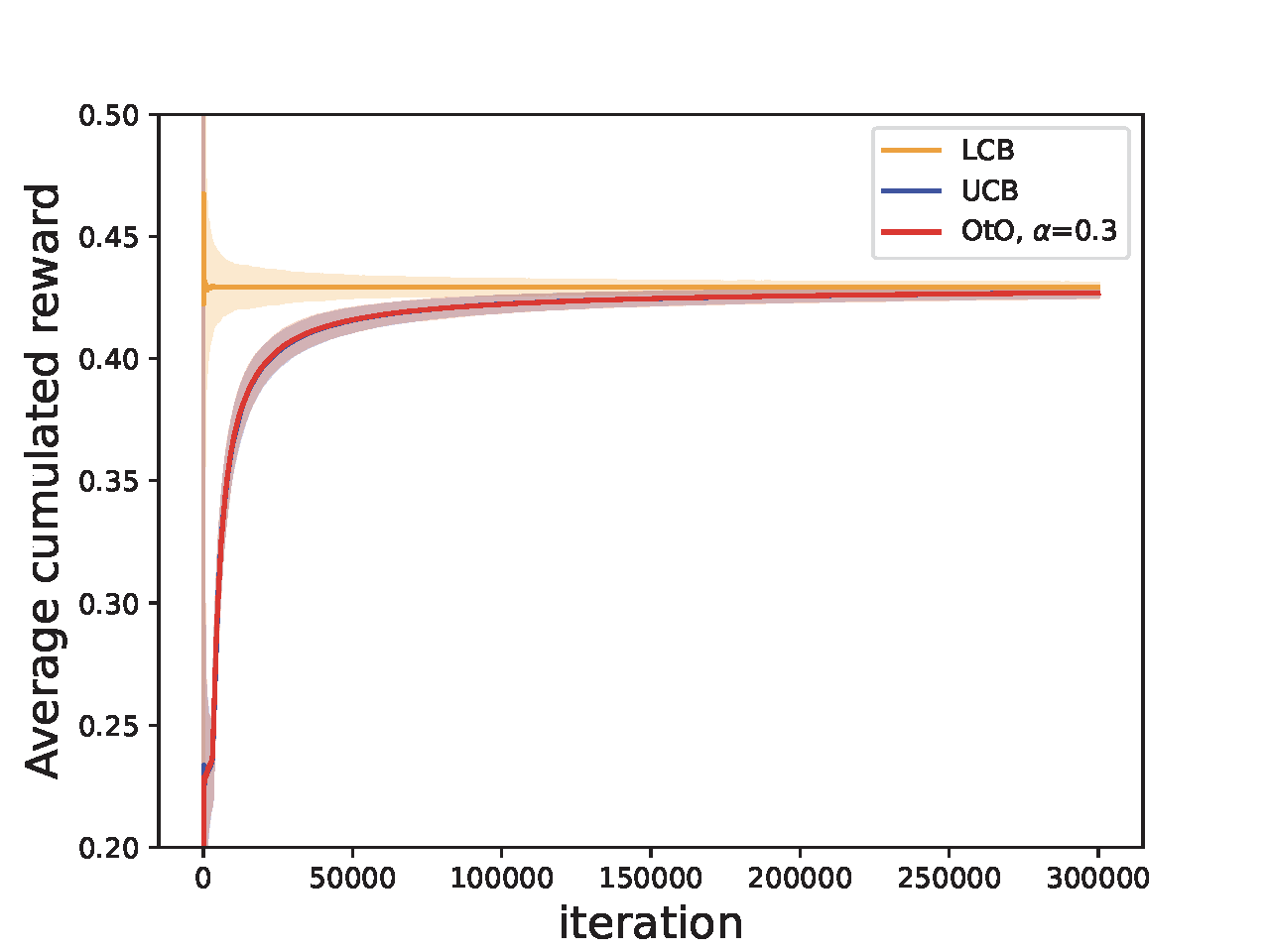}
\end{subfigure}
\begin{subfigure}{.48\textwidth}
  \centering
  \includegraphics[width=\linewidth]{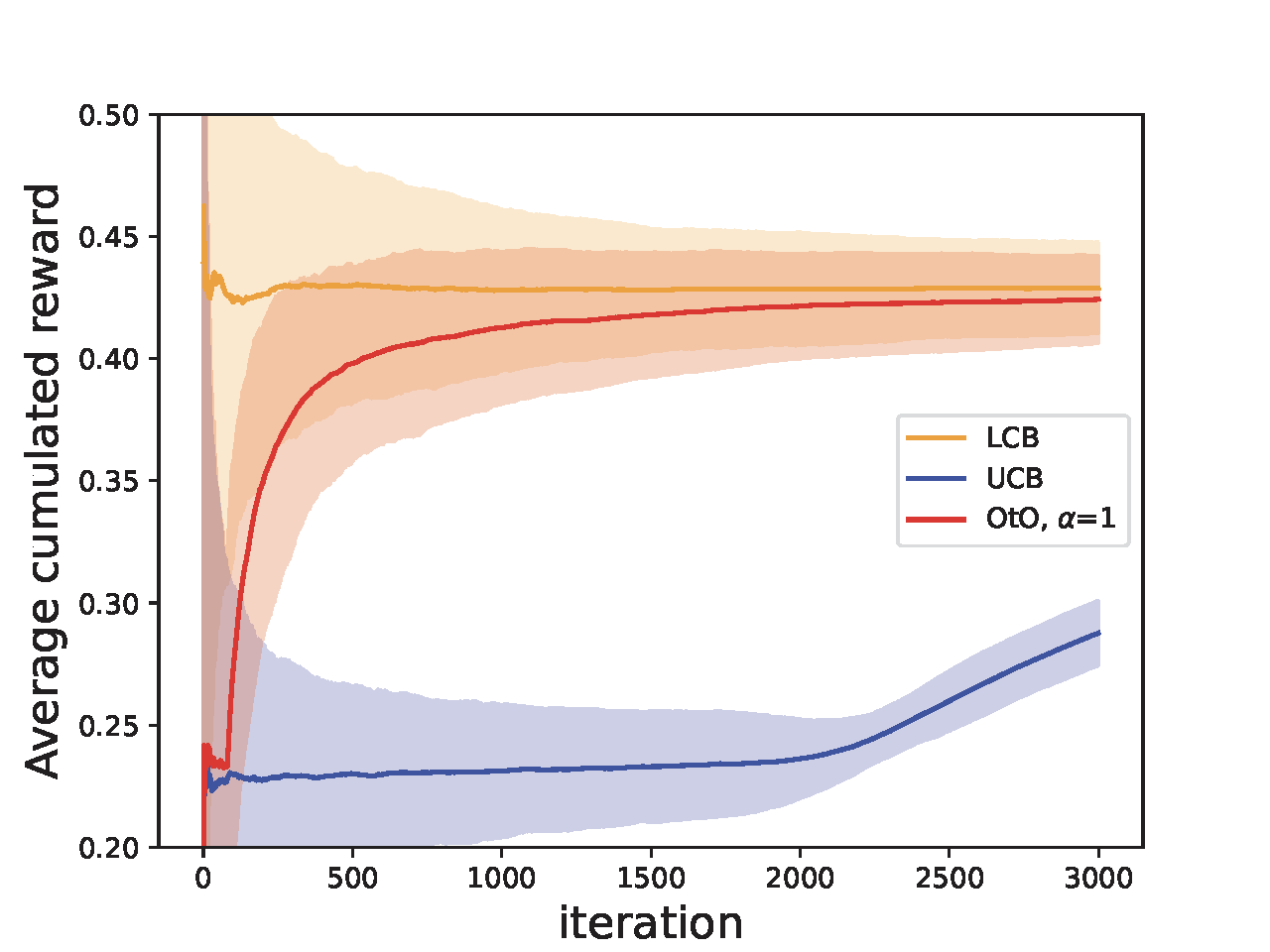}
\end{subfigure}%
\begin{subfigure}{.48\textwidth}
  \centering
  \includegraphics[width=\linewidth]{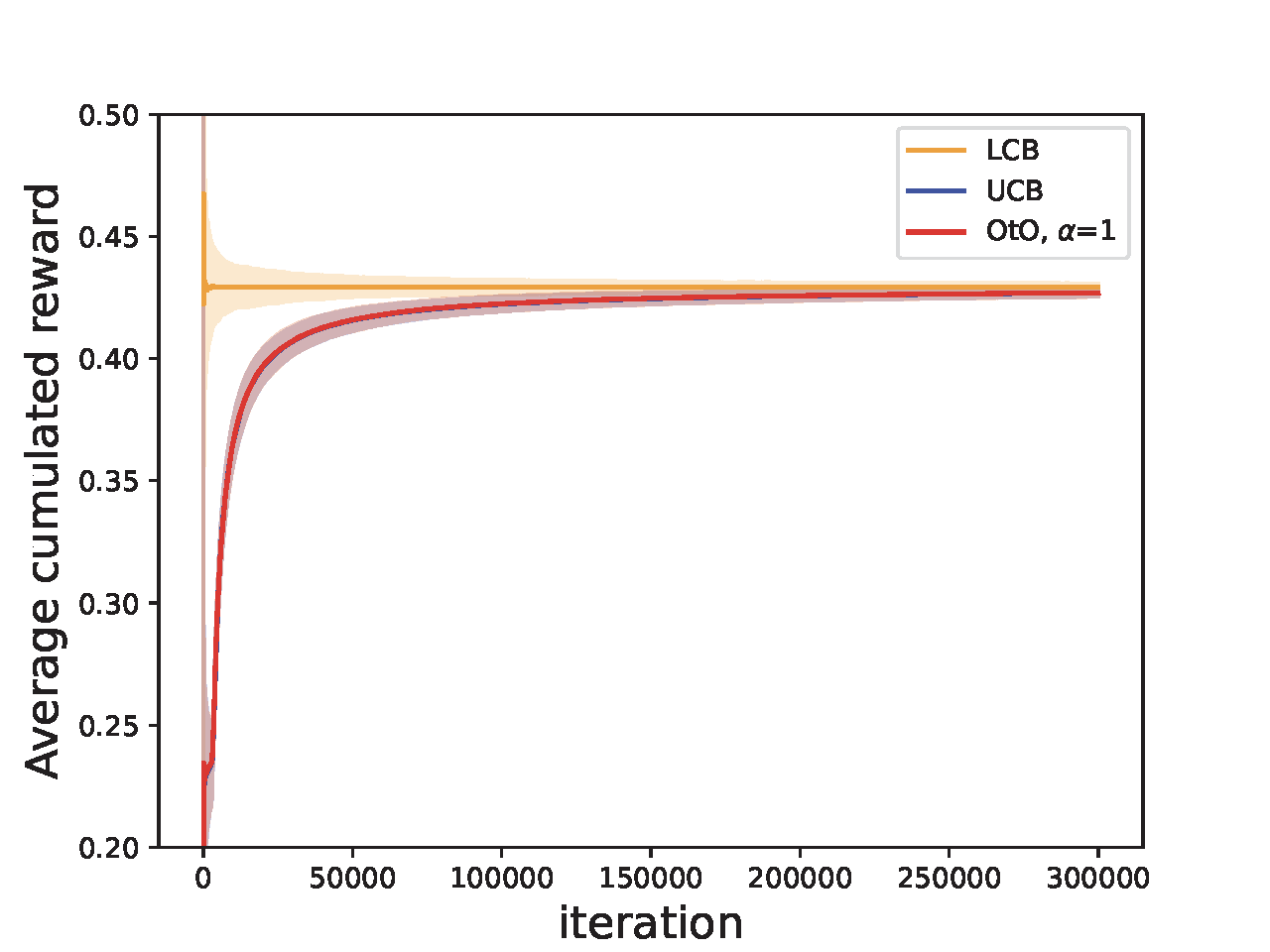}
\end{subfigure}
\begin{subfigure}{.48\textwidth}
  \centering
  \includegraphics[width=\linewidth]{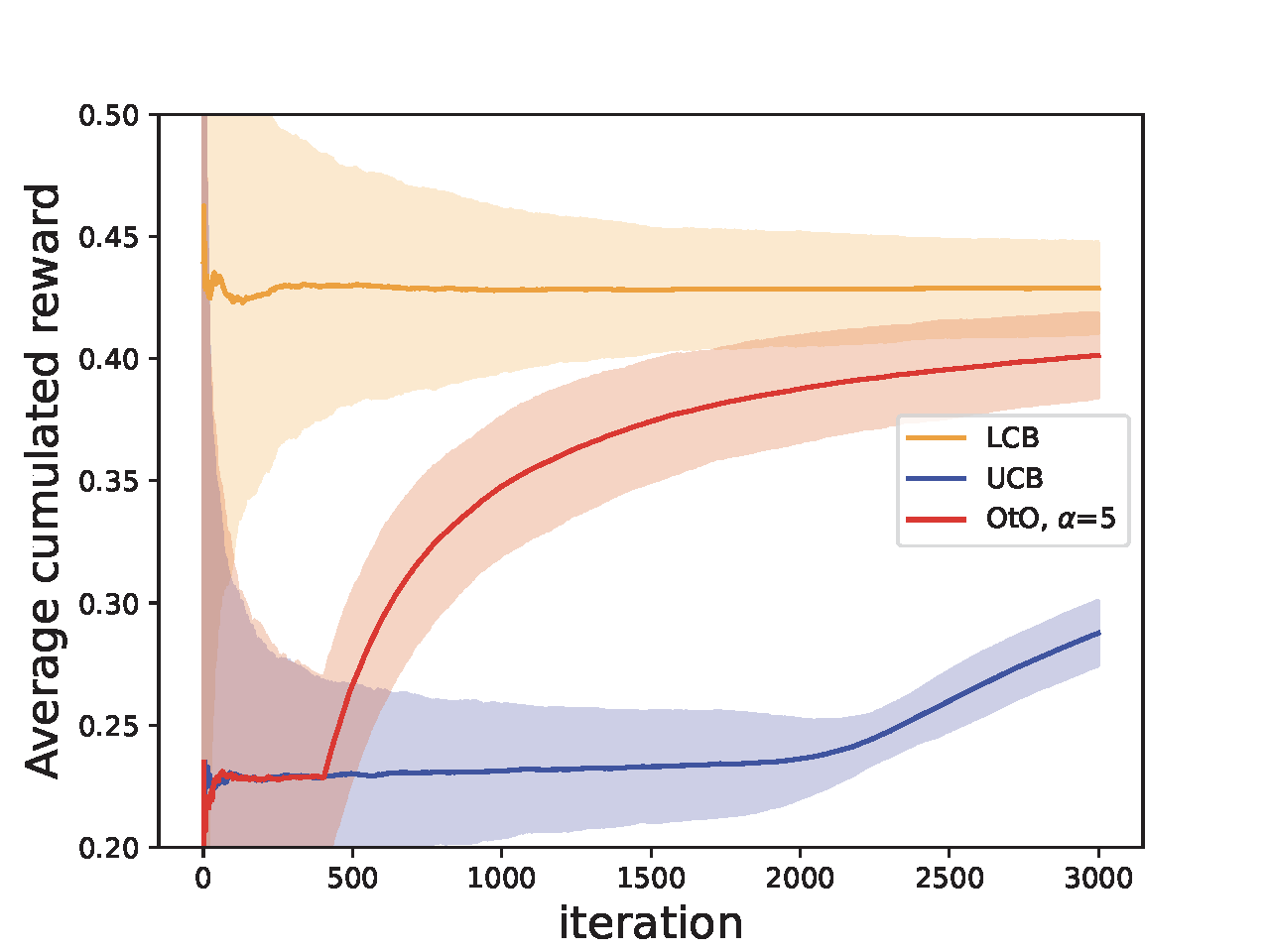}
\end{subfigure}%
\begin{subfigure}{.48\textwidth}
  \centering
  \includegraphics[width=\linewidth]{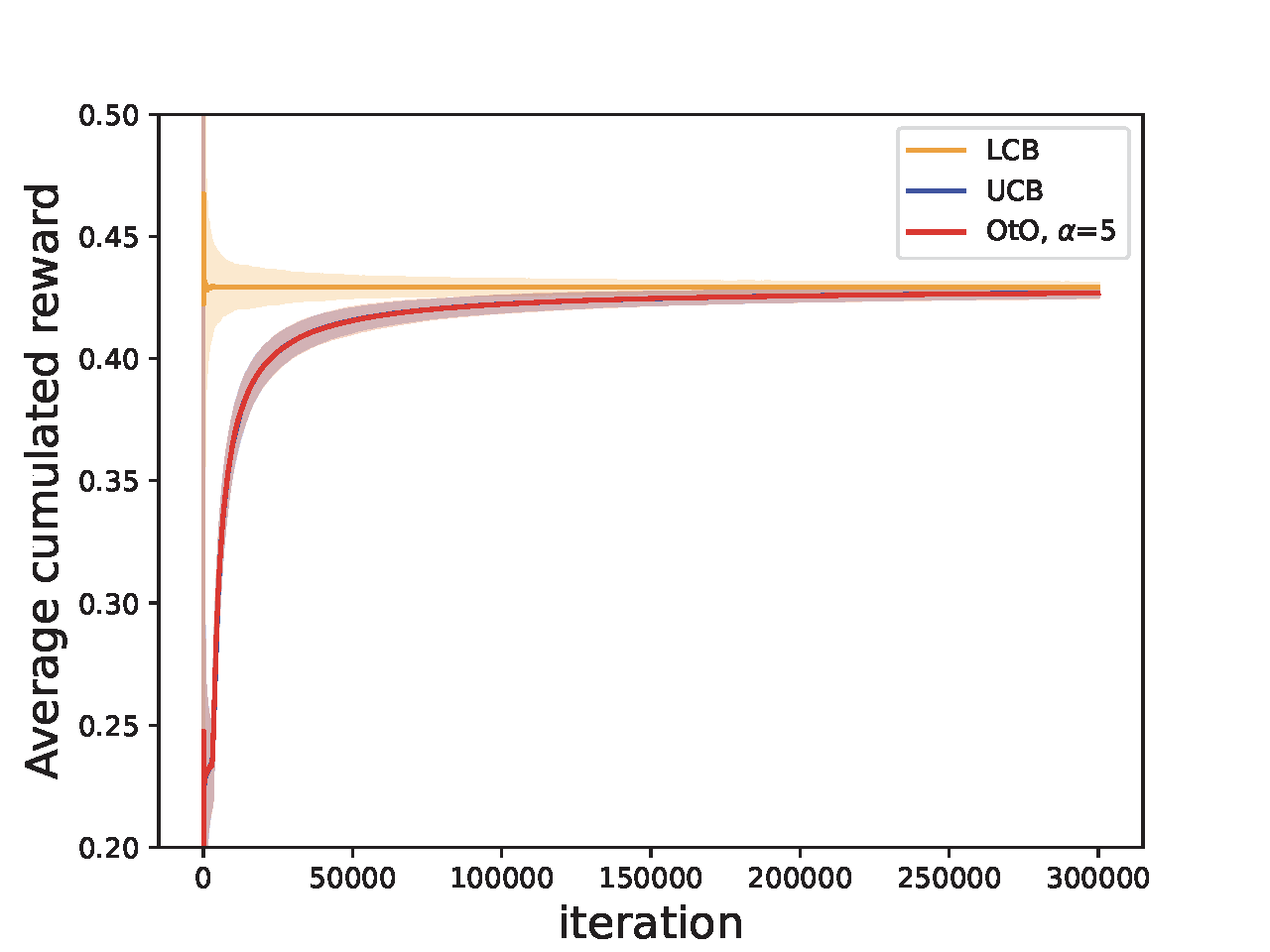}
\end{subfigure}

\caption{Cumulated reward of the three algorithms on the CTR prediction data in Setting 2, for various values of parameter $\alpha$, horizon $T=3000$ on the left, $T=300000$ on the right}
\label{fig:plotavcumrewardsetting2}

\end{centering}
\end{figure}

\section{Omitted Proofs}\label{sec:omittedproofs}

\input{proofsappendix}

\end{APPENDICES}

%% file: proofsappendix.tex
\subsection{Proof of \cref{prop:lowerboundminimax}}
 
 During this proof, we assume wlog that $\argmax_i m_i= m_1$. Fix any algorithm $\mathcal{A}$, and denote $\Delta>0$ a constant to be optimized later. We define a first instance $\theta_1$, where the reward of each arm $i\in [K]$ is a gaussian $\mathcal{N}(\mu_i,1)$, with $\mu_1=\Delta$ and $\mu_i=0$ for any $i>1$. Define
\[
j = \argmin_{i\neq 1} \mathbb{E}_{\theta_1,\mathcal{A}}[m_i+T_i(T-1)].
\]
Note that we have:
\[
\mathbb{E}_{\theta_1,\mathcal{A}}[m_j+T_j(T-1)] \leq \min_{J\subseteq [K]}\frac{T-1+\sum_{j \in J} m_j}{|J|}.
\]
Define alternative instance $\theta_2$ where the distribution of the reward of each arm is the same as in $\theta_1$ for all arm $i\neq j$, and the reward of arm $j$ is  $\mathcal{N}(2\Delta,1)$.  We have:
\[
\mathbb{P}_{\theta_1,\mathcal{A}}\left(R_\mathcal{A}(T)\geq \frac{T}{2}\Delta\right)\geq \mathbb{P}_{\theta_1, \mathcal{A}}\left(T_1(T)\leq \frac{T}{2}\right), 
\]
and
\begin{align*}
\mathbb{P}_{\theta_2,\mathcal{A}}\left(R_\mathcal{A}(T)\geq \frac{T}{2}\Delta\right)\geq& \mathbb{P}_{\theta_2, \mathcal{A}}\left(T_j(T)\leq \frac{T}{2}\right)\\
\geq& \mathbb{P}_{\theta_2, \mathcal{A}}\left(T_1(T)\geq \frac{T}{2}\right).
\end{align*}
By Bretagnolles-Huber:
\[
\mathbb{P}_{\theta_1, \mathcal{A}}\left(T_1(T)\leq \frac{T}{2}\right)+\mathbb{P}_{\theta_2, \mathcal{A}}\left(T_1(T)\geq \frac{T}{2}\right)\geq \frac{1}{2}\exp(-D(\mathbb{P}_{\theta_1,\mathcal{A}}, \mathbb{P}_{\theta_2,\mathcal{A}})).
\]

By the data processing inequality, we have:
\[
D(\mathbb{P}_{\theta_1,\mathcal{A}}, \mathbb{P}_{\theta_2,\mathcal{A}}) \leq \mathbb{E}_{\theta_1,\mathcal{A}}[m_j+T_j(T-1)] \frac{(2\Delta^2)}{2} .
\] 
Hence:
\[
D(\mathbb{P}_{\theta_1,\mathcal{A}}, \mathbb{P}_{\theta_2,\mathcal{A}})\leq \min_{J\subseteq [K]}\frac{T-1+\sum_{j \in J} m_j}{|J|}2\Delta^2.
\]
Also:
\begin{align*}
    \mathcal{R}_\mathcal{A}(T)\geq& \max_{\theta_1,\theta_2}\left(\mathbb{E}_{\theta_1,\mathcal{A}}\left[R(T)\right],\mathbb{E}_{\theta_2,\mathcal{A}}\left[R(T)\right]\right)\\
    \geq& \frac{T}{4}\Delta\left(\mathbb{P}_{\theta_1,\mathcal{A}}\left(R_\mathcal{A}(T)\geq \frac{T}{2}\Delta\right)+\mathbb{P}_{\theta_2,\mathcal{A}}\left(R_\mathcal{A}(T)\geq \frac{T}{2}\Delta\right)\right),
\end{align*}
where the second line uses inequality $\max(a,b)\geq \frac{1}{2}(a+b)$.
Putting everything together, this entails:
\[
\mathcal{R}_\mathcal{A}(T)\geq \frac{T}{8}\Delta\exp\left(-\min_{J\subseteq [K]}\frac{T-1+\sum_{j \in J} m_j}{|J|}2\Delta^2\right).
\]

Setting $\Delta= \sqrt{\max_{J\subseteq [K]} \frac{|J|}{2(T-1+\sum_{j \in J} m_j)}}$, we obtain:

\[
\mathcal{R}_\mathcal{A}(T)\geq \frac{T}{8}\sqrt{\max_{J\subseteq [K]} \frac{|J|}{2(T-1+\sum_{j \in J} m_j)}}\exp\left(-1\right).
\]
\hfill \(\Box\)

\subsection{Proof of \cref{prop:regretminimaxucb}}

 By \cref{lem:hoeff} and a union bound over the two inequalities, we have for all $t\leq T, i \in [K]$:
\[
\overline{\mu}_i(t)>\mu_i>\underline{\mu}_i(t).
\]

Denote $\mathcal{E}$ that event and assume in the following steps that it holds. If sub-optimal arm $i$ is pulled at iteration $t$, we have $\overline{\mu}_i(t)\geq \overline{\mu}_*(t)\geq \mu_*$, hence:
\begin{align}\label{eq:boundT_i(t)}
 \mu_i+\sqrt{\frac{2\log(K/\delta)}{T_i(t)+m_i}} \geq& \mu_*.
\end{align}
This gives the following bound on the number of pulls of sub-optimal arm $i$:
\begin{align}\label{eq:boundtotalnumpulls}
T_i(T) \leq \left( \frac{2}{\Delta_i^2} \log(K/\delta)-m_i\right)_++1.
\end{align}

Then, we have
\begin{align*}
    R(T)&=\mathbb{E}\left[\sum_i \Delta_i T_i(T)\right]\\
    &\leq \sum_i \Delta_i \mathbb{E}\left[T_i(T)|\mathcal{E}\right]+T\mathbb{P}\left(\bar{\mathcal{E}}\right)\\
    &\leq \underbrace{\sum_i \Delta_i \mathbb{E}\left[T_i(T)|\mathcal{E}\right]}_{(i)}+2T^2\delta\\
    &\leq\sum_i \Delta_i\left( \frac{2}{\Delta_i^2} \log(K/\delta)-m_i\right)_++\sum_i \Delta_i +2T^2\delta.
\end{align*}
The last line corresponds to the parameter-dependent upper bound. To get the minimax upper bound, we focus on bounding $(i)$. Define
\[
\Delta:= \max_{J\subseteq [K]}\sqrt{\frac{2|J|}{T+\sum_{j\in J}m_i}\log(K/\delta)}
\]
and 
\[
A:= \left\{ i \text{ s.t. }  \frac{2}{\Delta^2} \log(K/\delta)-m_i\geq 0\right\}.
\]
By \cref{eq:boundT_i(t)}, under event $\mathcal{E}$ an arm $i$ can only be pulled if $m_i \leq \frac{1}{\Delta_i^2}2\log(K/ \delta)$. If, in addition, that arm $i$ is such that $\Delta_i >\Delta$, then that arm can only be pulled if it belongs to ${A}$. Hence, combining with \cref{eq:boundtotalnumpulls}, we have:
\begin{align}\label{eq:interdeltai}
    (i)\leq& T\Delta+\sum_{\Delta_i>\Delta, i \in {A}}\Delta_i\left( \frac{2}{\Delta_i^2} \log(K/\delta)-m_i\right)_++\Delta_i.
\end{align}
Define now $J^*=\argmax_{J\subseteq [K]}\sqrt{\frac{2|J|}{T+\sum_{j\in J}m_j}\log(K/\delta)}$, breaking ties by selecting the element of maximum cardinality. Let us show that $A=J^*$. We start with $A\subseteq J^*$. Assume it is not the case, and we can find $i \in A \setminus J^*$. We have:

\[
m_i\leq \frac{T+\sum_{j\in J^*}m_j}{|J^*|}.
\]

Thus:
\begin{align*}
    \frac{T+\sum_{j\in J^*}m_j+m_i}{|J^*|+1}\leq &\frac{T+\sum_{j\in J^*}m_j+\frac{T+\sum_{j\in J^*}m_j}{|J^*|}}{|J^*|+1}\\
    =&\frac{T+\sum_{j\in J^*}m_j}{|J^*|},
\end{align*}
and this contradicts the optimality of $J^*$. Therefore $A\subseteq J^*$. On the other hand, assume we can find $i \in J^* \setminus A$. Then we have:
\[
m_i >\frac{T+\sum_{j\in J^*}m_j}{|J^*|}.
\]
Hence:
\begin{align*}
    \frac{T+\sum_{j\in J^*}m_j-m_i}{|J^*|-1}< &\frac{T+\sum_{j\in J^*}m_j-\frac{T+\sum_{j\in J^*}m_j}{|J^*|}}{|J^*|-1}\\
    =&\frac{T+\sum_{j\in J^*}m_j}{|J^*|},
\end{align*}
which again contradicts the optimality of $J^*$. Therefore we can conclude that $A=J^*$.

From this, we can combine with \cref{eq:interdeltai} and obtain:
\begin{align*}
    (i)\leq& \sum_{j \in J^*}\Delta_j\left( \frac{2}{\Delta_j^2} \log(K/\delta)-m_j\right)_++T\Delta+|J^*|\\
    \leq& \sum_{j \in J^*} \frac{2}{\Delta} \log(K/\delta)-\sum_{j \in J^*}m_j \Delta+T\Delta+|J^*|\\
    \leq& \sqrt{2|J^*|(T+\sum_{j\in J^*}m_j)\log(K/\delta)}-\sum_{j \in J^*}m_j\sqrt{\frac{2|J^*|}{T+\sum_{j\in J^*}m_j}\log(K/\delta)}  \\
    &+T\sqrt{\frac{2|J^*|}{T+\sum_{j\in J^*}m_j}\log(K/\delta)}+|J^*|\\
    =&\sqrt{2|J^*|\log(K/\delta)}\left(\sqrt{T+\sum_{j\in J^*}m_j}-\frac{\sum_{j\in J^*}m_j}{\sqrt{T+\sum_{j\in J^*}m_j}}\right)\\
    &+T\sqrt{\frac{2|J^*|}{T+\sum_{j\in J^*}m_j}\log(K/\delta)}+|J^*|\\
    =&2T\sqrt{\frac{2|J^*|}{T+\sum_{j\in J^*}m_j}\log(K/\delta)}+|J^*|.
\end{align*}
Also, if $\min_{i \in [K]}m_i>0$ define 
\[
\Delta':= \sqrt{\frac{2}{\min_{i \in [K]}m_i}\log(K/\delta)}.
\]

By \cref{eq:boundT_i(t)}, an arm $i$ can be sampled at least once only if $\Delta_i\leq \Delta'$. Hence, we also have the bound:
\[
(i)\leq T \Delta'.
\]
Putting the bounds together, we obtain:
\[
\mathcal{R}_{\algucb}(T)\leq \min\left(\sqrt{\frac{2}{\min_{i \in [K]}m_i}\log(K/\delta)};\max_{J\subseteq [K]}2T\sqrt{\frac{2|J|}{T+\sum_{j\in J}m_j}\log(K/\delta)}+|J|\right)+2T^2 \delta.
\]

\hfill \(\Box\)

\subsection{Proof of \cref{prop:minimaxregretlcb}}

Let us start by proving the upper bound. By \cref{lem:hoeff} and a union bound over the two bounds, with probability at least $1-2T\delta$, we have for all $t\leq T, i \in [K]$:
\[
\overline{\mu}_i(t)>\mu_i>\underline{\mu}_i(t).
\]
In this case, for all $t\leq T$, the following holds:
\begin{align*}
 \mu_{L(t)}\geq & \underline{\mu}_{L(t)}(t)\\
 \geq& \underline{\mu}_{*}(t)\\
 \geq& \mu_*-\sqrt{\frac{2\log(K/\delta)}{(m_*+T_*(t))}}\\
 \geq& \mu_*-\sqrt{\frac{2\log(K/\delta)}{\min_i m_i}}.
\end{align*}
Summing over all $T$ iterations and taking the expectation gives:
\[
 \mathcal{R}_\alglcb(T) \leq T\sqrt{\frac{2\log(K/\delta)}{\min_i m_i}}+ 2T^2\delta.
\]

We now turn to the proof of the lower bound. First if $\min_i m_i =0$, the result is obtained by setting the mean of arm(s) $i$ with $m_i=0$ to $1$ and the reward of all other arms to $0$. Then the best arm is never selected and   $R(T)=T$.

Assume now that $\min_i m_i >0$ and  wlog that $\argmin_i m_i=1$. Set the reward of arm $1$ to be $\mathcal{N}\left(\min\left(1;0.5+\sqrt{\frac{1}{m_1}}\right),1\right)$, the reward of arm $2$ to be $0.5$ deterministically and the reward of all the other arms to be $0$ deterministically. Note that if $\hat{\mu}_1(0) < \hat{\mu}_2(0)$, then \alglcb will pull arm $2$ at every iteration. Indeed, this arm has the highest lower bound in the first iteration. Also, the empirical mean of arm $1$ and $2$ remain unchanged unless arm $1$ is pulled. Then, by induction, arm $2$ will have the highest lower bound at every iteration. We can lower bound the probability that this event will happen:
\begin{align*}
    \mathbb{P}(\hat{\mu}_1(0) < \hat{\mu}_2(0)) &= \mathbb{P}(\hat{\mu}_1(0) \leq 0.5) \\
    &= \mathbb{P}(X\leq 0), \text{ where } X\sim \mathcal{N}\left(\min\left(1;0.5+\sqrt{\frac{1}{m_1}}\right),\frac{1}{\sqrt{m_1}}\right)\\
    &\geq 0.15.
\end{align*}

Hence:

\begin{equation*}
\mathcal{R}_{\textsc{LCB}}(T)\geq 0.15 T\min\left(0.5;\sqrt{\frac{1}{\min_i m_i}}\right).
\end{equation*}
\hfill \(\Box\)

\subsection{Proof of \cref{prop:loginucbT=1}}

 Let us start with the upper bound. We have:
\begin{align*}
    R_\algucb^{\text{log}}(1)\leq R_\algucb(1).
\end{align*}

Thus, by \cref{prop:regretminimaxucb}:
\begin{align*}
    R_\algucb^{\text{log}}(1)\leq \sqrt{\frac{2}{\min_i m_i}\log\left(\frac{K}{\delta}\right)}+2\delta.
\end{align*}

For the lower bound, assume wlog that $m_1=\min_i m_i$. If $m_1=0$, then for all arms $i$ s.t. $m_i=0$, set $\mu_i=0$, and for all other arms $i$ with $m_i>0$, set $\mu_i=1$. In that setting, \algucb\ pulls one of the arms $i$ with $m_i=0$ and $R^{\text{log}}(1)=1$. Assume now $m_1>0$.  Set the reward for arm $1$ to be $\mathcal{N}(0,1)$, and set the reward of all the other arms to be $\min \left(1;\sqrt{\frac{1}{m_1}}\right)$ deterministically. We have:

\[
\mu_0=\frac{\sum_i m_i \mu_i}{\sum_i m_i}\geq \min \left(1;\sqrt{\frac{1}{m_1}}\right) -\frac{1}{K} \min \left(1;\sqrt{\frac{1}{m_1}}\right) \geq 0.5\min \left(1;\sqrt{\frac{1}{m_1}}\right).
\]
Hence:
\[
  R_\algucb^{\text{log}}(1)\geq \mathbb{P}\left(I(1)=1\right) 0.5 \min \left(1;\sqrt{\frac{1}{m_1}}\right).
\]

On the other hand:
\begin{align*}
    \mathbb{P}\left(I(1)=1\right)\geq &\mathbb{P}\left(\hat{\mu}_1\geq \sqrt{\frac{1}{m_1}}\right)\\
    \geq & 0.15.
\end{align*}
This implies:
\[
  R^{\text{log}}_\algucb(1)\geq 0.07\min \left(1;\sqrt{\frac{1}{\min_i m_i}}\right).
\]
\hfill \(\Box\)

\subsection{Proof of \cref{prop:lowerboundUCBanyT}}

 Assume wlog that $m_1\geq \ldots \geq m_K$. The reward for all the arms in the constructed lower bound instance is deterministic.  Set $\mu_1=1$ and for all $i \in [2;K]$:
\[
\mu_i = 1-\underbrace{\left(\sqrt{\frac{\log(K/\delta)}{2(m_i+\frac{T}{K})}}-\sqrt{\frac{\log(K/\delta)}{2(m_1+\frac{T}{K})}}\right)}_{=\Delta_i}.
\]
Note that the formula also holds for arm $1$, as it gives $\Delta_1=0$. We will show  by contradiction that for any arm $i \in [K]$, we have $T_i(T)\leq \lceil\frac{T}{K}\rceil$. Assume $\exists i \in [K]$ s.t. $T_i(T)>\lceil\frac{T}{K}\rceil$. This implies the existence of another arm $j\neq i$ s.t. $T_j(T)<\frac{T}{K}$, as well as an iteration $t$ where $T_i(t)= \lceil\frac{T}{K}\rceil$ and
\[
\overline{\mu}_i(t)\geq\overline{\mu}_j(t), 
\]
hence:
\begin{align*}
-\Delta_i+\sqrt{\frac{\log(K/\delta)}{2(m_i+\frac{T}{K})}}\geq& -\Delta_j+\sqrt{\frac{\log(K/\delta)}{2(m_j+T_j(t))}}\\
\geq& -\Delta_j+\sqrt{\frac{\log(K/\delta)}{2(m_j+\frac{T}{K}-1)}}
\end{align*}
Injecting the formulas for the gaps and simplifying, we get:
\begin{align*}
\sqrt{\frac{\log(K/\delta)}{m_1+\frac{T}{K}}}\geq& \sqrt{\frac{\log(K/\delta)}{m_1+\frac{T}{K}}}-\sqrt{\frac{\log(K/\delta)}{m_j+\frac{T}{K}}}+\sqrt{\frac{\log(K/\delta)}{m_j+\frac{T}{K}-1}}.
\end{align*}
Hence:
\begin{align*}
\sqrt{\frac{\log(K/\delta)}{m_j+\frac{T}{K}}}\geq \sqrt{\frac{\log(K/\delta)}{m_j+\frac{T}{K}-1}},
\end{align*}
which is impossible. We thus obtain $T_i(T)\leq \lceil\frac{T}{K}\rceil$ for any $i\in [K]$. As on the other hand, we have $\sum_i T_i(T)=T$, we obtain
\[
T_i(T)\in \left[\bigg\lfloor\frac{T}{K}\bigg \rfloor;\bigg \lceil\frac{T}{K}\bigg \rceil\right] \quad \forall i \in [K].
\]
Thus, when $\frac{T}{K}$ is an integer, on that instance we have:
\begin{align*}
R^{\text{log}}_{\textsc{UCB}}(T)=&T\frac{1}{m}\sum_{i=1}^Km_i \mu_i- \sum_{i=1}^K\frac{T}{K} \mu_i \\
=&\sum_{i=1}^K\frac{T}{K} \Delta_i -T\frac{1}{m}\sum_{i=1}^Km_i \Delta_i\\
=&T\sum_{i=1}^K\left(\frac{1}{K}-\frac{m_i}{m}\right)\left[\sqrt{\frac{1}{2(m_i+\frac{T}{K})}}-\sqrt{\frac{1}{2(m_1+\frac{T}{K})}}\right].
\end{align*}
\hfill \(\Box\)

\subsection{Proof of \cref{prop:regretlogginglcb}}

Let us start with the upper bound. We have by definition of \textsc{LCB}:
\begin{align*}
    \underline{\mu}_{L(t)}(t)\geq \sum_{i=1}^K\frac{m_i \underline{\mu}_{i}(t)}{m}.
\end{align*}
By \cref{lem:hoeff} and a union bound, with probability at least $1-2T\delta$ this entails, 
\begin{align*}
    \mu_{L(t)}\geq &\frac{1}{\sum_i m_i}\sum_i m_i\left(\mu_i-\sqrt{2\frac{\log(\frac{K}{\delta})}{m_i+T_i(t)}}\right)\\
   \geq & \frac{1}{\sum_i m_i}\sum_i m_i\left(\mu_i-\sqrt{2\frac{\log(\frac{K}{\delta})}{m_i}}\right)\\
\end{align*}
Hence:
\begin{align}\label{eq:boundlcbaux}
    \mu_0-\mu_{L(t)}\leq \frac{\sum_i \sqrt{m_i}}{\sum_i m_i}\sqrt{2\log(\frac{K}{\delta})}.
\end{align}
Summing over all possible iterations and taking the expectation, this implies:
\begin{align*}
    \mathcal{R}_\alglcb^{\text{log}}(T)\leq &T\frac{\sum_i \sqrt{m_i}}{\sum_i m_i}\sqrt{2\log(\frac{K}{\delta})}+2T^2\delta.
\end{align*}

We now turn to the lower bound. First, assume wlog that $m_1\geq m_2\geq \ldots\_eq m_K$. We will distinguish two cases depending on whether $m_2=0$ or not. 

If $m_2=0$, then there is only samples for arm $1$ at the first iteration, hence, $\alglcb(t)=1$. By induction, this remains true at every iteration, and $\mathcal{R}^{\text{log}}(T)=0$.

Assume now that $m_2>0$. Set the reward for arm $2$ to be $\mathcal{N}(0.5+\sqrt{\frac{1}{m_2}},1)$, and set the reward of all the other arms to be $0.5$ deterministically. By definition, for any $t\leq T$:
\begin{align*}
\underline{\mu_1}(t)\geq &\underline{\mu_1}(0)\\ =&
0.5-\sqrt{\frac{\log(K/\delta)}{2m_1}}.
\end{align*}
This implies that if 
  $ \underline{\mu_2}(t)< \underline{\mu_1}(0)$, then   $ \underline{\mu_2}(t)< \underline{\mu_1}(t)$. Moreover, if $   \underline{\mu_2}(t)< \underline{\mu_1}(t)$ holds for some $t\leq T$, then arm $2$ is not chosen, which implies   $ \underline{\mu_2}(t+1)< \underline{\mu_1}(t+1)$. By induction, this gives:
\begin{align*}
\mathbb{P}\left(L(t) \neq 2 \text{ for all } t\leq T\right) \geq&\mathbb{P}\left(   \underline{\mu_2}(t)< \underline{\mu_1}(t) \text{ for all } t\leq T\right) \\
    \geq &\mathbb{P}\left(   \underline{\mu_2}(t)< \underline{\mu_1}(t) \right) \\
    =&\mathbb{P}\left(\hat{\mu}_2(0)< 0.5\right)\\
    \geq& 0.15. 
\end{align*}

On the other hand
\[
\frac{\sum_i m_i \mu_i}{\sum_i m_i}= 0.5 +\frac{\sqrt{m_2}}{m}.
\]
This implies:
\begin{align*}
  R_\alglcb^{\text{log}}(T)\geq& 0.15 T\frac{\sqrt{m_2}}{m}.
\end{align*}

\hfill \(\Box\)

%% file: main.bbl
\begin{thebibliography}{}

\bibitem[Agrawal and Goyal, 2012]{agrawal2012analysis}
Agrawal, S. and Goyal, N. (2012).
\newblock Analysis of thompson sampling for the multi-armed bandit problem.
\newblock In {\em Conference on learning theory}, pages 39--1. JMLR Workshop
  and Conference Proceedings.

\bibitem[Auer and Ortner, 2010]{auer2010ucb}
Auer, P. and Ortner, R. (2010).
\newblock Ucb revisited: Improved regret bounds for the stochastic multi-armed
  bandit problem.
\newblock {\em Periodica Mathematica Hungarica}, 61(1-2):55--65.

\bibitem[Ball et~al., 2023]{ball2023efficient}
Ball, P.~J., Smith, L., Kostrikov, I., and Levine, S. (2023).
\newblock Efficient online reinforcement learning with offline data.
\newblock In {\em International Conference on Machine Learning}, pages
  1577--1594. PMLR.

\bibitem[Bastani and Bayati, 2020]{bastani2020online}
Bastani, H. and Bayati, M. (2020).
\newblock Online decision making with high-dimensional covariates.
\newblock {\em Operations Research}, 68(1):276--294.

\bibitem[Ben-Tal and Nemirovski, 2002]{ben2002robust}
Ben-Tal, A. and Nemirovski, A. (2002).
\newblock Robust optimization--methodology and applications.
\newblock {\em Mathematical programming}, 92:453--480.

\bibitem[Bertsimas et~al., 2011]{bertsimas2011theory}
Bertsimas, D., Brown, D.~B., and Caramanis, C. (2011).
\newblock Theory and applications of robust optimization.
\newblock {\em SIAM review}, 53(3):464--501.

\bibitem[Bertsimas and Thiele, 2006]{bertsimas2006robust}
Bertsimas, D. and Thiele, A. (2006).
\newblock A robust optimization approach to inventory theory.
\newblock {\em Operations research}, 54(1):150--168.

\bibitem[Bu et~al., 2023]{bu2023offline}
Bu, J., Simchi-Levi, D., and Wang, L. (2023).
\newblock Offline pricing and demand learning with censored data.
\newblock {\em Management Science}, 69(2):885--903.

\bibitem[Bu et~al., 2022]{bu2021onlinepricingofflinedata}
Bu, J., Simchi-Levi, D., and Xu, Y. (2022).
\newblock Online pricing with offline data: Phase transition and inverse square
  law.
\newblock {\em Management Science}, 68(12):8568--8588.

\bibitem[Buckman et~al.,
  2020]{buckman2020importancepessimismfixeddatasetpolicy}
Buckman, J., Gelada, C., and Bellemare, M.~G. (2020).
\newblock The importance of pessimism in fixed-dataset policy optimization.

\bibitem[Cai, 2024]{kaggle_deepctr_difm}
Cai, B. (2024).
\newblock Deepctr difm: Demonstrating deepctr with difm model on kaggle.
\newblock Accessed: 2024-11-18.

\bibitem[Caro and Gallien, 2007]{caro2007dynamic}
Caro, F. and Gallien, J. (2007).
\newblock Dynamic assortment with demand learning for seasonal consumer goods.
\newblock {\em Management science}, 53(2):276--292.

\bibitem[Chen et~al., 2022]{chen2022data}
Chen, X., Shi, P., and Pu, S. (2022).
\newblock Data-pooling reinforcement learning for personalized healthcare
  intervention.
\newblock {\em arXiv preprint arXiv:2211.08998}.

\bibitem[Cheng et~al., 2022]{cheng2022adversarially}
Cheng, C.-A., Xie, T., Jiang, N., and Agarwal, A. (2022).
\newblock Adversarially trained actor critic for offline reinforcement
  learning.
\newblock In {\em International Conference on Machine Learning}, pages
  3852--3878. PMLR.

\bibitem[Cheung and Lyu, 2024]{pmlr-v235-cheung24a}
Cheung, W.~C. and Lyu, L. (2024).
\newblock Leveraging ({B}iased) information: Multi-armed bandits with offline
  data.
\newblock In Salakhutdinov, R., Kolter, Z., Heller, K., Weller, A., Oliver, N.,
  Scarlett, J., and Berkenkamp, F., editors, {\em Proceedings of the 41st
  International Conference on Machine Learning}, volume 235 of {\em Proceedings
  of Machine Learning Research}, pages 8286--8309. PMLR.

\bibitem[Fujimoto and Gu, 2021]{fujimoto2021minimalist}
Fujimoto, S. and Gu, S.~S. (2021).
\newblock A minimalist approach to offline reinforcement learning.
\newblock {\em Advances in neural information processing systems},
  34:20132--20145.

\bibitem[Fujimoto et~al., 2018]{fujimoto2019offpolicydeepreinforcementlearning}
Fujimoto, S., Meger, D., and Precup, D. (2018).
\newblock Off-policy deep reinforcement learning without exploration.
\newblock In {\em International Conference on Machine Learning}.

\bibitem[Guo et~al., 2017]{guo2017deepfmfactorizationmachinebasedneural}
Guo, H., Tang, R., Ye, Y., Li, Z., and He, X. (2017).
\newblock Deepfm: a factorization-machine based neural network for ctr
  prediction.
\newblock In {\em Proceedings of the 26th International Joint Conference on
  Artificial Intelligence}, IJCAI'17, page 1725–1731. AAAI Press.

\bibitem[Gur and Momeni, 2022]{gur2020adaptive}
Gur, Y. and Momeni, A. (2022).
\newblock Adaptive sequential experiments with unknown information arrival
  processes.
\newblock {\em Manufacturing \& Service Operations Management},
  24(5):2666--2684.

\bibitem[Jin et~al., 2022]{jin2022policy}
Jin, Y., Ren, Z., Yang, Z., and Wang, Z. (2022).
\newblock Policy learning" without''overlap: Pessimism and generalized
  empirical bernstein's inequality.
\newblock {\em arXiv preprint arXiv:2212.09900}.

\bibitem[Kidambi et~al., 2020a]{kidambi2020morel}
Kidambi, R., Rajeswaran, A., Netrapalli, P., and Joachims, T. (2020a).
\newblock Morel: Model-based offline reinforcement learning.
\newblock {\em Advances in neural information processing systems},
  33:21810--21823.

\bibitem[Kidambi et~al., 2020b]{MorelPessimistisOfflineLearning}
Kidambi, R., Rajeswaran, A., Netrapalli, P., and Joachims, T. (2020b).
\newblock Morel: Model-based offline reinforcement learning.
\newblock In Larochelle, H., Ranzato, M., Hadsell, R., Balcan, M., and Lin, H.,
  editors, {\em Advances in Neural Information Processing Systems}, volume~33,
  pages 21810--21823. Curran Associates, Inc.

\bibitem[Lattimore, 2016]{lattimore2016regretanalysisanytimeoptimally}
Lattimore, T. (2016).
\newblock Regret analysis of the anytime optimally confident ucb algorithm.

\bibitem[Lattimore and Szepesv{\'a}ri, 2020]{lattimore2020bandit}
Lattimore, T. and Szepesv{\'a}ri, C. (2020).
\newblock {\em Bandit algorithms}.
\newblock Cambridge University Press.

\bibitem[Lee et~al., 2021]{lee2021offlinetoonlinereinforcementlearningbalanced}
Lee, S., Seo, Y., Lee, K., Abbeel, P., and Shin, J. (2021).
\newblock Offline-to-online reinforcement learning via balanced replay and
  pessimistic q-ensemble.
\newblock In {\em Conference on Robot Learning}.

\bibitem[Li et~al., 2024a]{li2022pessimismofflinelinearcontextual}
Li, G., Ma, C., and Srebro, N. (2024a).
\newblock Pessimism for offline linear contextual bandits using lp confidence
  sets.
\newblock In {\em Proceedings of the 36th International Conference on Neural
  Information Processing Systems}, NIPS '22, Red Hook, NY, USA. Curran
  Associates Inc.

\bibitem[Li et~al., 2023]{Li2023-lv}
Li, G., Zhan, W., Lee, J.~D., Chi, Y., and Chen, Y. (2023).
\newblock Reward-agnostic fine-tuning: Provable statistical benefits of hybrid
  reinforcement learning.
\newblock {\em arXiv preprint arXiv:2305.10282}.

\bibitem[Li et~al., 2024b]{li2024reward}
Li, G., Zhan, W., Lee, J.~D., Chi, Y., and Chen, Y. (2024b).
\newblock Reward-agnostic fine-tuning: Provable statistical benefits of hybrid
  reinforcement learning.
\newblock {\em Advances in Neural Information Processing Systems}, 36.

\bibitem[Lu et~al., 2020]{Lu2020ADI}
Lu, W., Yu, Y., Chang, Y., Wang, Z., Li, C., and Yuan, B. (2020).
\newblock A dual input-aware factorization machine for ctr prediction.
\newblock In {\em International Joint Conference on Artificial Intelligence}.

\bibitem[Pandey et~al., 2007]{pandey2007bandits}
Pandey, S., Agarwal, D., Chakrabarti, D., and Josifovski, V. (2007).
\newblock Bandits for taxonomies: A model-based approach.
\newblock In {\em Proceedings of the 2007 SIAM international conference on data
  mining}, pages 216--227. SIAM.

\bibitem[Perakis and Roels, 2008]{perakis2008regret}
Perakis, G. and Roels, G. (2008).
\newblock Regret in the newsvendor model with partial information.
\newblock {\em Operations research}, 56(1):188--203.

\bibitem[Rajaraman et~al.,
  2020]{rajaraman2020fundamentallimitsimitationlearning}
Rajaraman, N., Yang, L.~F., Jiao, J., and Ramchandran, K. (2020).
\newblock Toward the fundamental limits of imitation learning.
\newblock In {\em Proceedings of the 34th International Conference on Neural
  Information Processing Systems}, NIPS '20, Red Hook, NY, USA. Curran
  Associates Inc.

\bibitem[Rashidinejad et~al.,
  2024]{rashidinejad2023bridgingofflinereinforcementlearning}
Rashidinejad, P., Zhu, B., Ma, C., Jiao, J., and Russell, S. (2024).
\newblock Bridging offline reinforcement learning and imitation learning: a
  tale of pessimism.
\newblock In {\em Proceedings of the 35th International Conference on Neural
  Information Processing Systems}, NIPS '21, Red Hook, NY, USA. Curran
  Associates Inc.

\bibitem[Ross and Bagnell, 2010]{Imitationlearningross}
Ross, S. and Bagnell, D. (2010).
\newblock Efficient reductions for imitation learning.
\newblock In Teh, Y.~W. and Titterington, M., editors, {\em Proceedings of the
  Thirteenth International Conference on Artificial Intelligence and
  Statistics}, volume~9 of {\em Proceedings of Machine Learning Research},
  pages 661--668, Chia Laguna Resort, Sardinia, Italy. PMLR.

\bibitem[Schwartz et~al., 2017]{advertising}
Schwartz, E.~M., Bradlow, E.~T., and Fader, P.~S. (2017).
\newblock Customer acquisition via display advertising using multi-armed bandit
  experiments.
\newblock {\em Marketing Science}, 36(4):500--522.

\bibitem[Shen, 2024]{deepctr_torch}
Shen, W. (2024).
\newblock Deepctr-torch: Easy-to-use, modular, and extendible pytorch framework
  for ctr prediction.
\newblock Accessed: 2024-11-18.

\bibitem[Shivaswamy and Joachims, 2012]{MABwithHistory}
Shivaswamy, P. and Joachims, T. (2012).
\newblock Multi-armed bandit problems with history.
\newblock In Lawrence, N.~D. and Girolami, M., editors, {\em Proceedings of the
  Fifteenth International Conference on Artificial Intelligence and
  Statistics}, volume~22 of {\em Proceedings of Machine Learning Research},
  pages 1046--1054, La Palma, Canary Islands. PMLR.

\bibitem[Song et~al., 2022]{Song2022-vg}
Song, Y., Zhou, Y., Sekhari, A., Andrew~Bagnell, J., Krishnamurthy, A., and
  Sun, W. (2022).
\newblock Hybrid {RL}: Using both offline and online data can make {RL}
  efficient.
\newblock {\em arXiv [cs.LG]}.

\bibitem[Swaminathan and Joachims, 2015a]{swaminathan2015batch}
Swaminathan, A. and Joachims, T. (2015a).
\newblock Batch learning from logged bandit feedback through counterfactual
  risk minimization.
\newblock {\em The Journal of Machine Learning Research}, 16(1):1731--1755.

\bibitem[Swaminathan and Joachims,
  2015b]{swaminathan2015counterfactualriskminimizationlearning}
Swaminathan, A. and Joachims, T. (2015b).
\newblock Batch learning from logged bandit feedback through counterfactual
  risk minimization.
\newblock {\em Journal of Machine Learning Research}, 16(52):1731--1755.

\bibitem[Thompson, 1933]{thompsonclinical}
Thompson, W.~R. (1933).
\newblock On the likelihood that one unknown probability exceeds another in
  view of the evidence of two samples.
\newblock {\em Biometrika}, 25(3/4):285--294.

\bibitem[Vershynin, 2018]{Vershynin_2018}
Vershynin, R. (2018).
\newblock {\em Frontmatter}, page i–ii.
\newblock Cambridge Series in Statistical and Probabilistic Mathematics.
  Cambridge University Press.

\bibitem[Wagenmaker and Pacchiano, 2023]{wagenmaker2023leveraging}
Wagenmaker, A. and Pacchiano, A. (2023).
\newblock Leveraging offline data in online reinforcement learning.
\newblock In {\em International Conference on Machine Learning}, pages
  35300--35338. PMLR.

\bibitem[Wang et~al., 2017]{wang2017deepcrossnetwork}
Wang, R., Fu, B., Fu, G., and Wang, M. (2017).
\newblock Deep \& cross network for ad click predictions.

\bibitem[Wang and Cukierski, 2014]{avazu-ctr-prediction}
Wang, S. and Cukierski, W. (2014).
\newblock Click-through rate prediction.
\newblock \url{https://kaggle.com/competitions/avazu-ctr-prediction}.
\newblock Kaggle.

\bibitem[Wu et~al., 2016]{wu2016conservative}
Wu, Y., Shariff, R., Lattimore, T., and Szepesv\'{a}ri, C. (2016).
\newblock Conservative bandits.
\newblock In {\em Proceedings of the 33rd International Conference on
  International Conference on Machine Learning - Volume 48}, ICML'16, page
  1254–1262. JMLR.org.

\bibitem[Wu et~al., 2019a]{wu2019behavior}
Wu, Y., Tucker, G., and Nachum, O. (2019a).
\newblock Behavior regularized offline reinforcement learning.
\newblock {\em arXiv preprint arXiv:1911.11361}.

\bibitem[Wu et~al., 2019b]{wu2019behaviorregularizedofflinereinforcement}
Wu, Y., Tucker, G., and Nachum, O. (2019b).
\newblock Behavior regularized offline reinforcement learning.

\bibitem[Xiao et~al., 2021a]{xiao2021optimalitybatchpolicyoptimization}
Xiao, C., Wu, Y., Lattimore, T., Dai, B., Mei, J., Li, L., Szepesvari, C., and
  Schuurmans, D. (2021a).
\newblock On the optimality of batch policy optimization algorithms.
\newblock In {\em International Conference on Machine Learning}.

\bibitem[Xiao et~al., 2021b]{Xiao2021OnTO}
Xiao, C., Wu, Y., Lattimore, T., Dai, B., Mei, J., Li, L., Szepesvari, C., and
  Schuurmans, D. (2021b).
\newblock On the optimality of batch policy optimization algorithms.
\newblock In {\em International Conference on Machine Learning}.

\bibitem[Xie et~al., 2022]{xie2022armormodelbasedframeworkimproving}
Xie, T., Bhardwaj, M., Jiang, N., and Cheng, C.-A. (2022).
\newblock Armor: A model-based framework for improving arbitrary baseline
  policies with offline data.

\bibitem[Xie et~al., 2024]{xie2022policyfinetuningbridgingsampleefficient}
Xie, T., Jiang, N., Wang, H., Xiong, C., and Bai, Y. (2024).
\newblock Policy finetuning: bridging sample-efficient offline and online
  reinforcement learning.
\newblock In {\em Proceedings of the 35th International Conference on Neural
  Information Processing Systems}, NIPS '21, Red Hook, NY, USA. Curran
  Associates Inc.

\bibitem[Xu et~al., 2022]{xu2022robust}
Xu, L., Zheng, Y., and Jiang, L. (2022).
\newblock A robust data-driven approach for the newsvendor problem with
  nonparametric information.
\newblock {\em Manufacturing \& Service Operations Management}, 24(1):504--523.

\bibitem[Yin et~al., 2021]{yin2020nearoptimalprovableuniformconvergence}
Yin, M., Bai, Y., and Wang, Y.-X. (2021).
\newblock Near-optimal provable uniform convergence in offline policy
  evaluation for reinforcement learning.
\newblock In {\em International Conference on Artificial Intelligence and
  Statistics}.

\bibitem[Yin and Wang, 2021]{yin2021towards}
Yin, M. and Wang, Y.-X. (2021).
\newblock Towards instance-optimal offline reinforcement learning with
  pessimism.
\newblock {\em Advances in neural information processing systems},
  34:4065--4078.

\bibitem[Yu et~al., 2020]{yu2020mopomodelbasedofflinepolicy}
Yu, T., Thomas, G., Yu, L., Ermon, S., Zou, J., Levine, S., Finn, C., and Ma,
  T. (2020).
\newblock Mopo: model-based offline policy optimization.
\newblock In {\em Proceedings of the 34th International Conference on Neural
  Information Processing Systems}, NIPS '20, Red Hook, NY, USA. Curran
  Associates Inc.

\bibitem[Zheng et~al., 2023]{zheng2023adaptivepolicylearningofflinetoonline}
Zheng, H., Luo, X., Wei, P., Song, X., Li, D., and Jiang, J. (2023).
\newblock Adaptive policy learning for offline-to-online reinforcement
  learning.
\newblock In {\em Proceedings of the Thirty-Seventh AAAI Conference on
  Artificial Intelligence and Thirty-Fifth Conference on Innovative
  Applications of Artificial Intelligence and Thirteenth Symposium on
  Educational Advances in Artificial Intelligence}, AAAI'23/IAAI'23/EAAI'23.
  AAAI Press.

\bibitem[Zhou et~al., 2024]{zhou2024sequential}
Zhou, Y., Fu, M.~C., and Ryzhov, I.~O. (2024).
\newblock Sequential learning with a similarity selection index.
\newblock {\em Operations Research}, 72(6):2526--2542.

\bibitem[Zhou et~al., 2023]{zhou2023offlinedataenhancedonpolicy}
Zhou, Y., Sekhari, A., Song, Y., and Sun, W. (2023).
\newblock Offline data enhanced on-policy policy gradient with provable
  guarantees.

\end{thebibliography}
